%% file: main.tex
\documentclass{bmvc2k}
\usepackage{makecell}
\usepackage{amsmath}
\usepackage{tikz}
\DeclareMathOperator*{\argmax}{argmax}
\usepackage{rotating}
\usepackage{float}
\usepackage{hyperref}
\usepackage{xr}

\newcommand\blfootnote[1]{%
	\begingroup
	\renewcommand\thefootnote{}\footnote{#1}%
	\addtocounter{footnote}{-1}%
	\endgroup
}

\newcommand{\beginsupplement}{%
	\setcounter{table}{0}
	\renewcommand{\thetable}{S\arabic{table}}%
	\setcounter{figure}{0}
	\renewcommand{\thefigure}{S\arabic{figure}}%
	\setcounter{section}{0}
	\renewcommand{\thesection}{\Alph{section}}
}

\title{READMem: Robust Embedding Association for a Diverse Memory in Unconstrained Video Object Segmentation}

\addauthor{Stéphane Vujasinovi\'{c}}{stephane.vujasinovic@iosb.fraunhofer.de}{1}
\addauthor{Sebastian Bullinger}{sebastian.bullinger@iosb.fraunhofer.de}{1}
\addauthor{Stefan Becker}{stefan.becker@iosb.fraunhofer.de}{1}
\addauthor{Norbert Scherer-Negenborn}{norbert.scherer-negenborn@iosb.fraunhofer.de}{1}
\addauthor{Michael Arens}{michael.arens@iosb.fraunhofer.de}{1}
\addauthor{Rainer Stiefelhagen}{rainer.stiefelhagen@kit.edu}{2}

\addinstitution{
	Fraunhofer IOSB\textsuperscript{{\normalsize\text{$^\star$}}}\\
	Ettlingen, Germany
}
\addinstitution{
	Karlsruhe Institut of Technology\\
	Karlsruhe, Germany
}

\runninghead{
	Vujasinovi\'{c},
	Bullinger,
	Becker,
	Scherer,
	Arens,
	Stiefelhagen
}{READMem}

\def\eg{\emph{e.g}\bmvaOneDot}

\def\etal{\emph{et al}\bmvaOneDot}

\input{macro/macro.tex}
\usepackage{caption}

\DeclareCaptionFont{BMVCblue}{\color[rgb]{0,.0,.4}}
\begin{document}

\newcommand{\tabref}[1]{Table~\ref{#1}}
\newcommand{\BV}[1]{\textbf{#1}}

\definecolor{deeppink}{rgb}{1.0, 0.08, 0.58}
\definecolor{electricgreen}{rgb}{0.0, 1.0, 0.0}
\definecolor{greentxt}{rgb}{0.0, 0.80, 0.33}
\definecolor{IBMINDIGO}{HTML}{0000FF}
\definecolor{IBMMAGENTA2}{HTML}{9D02D7}
\definecolor{IBMMAGENTA1}{HTML}{CD34B5}
\definecolor{IBMPINK}{HTML}{EA5F94}
\definecolor{IBMLIGHTORANGE}{HTML}{FA8775}
\definecolor{IBMORANGE}{HTML}{FFB14E}
\definecolor{IBMGOLD}{HTML}{FFD700}
\definecolor{IBMTURQUOISE}{HTML}{22D8D7}

\maketitle

\begin{abstract}
We present READMem (Robust Embedding Association for a Diverse Memory), a modular framework for semi-automatic video object segmentation (sVOS) methods designed to handle unconstrained videos. 
Contemporary sVOS works typically aggregate video frames in an ever-expanding memory, demanding high hardware resources for long-term applications.
To mitigate memory requirements and prevent near object duplicates (caused by information of adjacent frames), previous methods introduce a hyper-parameter that controls the frequency of frames eligible to be stored.
This parameter has to be adjusted according to concrete video properties (such as rapidity of appearance changes and video length) and does not generalize well.
Instead, we integrate the embedding of a new frame into the memory only if it increases the diversity of the memory content.
Furthermore, we propose a robust association of the embeddings stored in the memory with the query embeddings during the update process.
Our approach avoids the accumulation of redundant data, allowing us in return, to restrict the memory size and prevent extreme memory demands in long videos.
We extend popular sVOS baselines with READMem, which previously showed limited performance on long videos. 
Our approach achieves competitive results on the Long-time Video dataset (LV1) while not hindering performance on short sequences. Our code is publicly available at \url{https://github.com/Vujas-Eteph/READMem}.
\end{abstract}


\definecolor{IBMINDIGO}{HTML}{0000FF}
\definecolor{IBMMAGENTA2}{HTML}{9D02D7}
\definecolor{IBMMAGENTA1}{HTML}{CD34B5}
\definecolor{IBMPINK}{HTML}{EA5F94}
\definecolor{IBMLIGHTORANGE}{HTML}{FA8775}
\definecolor{IBMORANGE}{HTML}{FFB14E}
\definecolor{IBMGOLD}{HTML}{FFD700}

\graphicspath{{./figures}}

\input{sections/introduction}

\input{sections/related_work}

\input{sections/methods}

\input{sections/results}

\input{sections/conclusion}

\bibliography{bibliography_vujasinovic}

\beginsupplement
\input{supplementary}

\end{document}

%% file: macro/macro.tex
\usepackage{xspace}
\usepackage{siunitx}

\def\numx#1e#2{{#1}\mathrm{\times 10}{^{#2}}}

\usepackage{xcolor}

\usepackage{multirow}
\usepackage{graphicx}
\usepackage{booktabs}

\usepackage{amssymb}
\usepackage{amsmath}

\usepackage{nccmath}

\usepackage{pifont}

\usepackage{mathtools}

\usepackage{tcolorbox}

\usepackage{colortbl}

\usepackage{array}

\newcolumntype{P}[1]{>{\centering\arraybackslash}p{#1}}

\usepackage[printonlyused]{acronym}
\acrodef{iou}[IoU]{Intersection over Union}
\acrodef{for}[\textit{FoR}]{\textit{frame of reference}}
\acrodef{convnet}[ConvNet]{Convolutional Neural Network}

\newcommand{\figref}[1]{Figure~\ref{#1}}

\newcommand{\secref}[1]{Section~\ref{#1}}

\newcommand{\mset}[1]{\ensuremath{\mathcal{#1}}}
\newcommand{\mtensor}[1]{\ensuremath{\mathbf{#1}}}

\newcommand*{\tran}{^{\mkern-1.5mu\mathsf{T}}}

\makeatletter
\DeclareRobustCommand\onedot{\futurelet\@let@token\@onedot}
\def\@onedot{\ifx\@let@token.\else.\null\fi\xspace}                
\def\eg{\emph{e.g}\onedot}             
            
\def\ie{\emph{i.e}\onedot}             
            
\def\vs{\emph{vs}\onedot}            
\def\wrt{w.r.t\onedot}                  
                   
\def\etal{\emph{et al}\@onedot}          

\makeatother

\def\JandF{$\mathcal{J}\mathcal{\&}\mathcal{F}$\xspace}

\def\JF@s{$\mathcal{J}\mathcal{\&}\mathcal{F}@60s$\xspace}
\def\J{$\mathcal{J}$\xspace}
\def\F{$\mathcal{F}$\xspace}
\def\JandFD17{$\mathcal{J}\mathcal{\&}\mathcal{F}_{\text{D}17}$\xspace}
\def\JandFLV1{$\mathcal{J}\mathcal{\&}\mathcal{F}_{\text{LV}1}$\xspace}

\def\MiVOS{MiVOS~\cite{MiVOS}\xspace}
\def\STCN{STCN~\cite{STCN}\xspace}
\def\XMEM{XMem~\cite{XMEM}\xspace}

\def\QDMN{QDMN~\cite{QDMN}\xspace}

\def\cSTM{\cite{STM}\xspace}

\def\cJOINT{\cite{JOINT}\xspace}

\def\cAFBURR{\cite{AFB-URR}\xspace}

\def\cTHOR{\cite{Sauer2019BMVC}}

\def\citexonlyxSTMxbasedxmethods{\cite{MiVOS,STCN,QDMN,XMEM,GSFM,RDE-VOS,SWEM}\xspace}

\colorlet{semigray}{black!40}
\newcommand{\refConf}[1]{\textcolor{black!50!white}{(#1)}}

\def\DeepLabv3+{DeepLabv3+~\cite{Chen2018}}
\def\ResNet50{ResNet50~\cite{He2016}}

\def\DAVIS17{DAVIS 2017~\cite{Pont-Tuset_arXiv_2017}\xspace}
\def\D17{D17~\cite{Pont-Tuset_arXiv_2017}\xspace}
\def\LV1{LV1~\cite{AFB-URR}\xspace}
\def\LongTimeVideo{Long-time Video~\cite{AFB-URR}\xspace}

\def\BL30K{{\href{https://github.com/hkchengrex/MiVOS}{BL30K}}~\cite{Cheng_CVPR_2021}}

\def\Pascal_VOC{PASCAL VOC~\cite{Everingham2011}}

\def\sota{state-of-the-art\xspace}

\def\vos{video object segmentation\xspace}

\def\svos{semi-automatic \vos}


%% file: sections/introduction.tex
\section{Introduction}
\label{sec:introduction}

Video object segmentation~(VOS),~\ie, the pixel-level discrimination of a foreground object from the background in videos, is a fundamental and challenging problem in computer vision. 
This paper focuses on the~\svos~(sVOS) setting~\cite{zhou2023survey}, where the goal consists of segmenting an object (or a collection of objects) in a video by only providing a segmentation mask as input.

\begin{figure}[t]
	\centering
	\begin{tabular}{c}
		\hspace{-4mm}
		\bmvaHangBox{\includegraphics[trim= {0 3mm 0 0},clip, width=0.98\textwidth]{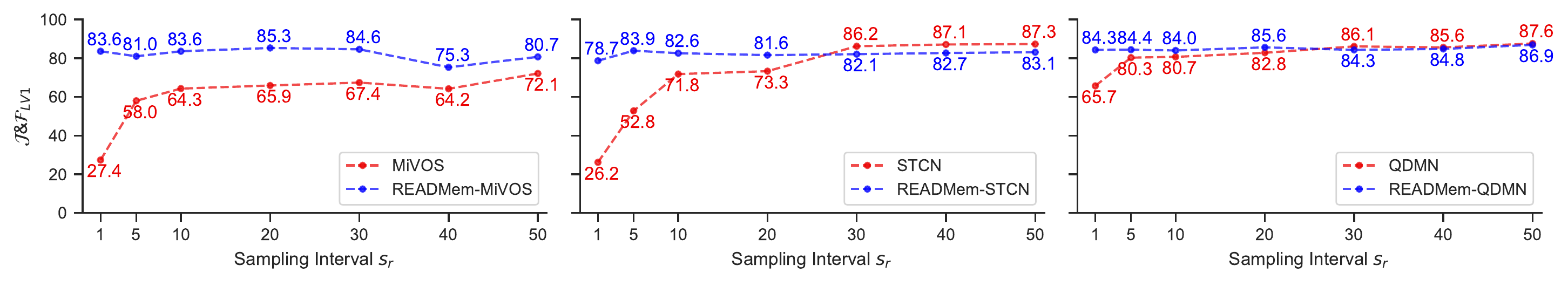}}
	\end{tabular}
	\begin{tabular}{P{4cm}P{4cm}P{4cm}}
		\hspace{+3mm}{\scriptsize (a) MiVOS \vs READMem-MiVOS} & \hspace{-4mm} {\scriptsize (b) STCN \vs READMem-STCN} & \hspace{-12mm} {\scriptsize (c) QDMN \vs READMem-QDMN}
	\end{tabular}
	\caption{Performance comparison of sVOS baselines (\MiVOS, \STCN, \QDMN) with and without the READMem extension on the \LV1 dataset when varying the sampling interval~$s_r$.
		The configuration employed for these experiments is consistent to the quantitative analysis in~\secref{sec:quantitativexresults}.
	}
	\label{FIGxBaselinesxVSxHoliMemxLV1xALLxinxOne}
\end{figure}

\textbf{Motivation}:
Current sVOS methods~\citexonlyxSTMxbasedxmethods predominantly utilize the space-time memory network design~\cite{STM}, which stores deep representations of previous frames and their estimated masks in an external memory for the segmentation of a query image.
Although some attention has recently been given to the memory update strategy~\cite{XMEM, QDMN, SWEM}, most methods rely on a simple frame aggregation approach, which inserts every~$t$-th frame into an ever-expanding memory. 
This approach works well for short sequences but fails on longer videos due to the saturation of the GPU memory. 
Furthermore, this strategy neither considers redundancies nor novel information while adding new embeddings to the memory.
A naive approach to better deal with long videos is to increase the sampling interval~($s_r$) of~\sota methods~\cite{STM,MiVOS,STCN}, as shown in~\figref{FIGxBaselinesxVSxHoliMemxLV1xALLxinxOne}, which displays that increasing the sampling interval leads to better performances. 
Due to a higher sampling interval, the memory stores embeddings of frames that are further apart in time from each other. 
These embeddings are more likely to reflect appearance variations resulting in a more diverse set of embeddings in the memory, contributing to better predictions.
However, a high sampling interval may lead to the omission of potential useful frames~\cite{AFB-URR,XMEM}, which is particularly detrimental for short sequences.
Moreover, as noted by~\cite{XMEM}, a higher sampling interval leads to unstable performances.

To address the aforementioned points, we propose a simple yet effective extension in the form of a module, named \textit{READMem}, for updating the memory.
Drawing inspiration from~\cTHOR\xspace in the visual object tracking field, our approach focuses on increasing the diversity of the embeddings stored in the memory, rather than simply aggregating every $t$-th frame.
This selective approach allows us to save only key embeddings in the memory, thereby alleviating the memory saturation constraint faced by previous methods on long videos and even on unconstrained video lengths.

\begin{figure}[t]
	\begin{tabular}{ccccccc}
		\bmvaHangBox{
			\includegraphics[width=0.135\textwidth]{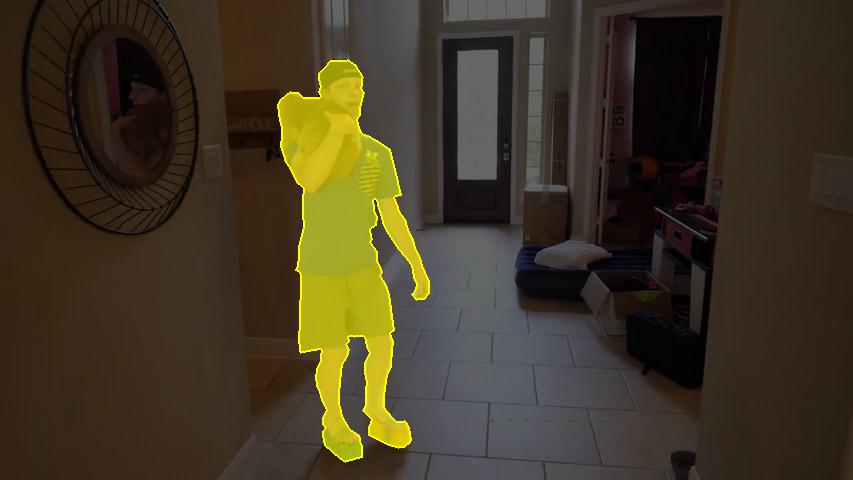}
			\put(-48.0,23.0){\makebox[0pt][l]{{\color{IBMGOLD}\textbf{\tiny \#180}}}}}
		&\hspace{-4.5mm}
		\bmvaHangBox{\includegraphics[width=0.135\textwidth]{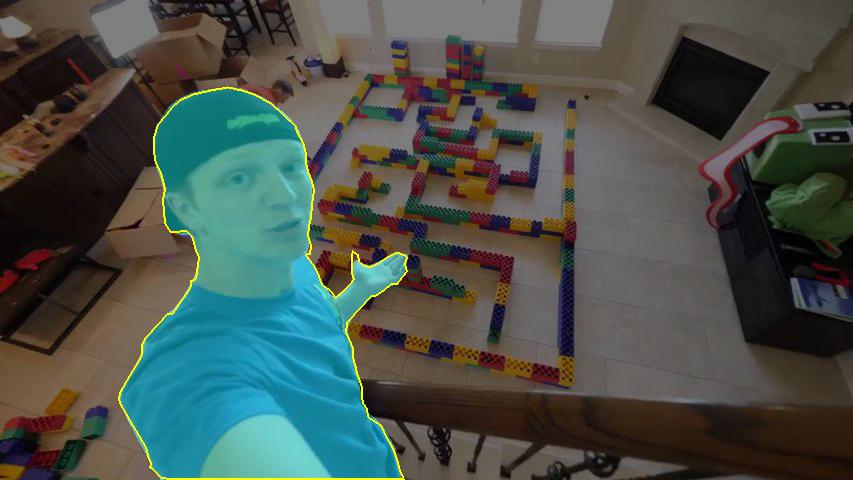}
			\put(-48.0,23.0){\makebox[0pt][l]{{\color{IBMGOLD}\textbf{\tiny \#4941}}}}}
		&\hspace{-4.5mm}
		\bmvaHangBox{\includegraphics[width=0.135\textwidth]{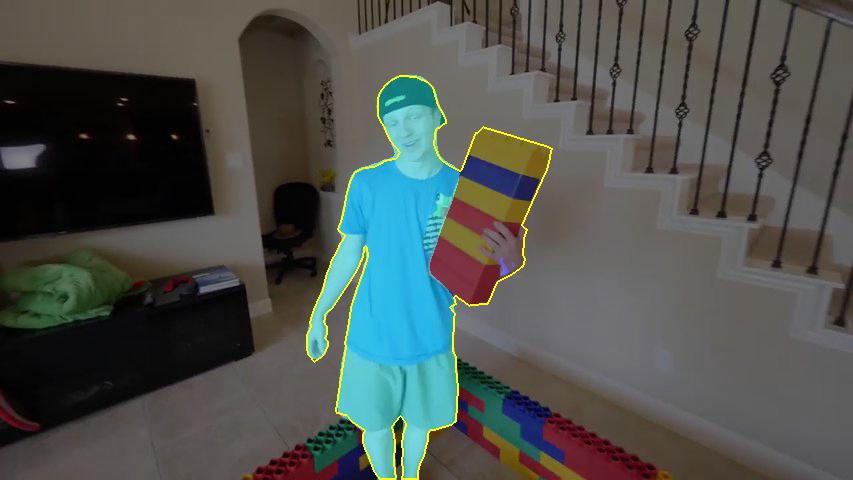}
			\put(-48.0,23.0){\makebox[0pt][l]{{\color{IBMGOLD}\textbf{\tiny \#6057}}}}}
		&\hspace{-4.5mm}
		\bmvaHangBox{\includegraphics[width=0.135\textwidth]{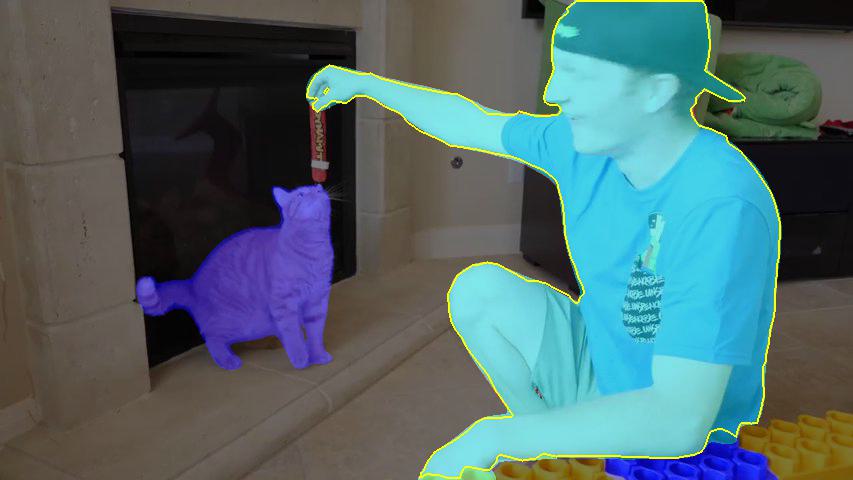}
			\put(-48.0,23.0){\makebox[0pt][l]{{\color{IBMGOLD}\textbf{\tiny \#11139}}}}}
		&\hspace{-4.5mm}
		\bmvaHangBox{\includegraphics[width=0.135\textwidth]{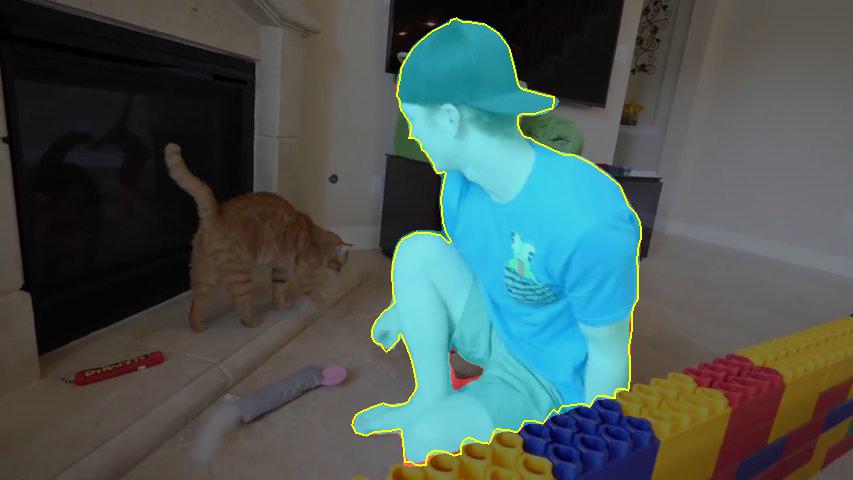}
			\put(-48.0,23.0){\makebox[0pt][l]{{\color{IBMGOLD}\textbf{\tiny \#11895}}}}}
		&\hspace{-4.5mm}
		\bmvaHangBox{\includegraphics[width=0.135\textwidth]{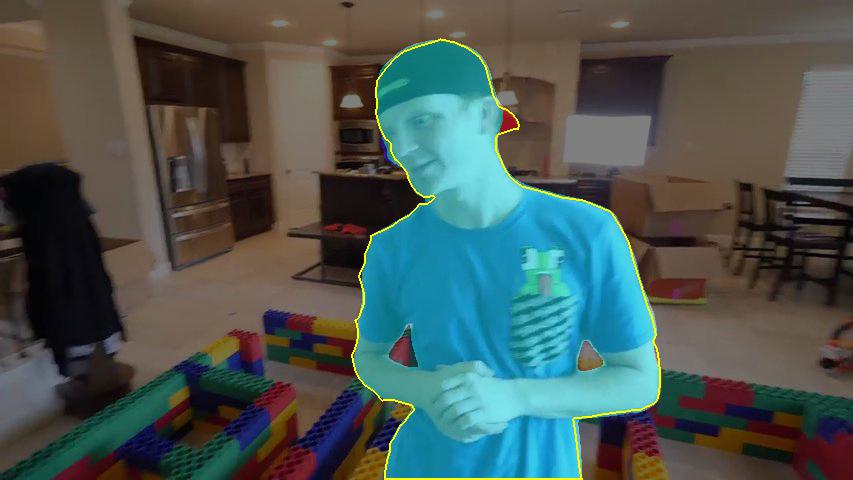}
			\put(-48.0,23.0){\makebox[0pt][l]{{\color{IBMGOLD}\textbf{\tiny \#12930}}}}}
		&\hspace{-4.5mm}
		\bmvaHangBox{\includegraphics[width=0.135\textwidth]{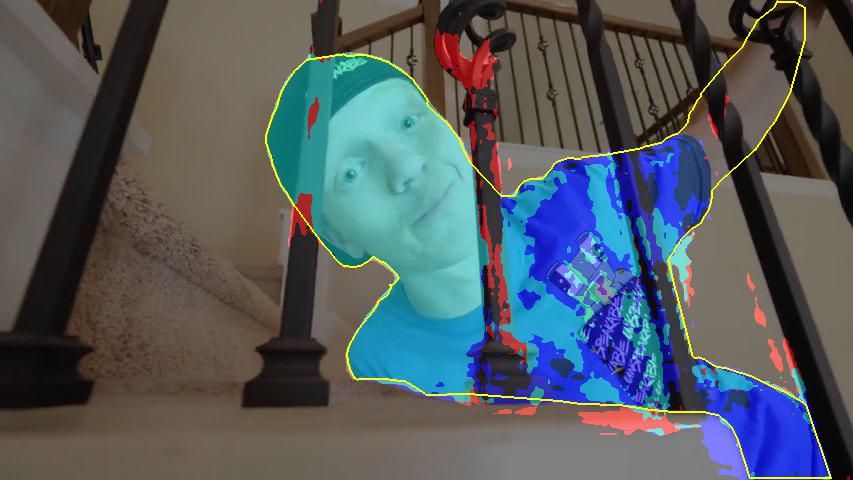}
			\put(-48.0,23.0){\makebox[0pt][l]{{\color{IBMGOLD}\textbf{\tiny \#18252}}}}}
		\\
		\bmvaHangBox{\includegraphics[width=0.135\textwidth]{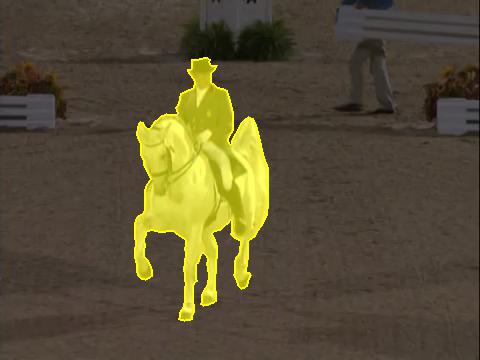}
			\put(-48.0,32.0){\makebox[0pt][l]{{\color{IBMGOLD}\textbf{\tiny \#1947}}}}}
		&\hspace{-4.5mm}
		\bmvaHangBox{\includegraphics[width=0.135\textwidth]{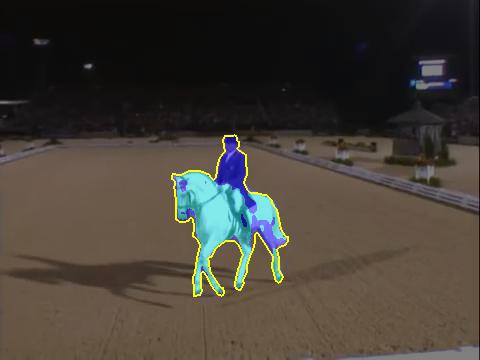}
			\put(-48.0,32.0){\makebox[0pt][l]{{\color{IBMGOLD}\textbf{\tiny \#3075}}}}}
		&\hspace{-4.5mm}
		\bmvaHangBox{\includegraphics[width=0.135\textwidth]{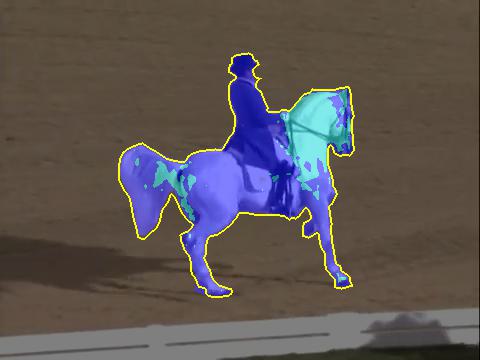}
			\put(-48.0,32.0){\makebox[0pt][l]{{\color{IBMGOLD}\textbf{\tiny \#3639}}}}}
		&\hspace{-4.5mm}
		\bmvaHangBox{\includegraphics[width=0.135\textwidth]{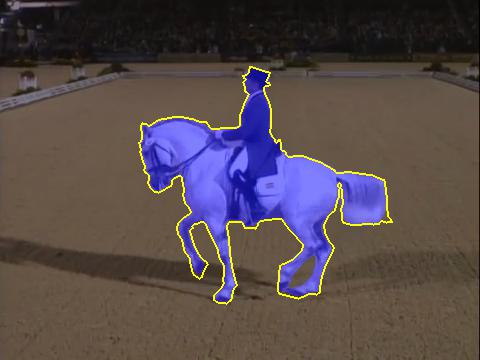}
			\put(-48.0,32.0){\makebox[0pt][l]{{\color{IBMGOLD}\textbf{\tiny \#4767}}}}}
		&\hspace{-4.5mm}
		\bmvaHangBox{\includegraphics[width=0.135\textwidth]{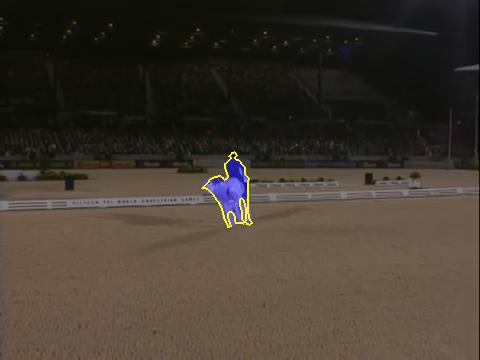}
			\put(-48.0,32.0){\makebox[0pt][l]{{\color{IBMGOLD}\textbf{\tiny \#10407}}}}}
		&\hspace{-4.5mm}
		\bmvaHangBox{\includegraphics[width=0.135\textwidth]{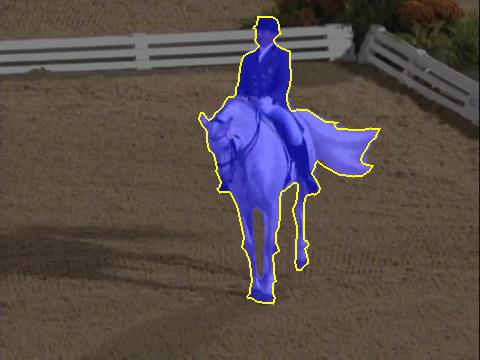}
			\put(-48.0,32.0){\makebox[0pt][l]{{\color{IBMGOLD}\textbf{\tiny \#12099}}}}}
		&\hspace{-4.5mm}
		\bmvaHangBox{\includegraphics[width=0.135\textwidth]{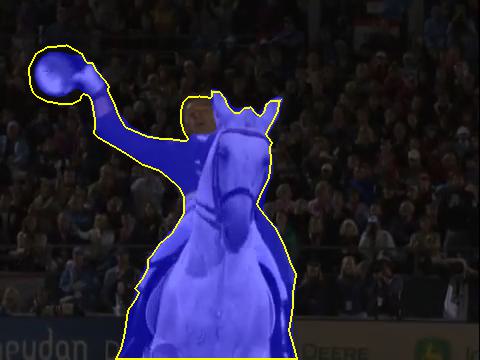}
			\put(-48.0,32.0){\makebox[0pt][l]{{\color{IBMGOLD}\textbf{\tiny \#12711}}}}}
		\\
		\bmvaHangBox{\includegraphics[width=0.135\textwidth]{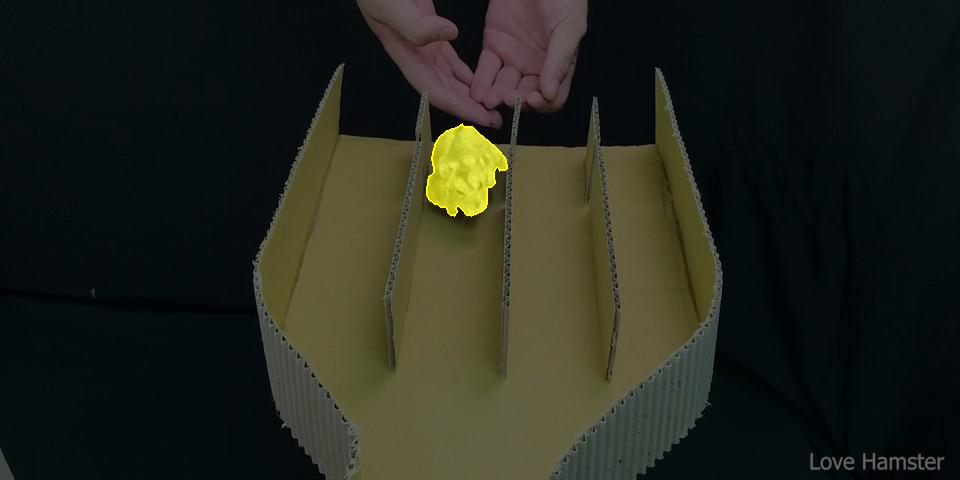}
			\put(-48.0,20.0){\makebox[0pt][l]{{\color{IBMGOLD}\textbf{\tiny \#1479}}}}}
		&\hspace{-4.5mm}
		\bmvaHangBox{\includegraphics[width=0.135\textwidth]{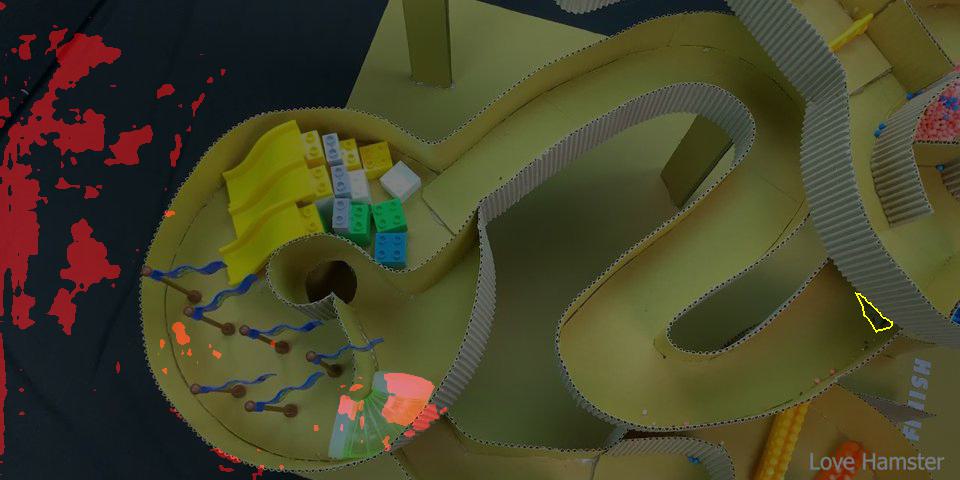}
			\put(-48.0,20.0){\makebox[0pt][l]{{\color{IBMGOLD}\textbf{\tiny \#2931}}}}}
		&\hspace{-4.5mm}
		\bmvaHangBox{\includegraphics[width=0.135\textwidth]{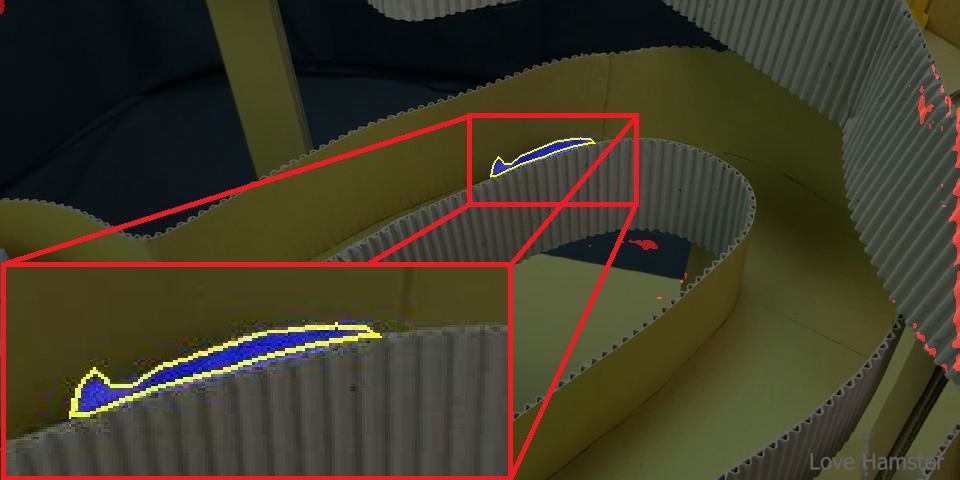}
			\put(-48.0,20.0){\makebox[0pt][l]{{\color{IBMGOLD}\textbf{\tiny \#4485}}}}}
		&\hspace{-4.5mm}
		\bmvaHangBox{\includegraphics[width=0.135\textwidth]{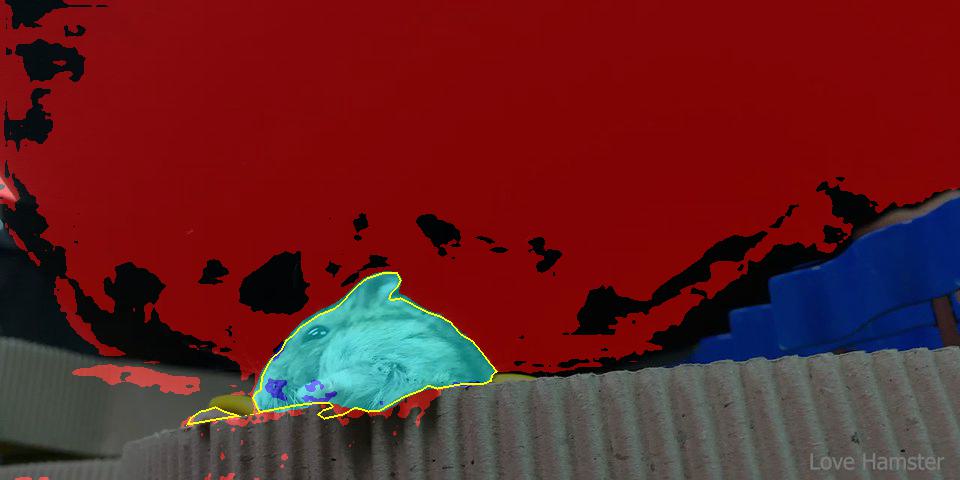}
			\put(-48.0,20.0){\makebox[0pt][l]{{\color{IBMGOLD}\textbf{\tiny \#4929}}}}}
		&\hspace{-4.5mm}
		\bmvaHangBox{\includegraphics[width=0.135\textwidth]{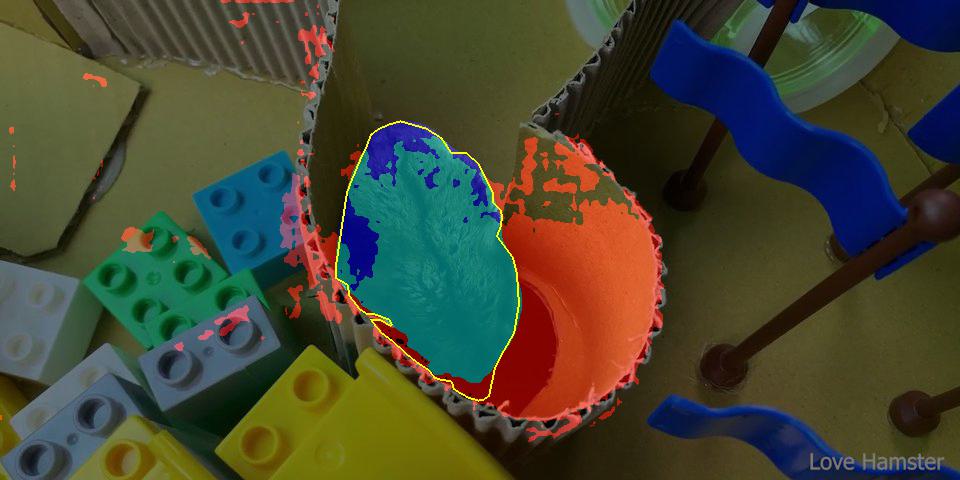}
			\put(-48.0,20.0){\makebox[0pt][l]{{\color{IBMGOLD}\textbf{\tiny \#5373}}}}}
		&\hspace{-4.5mm}
		\bmvaHangBox{\includegraphics[width=0.135\textwidth]{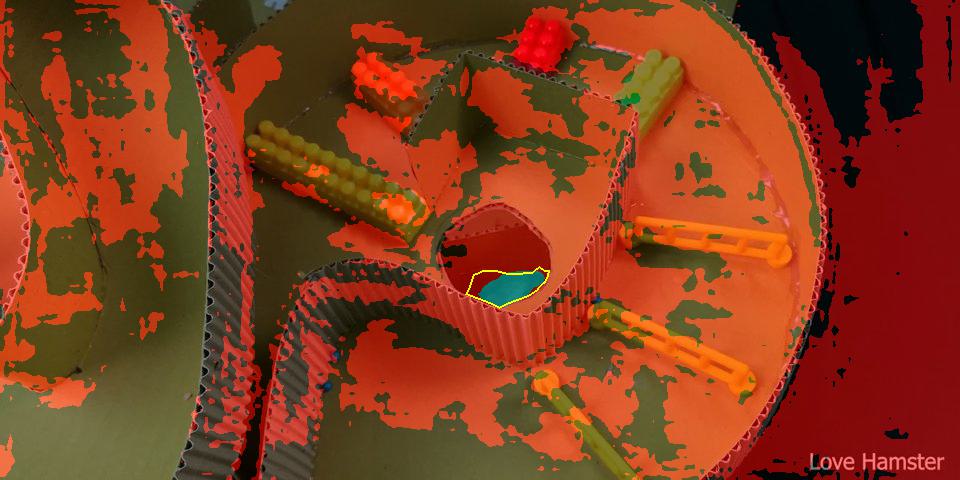}
			\put(-48.0,20.0){\makebox[0pt][l]{{\color{IBMGOLD}\textbf{\tiny \#5817}}}}}
		&\hspace{-4.5mm}
		\bmvaHangBox{\includegraphics[width=0.135\textwidth]{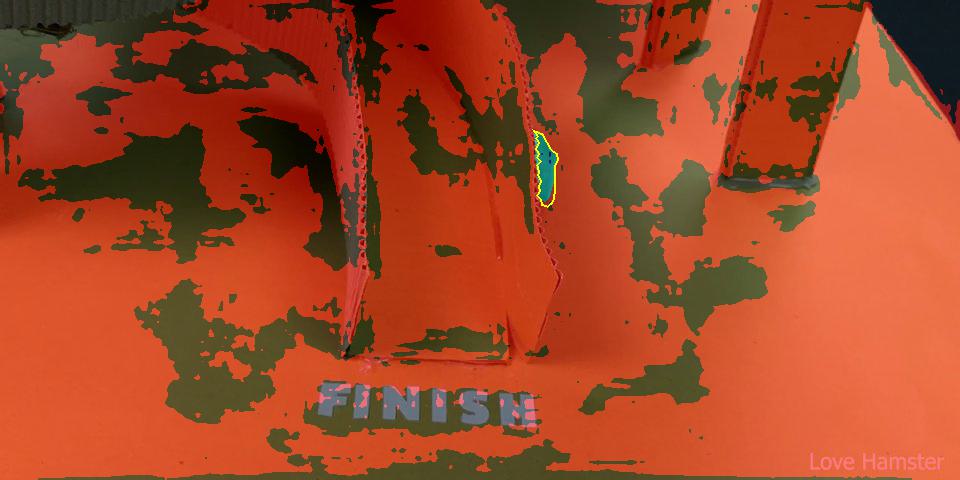}
			\put(-48.0,20.0){\makebox[0pt][l]{{\color{IBMGOLD}\textbf{\tiny \#6039}}}}}
	\end{tabular}
	\caption{Results of \MiVOS (baseline) in~{\color{red!70!black}red} and READMem-MiVOS (READMem added to \MiVOS) in~{\color{IBMINDIGO!}blue} on the~\LV1 dataset.
		We depict the intersection between both predictions in~{\color{IBMTURQUOISE!70!black}turquoise}.
		We indicate the ground-truth with {\color{IBMGOLD!70!black}yellow} contours.
		The first column displays the input frames (\ie, ground-truth mask in {\color{IBMGOLD!70!black}yellow}) for each sequence of the~\LV1 dataset used to initialize the methods.
		The experimental configuration of these results is consistent with the quantitative experiments in \secref{sec:quantitativexresults}.
	}
	\label{fig:LVONE_MiVOS_Holi_MiVOS_All_in_One}
\end{figure}

\vspace{1mm}
\textbf{Our contributions are as follows}: (1)~We propose READMem, a novel module suitable to extend existing sVOS methods to handle unconstrained videos (arbitrary rapidity of appearance changes and arbitrary video length) without requiring re-training or fine-tuning the sVOS baseline. 
Our approach improves the diversity of the embeddings stored in the memory and performs a robust association with query embeddings during the update process.
(2)~When using a small sampling interval (which is desired since this improves the stability of the methods~\cite{XMEM} and avoids missing key information~\cite{XMEM,AFB-URR}) the proposed READMem module consistently improves the results of current sVOS baselines on long videos~(\LV1), while not hindering the performance on short videos (\ie, 2017 DAVIS validation set~\cite{Pont-Tuset_arXiv_2017} (D17)). Overall, we attain competitive results against the state-of-the-art. 
(3)~We provide insight into our approach through an extensive ablation study. 
(4)~Finally, we make our code publicly available to promote reproducibility and facilitate the implementation of our method.

%% file: sections/related_work.tex
\section{Relate Work}
\label{sec:relatework}
\textbf{Short-term sVOS Methods}:
\textit{Online fine-tuning approaches}~\cite{voigtlaenderonline, caelles2017one, maninis2018video, xiao2018monet, xiao2019online, meinhardt2020make} adapt the network parameters on-the-fly based on an initial mask indicating the object of interest, but suffer from slow inference and poor generalization~\cite{zhou2023survey}. 

In contrast, \textit{propagation-based methods}~\cite{perazzi2017learning, jang2017online, hu2018motion, zhang2019fast, johnander2019generative} utilize strong priors offered by previous (adjacent) frames to propagate the mask, allowing them to better deal with fast appearance changes.
However, they are prone to error accumulation and lack long-term context for identifying objects after occlusion~\cite{zhou2023survey}.

\textit{Matching-based methods}~\cite{zhou2023survey}
use features of the initial frame along with previously processed frames (intermediate frames) as references for feature matching during inference.
In~\cite{voigtlaender2019feelvos,CFBI,CFBIP,AOT,DEAOT,TBD}, the embeddings of the initial frame and the adjacent frame are matched with the embeddings of the query frame through global and local matching, while Zhou \etal~\cJOINT also combine online-fine tuning approaches.
However, leading sVOS methods rely mostly on the Space-Time Memory~(STM) design introduced by Oh \etal~\cSTM .
Using a non-local approach, the method matches the embeddings from the query frame with the embeddings of reference frames.
Based on STM~\cSTM, Xie \etal~\cite{RMNet} and Seong \etal~\cite{KMN} perform a local-to-local matching to alleviate distractors-based mismatches. A memory matching at multiple scales is also conducted by Seong \etal~\cite{HMMN}. 
To enhance pixel discrimination Yong \etal~\cite{GSFM} explore a unique approach that uses both, the frequency and spectral domains. 
To reduce noisy predictions due to the ever-expanding memory, Cheng \etal~\cite{MiVOS} add a top-$k$ filter during non-local matching and propose an efficient extension~\cite{STCN}, which decouples the feature extraction of the frame and corresponding mask while improving the matching process.
Park \etal~\cite{PCVOS} deviate from the frame-to-frame propagation scheme and adopt instead a clip-to-clip approach to speed-up the propagation process.
However, as space-time memory networks-based methods display incredible performance on short-term benchmarks~\cite{Pont-Tuset_arXiv_2017, xu2018youtube}, they are hampered in real-world applications by their ever-expanding memory.

\textbf{Long-term sVOS Methods}: To address memory limitations, Li~\etal~\cite{FVOSGCM} propose a compact global module to summarize object segmentation information within a fixed memory size.
Similarly, Liang~\etal~\cite{AFB-URR} apply a weighted average on the extracted features to merge new features with existing ones into the memory and discard obsolete features through a \textit{least frequently used} (LFU) approach.
Liu~\etal~\cite{QDMN} explore the use of a quality-aware-module~(QAM)~\cite{huang2019msrcnn} to assess the quality of the predicted mask on-the-fly before integrating it in the memory. 
Li~\etal~\cite{RDE-VOS} use a spatio-temporal aggregation module to update a fixed-sized memory bank.
Cheng and Schwing~\cite{XMEM} follow the Atkinson-Shiffrin~\cite{atkinson1968human} model for their sVOS pipeline (XMem), which comprises: 
a \textit{sensory memory}~\cite{cho2014properties} that learns an appearance model for every incoming frame, a \textit{working memory} based on STCN~\cite{STCN}, and a \textit{long-term memory} that consolidates the working memory embeddings in a compact representation.

%% file: sections/methods.tex
\section{Method}
\label{sec:method}
\begin{figure*}[t]
	\centering
	\includegraphics[width=1\textwidth]{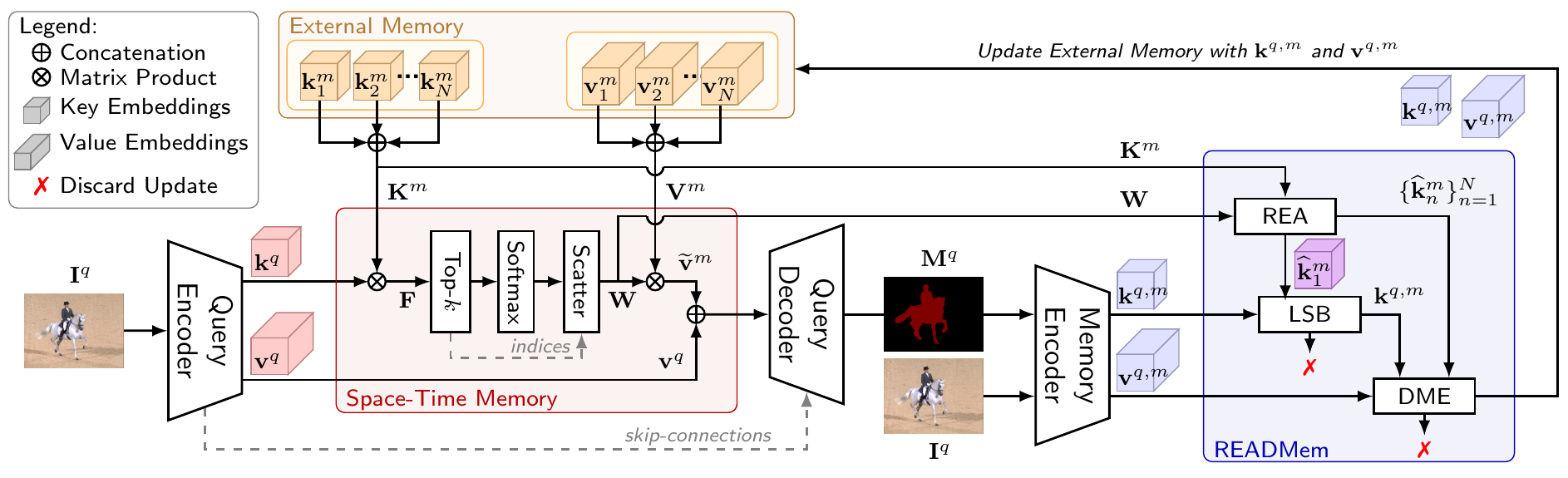}
	\caption{Overview of READMem-MiVOS~(\MiVOS using READMem) after initialization. We omit the temporary memory for simplicity.
	}
	\label{fig:Overview}
\end{figure*}

This section presents our READMem module~\textendash\xspace an extension for space-time memory-based sVOS pipelines~\cite{STM,MiVOS,STCN}.
\figref{fig:Overview} illustrates our module embedded in~\MiVOS.

\subsection{Video Object Segmentation Model Structure}
\label{subsectionVOSbaseline}
Since READMem is built upon the popular space-time memory network paradigm~\cite{STM}, we provide a brief description of the corresponding components.

(1) A query encoder~(an extension of~\ResNet50) that encodes a query frame~$\mtensor{I}^q$ seen at time step~$t$ into a pair of a query key feature~$\mtensor{k}^q \in \mathbb{R}^{C_k \times HW}$ and a query value feature $\mtensor{v}^q \in \mathbb{R}^{C_v \times HW}$.
With~$C_k$ and~$C_v$ we denote the number of feature channels, while~$H$ and~$W$ indicate the spatial image dimensions using a stride of 16.

(2) A memory encoder that extracts latent feature representations from a frame and its corresponding segmentation mask into a pair of a memory key~$\mtensor{k}^m \in \mathbb{R}^{C_k \times HW}$ and a memory value~$\mtensor{v}^m \in \mathbb{R}^{C_v \times HW}$.
The key features encode visual semantic information that is robust to appearance changes (needed for soft-attention, refer to (4)), while value features encode boundary and texture cues~\cite{STM} useful at the decoder stage.

(3) An external memory that maintains a set of memory keys~$\mset{M}^k = \big\{\mtensor{k}_n^m\big\}_{n=1}^{N}$ and a set of memory values~$\mset{M}^v = \big\{\mtensor{v}_n^m\big\}_{n=1}^{N}$.
The variable~$N$ denotes the total number of memory slots, while $n$ represents the index of a slot.
In addition, some methods~\cite{STM,MiVOS,QDMN} use a temporary memory that only contains the latent representation (\ie, memory key and value features) of the previous adjacent frame.

(4) A \textit{Space-Time Memory}~(STM) component that serves as an attention mechanism to determine relevant memory information. 
It matches the channels of the query key~$\mtensor{k}^q$ with all channels of every memory key~$\mtensor{k}^m_n \in \mset{M}^k$ on a temporal and spatial aspect by inferring an affinity matrix~$\mtensor{F} \in \mathbb{R}^{NHW\times HW}$.
Concretely, the channel level similarities of the query~$\mtensor{k}^q$ and every memory key in~$\mtensor{K}^m$ are computed through
\begin{equation} \label{Affinity_Matrix} \mtensor{F} = (\mtensor{K}^m)\tran \mtensor{k}^q, \end{equation}
where~$\mtensor{K}^m \in \mathbb{R}^{C_k \times NHW}$ denotes the matrix obtained from concatenating every~$\mtensor{k}^m_n \in \mset{M}^k$ along the column dimension, such that~$\mtensor{K}^m = \left[ \mtensor{k}^m_{n=1},\mtensor{k}^m_{n=2},\dots,\mtensor{k}^m_{n=N}\right]$.
To reduce noise (\ie, low affinities) in the affinity matrix~$\mtensor{F}$, a top-$k$ filter is applied along the columns (memory dimension) following~\cite{MiVOS}. 
This produces a sparse matrix~$\mtensor{S} \in \mathbb{R}^{NHW\times HW}$ by 
\begin{equation}
	\label{EQ_Sparse_Matrix}
	\mtensor{S}_{i,j} =
	\begin{cases} 		
		\mtensor{F}_{i,j} & \text{, if } \mtensor{F}_{i,j} \in \underset{\mset{F}_j \subset \left\lbrace  f \mid f \in \mtensor{F}_{\ast,j}\right\rbrace , \lvert \mset{F}_j \rvert = k} \argmax \left( \displaystyle\sum_{f\in \mset{F}_j} f \right)
		\\
		0 & \text{, otherwise}
	\end{cases},
\end{equation}
where $i$ and $j$ denote the row and column indices respectively, and $\ast$ refers to either the complete row or column of a matrix.
Based on the sparse affinity matrix~$\mtensor{S}$, the method computes soft-weights~$\mtensor{W} \in \mathbb{R}^{NHW\times HW}$ using
\begin{equation}
	\label{EQ_Softmax_on_affinity}
	\mtensor{W}_{i,j} =  
	\frac{\exp(\mtensor{S}_{i,j})}
	{
		\sum^{NHW}_{l=1} \exp(\mtensor{S}_{l,j})
	}.
\end{equation} 

All the channels of every memory value feature~$\mtensor{v}^m_n \in \mset{M}^v$ are weighted with~$\mtensor{W}$ to compute a pseudo memory feature representation~$\widetilde{\mtensor{v}}^m \in \mathbb{R}^{C_v \times HW}$ through $\widetilde{\mtensor{v}}^m = \mtensor{V}^m \mtensor{W}$, where $\mtensor{V}^m \in \mathbb{R}^{C_k \times NHW}$ is the concatenated form of the memory values~$\mtensor{v}^m_n$ along the columns.

(5) Lastly, a decoder inspired by~\cSTM that predicts a segmentation mask~$\mtensor{M}^q$ for the query frame~$\mtensor{I}^q$ using refinement modules~\cite{oh2018fast}, skip-connections and bilinear upsampling. 

We use the original weights for the sVOS baselines in our experiments~(\ie, \MiVOS, \STCN and \QDMN).
The training of the sVOS baselines uses a bootstrapped cross-entropy loss~\cite{MiVOS} and consists of two stages: a \textit{pre-training} using static image datasets, and a \textit{main-training} using DAVIS~\cite{Pont-Tuset_arXiv_2017} and Youtube-VOS~\cite{xu2018youtube}. 
An intermediate stage can also be performed using the BL30K~\cite{MiVOS} dataset.

\subsection{READMem}

\textbf{Diversification of Memory Embeddings (DME)}: 
As mentioned, contemporary methods insert the embeddings of every $t$-th frame into the memory, along with the corresponding mask.
This leads to an inevitable memory overload when accumulating embeddings of unconstrained sequences.
In contrast, READMem proposes an alternative frame insertion scheme, maximizing the diversity of latent representations stored in the memory by only adding frames that enhance information. This allows us to limit the number of memory slots without degrading the performance and avoid GPU memory saturation when processing long videos.

In pursuit of enhancing the diversity of the memory key embeddings~$\mset{M}^k = \{\mathbf{k}_{n}^{m}\}_{n=1}^{N}$, we require a mean to quantify this diversity.
If we conceptually consider the embeddings to form a parallelotope, where the key embeddings~$\{\mathbf{k}_{n}^{m}\}_{n=1}^{N}$ act as edges, we can quantify the diversity by calculating the volume of this geometric object. 
A conventional approach to estimate the volume involves concatenating the keys $\mathbf{k}_{n}^{m} \in \mathbb{R}^{C_{k}HW}$ (we take a flattened representation of the keys $\mathbf{k}_{n}^{m} \in \mathbb{R}^{C_{k} \times HW}$) into a matrix $\mathbf{X} \in \mathbb{R}^{C_{k}HW \times N}$ and inferring the determinant of $\mathbf{X}$.
However, since $N \ll C_k \times HW$, $\mathbf{X}$ is a non-square matrix, which prevents the computation of the determinant. 
To circumvent this issue, we leverage the Gram matrix~$\mtensor{G} \in \mathbb{R}^{N \times N}$, computed through $\mtensor{G} = \mtensor{X}\tran \mtensor{X}$.
As demonstrated by~\cite{vinberg2003course} (pages $195$-$196$) the determinant of the Gram matrix (\ie, Gramian~\cite{vinberg2003course}) allows us to compute the square volume ($vol$) of the parallelotope, such that~$(vol P(\{\mtensor{k}_{n}^{m}\}_{n=1}^{N}))^2 = \det(\mtensor{G})$. 
Therefore, we can quantify the diversity of our memory with $\det(\mtensor{G})$.
We use the set of memory keys~$\mset{M}^k = \{\mtensor{k}_{n}^{m}\}_{n=1}^{N}$ to compute the diversity of the memory, as they encode visual semantics that are robust to appearance variation~\cSTM.
Hence, by computing the similarity~$s_{a,b}$ between the $a$-th and $b$-th flatten memory key in~$\mset{M}^k$ through an inner product $g : \mathbb{R}^{C_kHW} \times \mathbb{R}^{C_kHW} \to \mathbb{R}$ with $g(\mtensor{k}_a^m,\mtensor{k}_b^m) \coloneqq s_{a,b}$, we construct~$\mtensor{G}$ by
\begin{equation}
	\label{eq_gram_matrix}
	\mtensor{G}(\mset{M}^k)
	=
	\begin{pmatrix}
		s_{1,1} & s_{1,2} & \cdots & s_{1,N} \\
		\vdots  & \vdots  & \ddots & \vdots  \\
		s_{N,1} & s_{N,2} & \cdots & s_{N,N} 
	\end{pmatrix}.
\end{equation}
The size of the memory $N$ is bounded by the dimension of the feature space since otherwise $\det{(\mtensor{G})}=0$~\cite{barth1999gramian}. 
However, this is not a relevant limitation since in practice, $N$ is several magnitudes smaller than the dimension of the feature space $C_k \times HW$ (\ie, $N \ll C_k \times HW$).

To increase the diversity of the embeddings stored in the memory, we want to maximize the absolute Gramian~$\lvert\det{(\mtensor{G})}\rvert$ of $\mtensor{G}(\mset{M}^k)$.
As the annotated frame~$\mtensor{I}_1$ provides accurate and controlled segmentation information~\cite{STM,QDMN}, the corresponding memory key~$\mtensor{k}_{n=1}^m$ and value~$\mtensor{v}_{n=1}^m$ are exempt from the following update strategy:

(1)~For each memory slot $n \in \{2,\dots,N\}$, we substitute the respective memory key~$\mtensor{k}_n^m$ with the query-memory key~$\mtensor{k}^{q,m}$ (memory key of the current query frame~$\mtensor{I}^q$ and mask~$\mtensor{M}^q$) and construct a temporary Gram matrix~$\mtensor{G}_n^t$.
Using the temporary matrices~$\mtensor{G}_n^t$, we build a set~$\mset{G}^t$ containing the absolute values of the determinants, such that~$\mset{G}^t = \big\{ \lvert\det{(\mtensor{G}_n^t)}\rvert \big\}_{n=2}^{N}$. 

(2) The memory is updated based on whether the highest value in $\mset{G}^t$ surpasses the absolute value of the current Gramian $\lvert\det{(\mtensor{G})}\rvert$. 
If this condition is met, then the corresponding memory slots~$n$ (position of the highest value in~$\mset{G}^t$) is replaced with the query-memory key~$\mtensor{k}^{q,m}$ and value~$\mtensor{v}^{q,m}$ embeddings.

\begin{figure*}[t]
	\centering
	\includegraphics[width=1\textwidth, trim={0mm 3.cm 0mm 0.3cm},clip]{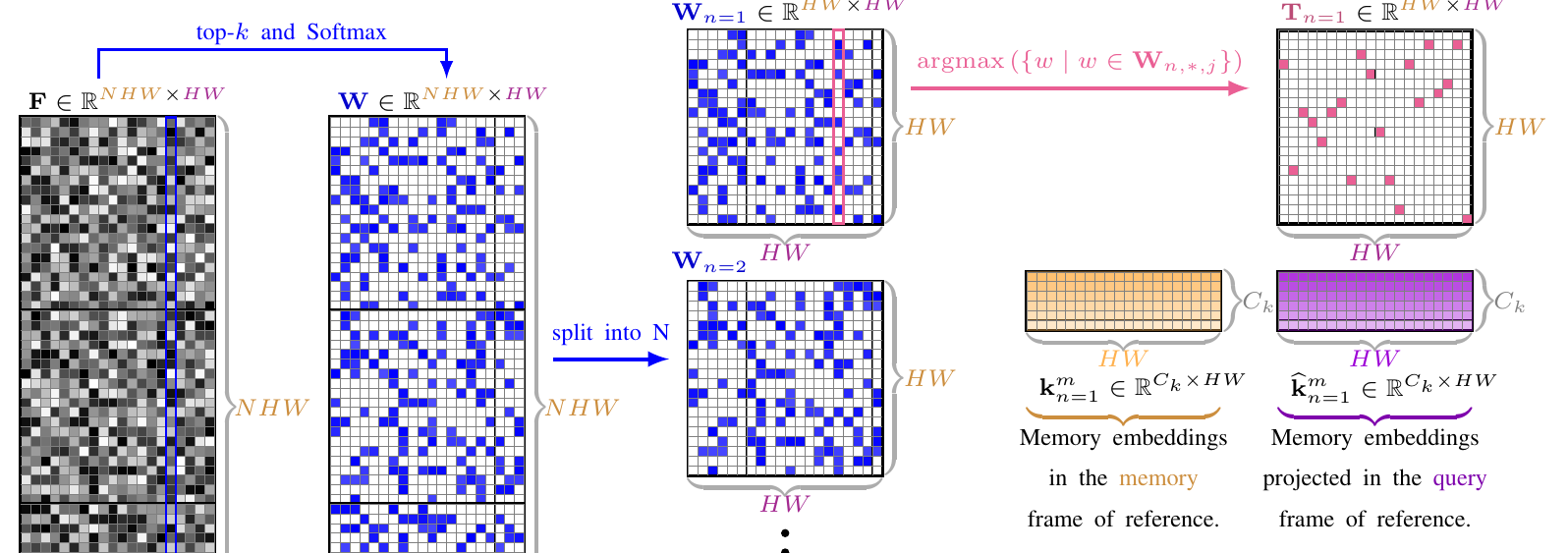}
	\vspace{-2mm}
	\caption{Inference of a transition matrix~$\mtensor{T}_n$ and the corresponding pseudo-memory embeddings $\widehat{\mtensor{k}}_n^m$ for robust embeddings association (REA).}
	\label{fig:OverViewxTransformation}
\end{figure*}

\textbf{Robust Embeddings Association}~(\textbf{REA}): 
Given that we compute an inner-product between two embeddings (enforced by the Gram matrix), we capture the global similarity of two frames in a single score. 
However, as two frames,~$\mtensor{I}_t$ and $\mtensor{I}_{t+\Delta t}$ are further apart in time, it becomes more likely for the object of interest to experience in-between: motion, ego-motion, or size variations.
Consequently, a channel-wise embedding may encode different information (specifically foreground \vs background) for the same image region in frame $\mtensor{I}_t$ and frame $\mtensor{I}_{t+\Delta t}$ due to the spatial disparity in the object's location. 
As a result, the resulting similarity between the two frames is inherently low when comparing their embeddings.
To address this issue and dampen positional variation without relying on a motion model or fixed-sized search region (which would introduce additional hyper-parameters), our proposal utilizes transition matrices $\{\mtensor{T}_{n}\}_{n=1}^N$.
Wherein, the core idea is to leverage the cross-attention already at hand in the memory-based networks (\ie, $\mtensor{W}$ and $\mtensor{F}$) to identify the best channel-wise mapping between two embeddings.
Specifically, through the transition matrix~$\mtensor{T}_n \in \mathbb{R}^{HW \times HW}$, we project the $n$-th memory embedding~$\mtensor{k}_n^m \in \mathbb{R}^{C_v \times HW}$ from its original \ac{for} to the query's \ac{for} (see \figref{fig:OverViewxTransformation}) by
\begin{equation}
	\label{eq_pseudo_memory_key}
	\widehat{\mtensor{k}}^m_n
	= \mtensor{k}^m_n \mtensor{T}_n
	,
\end{equation}
where $\widehat{\mtensor{k}}^m_n \in \mathbb{R}^{C_v \times HW}$ denotes the pseudo-embedding of the embedding~$\mtensor{k}_n^m$, but expressed in the query's \textit{frame of reference}. 
This approach effectively compensates for the target's spatial disparity between two distant frames (refer to \tabref{tab_TableAblations_tres} and \ref{TableTransformationAblation}).

Thus, we compute the similarity \wrt~$\mtensor{k}^{q,m}$ not with the memory keys $\{\mtensor{k}^m_n\}_{n=1}^{N}$ but with the pseudo memory keys~$\{\widehat{\mtensor{k}}^m_n\}_{n=1}^{N}$.
\figref{fig:OverViewxTransformation} depicts the operation that maps key embeddings from the memory's \ac{for} to the query's \ac{for} through the transition matrix~$\mtensor{T}_n$.
We use $\mtensor{W}$, a filtered version of $\mtensor{F}$~(see~\eqref{EQ_Softmax_on_affinity}), to only consider strong affinities between the point-wise embeddings (channels) of the query and memory key which are more likely to encode a similar information.
This ensures that the corresponding mapping matrix~$\mtensor{T}_n$ is constructed based on relevant channels~(refer to~\tabref{TableTransformationAblation}).
Although $\mtensor{W}$ entails strong similarities, using~$\mtensor{W}_n$ as transition matrix~$\mtensor{T}_n$ potentially leads to the aggregation of multiple memory channels onto a single pseudo memory channel~\textendash~which is certain to degrade the similarity.
Hence, to avoid the summation, we need to map at most one element from the memory \ac{for} to the query \ac{for},~\ie, filtering $\mtensor{W}_n$.
Generating a bijective mapping between the memory key~$\mtensor{k}_n^m$ and the corresponding pseudo query key~$\widehat{\mtensor{k}}_n^m$ would constraint the foreground-background ratio of $\widehat{\mtensor{k}}_n^m$ based on~$\mtensor{k}_n^m$.
However, it is unlikely that the area taken by the object in the image is unchanged from one frame to the other.
To avoid this foreground-background restriction, we should allow point-wise embeddings from the memory key~$\mtensor{k}_n^m$ to be re-used for the creation of the pseudo memory key~$\widehat{\mtensor{k}}_n^m$, as long as they are sufficiently similar.
Thus, an appropriate function that validates the aforementioned criteria is $\argmax$ applied along the columns of~$\mtensor{W}$ to maximize the re-usability of point-wise embeddings when producing the pseudo memory key~$\widehat{\mtensor{k}}_n^m$ on the query \ac{for}.
We obtain $\mtensor{T}_n \in \mathbb{R}^{HW \times HW}$ 
by dividing~$\mtensor{W} \in \mathbb{R}^{NHW \times HW}$ into~$N$-square matrices, such that~$\mtensor{W}_n \in \mathbb{R}^{HW \times HW}$, and by applying $\argmax$ along the columns by 

\begin{equation}
	\label{eq_permutation_base_on_affinity}
	\mtensor{T}_{n,i,j}
	= 
	\begin{cases} 
			1& \text{, if } i \in \argmax\left( \left\lbrace w \mid w \in \mtensor{W}_{n,\ast,j} \right\rbrace  \right) 
			\\
			0& \text{, otherwise}
			\\
		\end{cases}
	.
\end{equation}

\textbf{Lower Similarity Bound of Memory Embeddings}~(\textbf{LSB}): To ensure the presence of the foreground object in a candidate frame, we set a lower bound on similarity $l_{bs} = 0.5$.
We validate object presence in~$\mtensor{I}_q$ by computing a similarity score such that ~$g({\widehat{\mtensor{k}}^m_1,\mtensor{k}^{q,m}}) > l_{bs}$.

\textbf{Initialization}: We integrate every~$t$-th frame (that satisfies the lower similarity bound) into the memory~\textendash~until the memory is completely filled.
Afterward, the method only integrates embeddings of frames that enhance the diversity of the memory.

%% file: sections/results.tex
\section{Experiments}
\label{sec:experiments}
In our experiments, we use the~\LongTimeVideo (LV1) dataset (long videos) as well as the validation set of the~\DAVIS17 (D17) dataset (short videos).
We use the~\J~(intersection over union) and~\F~(contour accuracy) metrics (higher is better) introduced by~\cite{Pont-Tuset_arXiv_2017}\xspace to quantify the performance of each method.
Both metrics are averaged into the~\JandF metric for ease of ranking.
We perform all experiments on an Nvidia GeForce GTX 1080 Ti.

\textbf{Quantitative Results}:
\label{sec:quantitativexresults}
\tabref{tab_TableQuantitativeResults} presents the quantitative results of recent sVOS methods (\MiVOS, \STCN and \QDMN) and their READMem based extensions along with the state-of-the-art sVOS method \XMEM on the \LV1 and the \D17 datasets~\textendash~we display the qualitative results of \MiVOS and READMem-MiVOS on \LV1 in~\figref{fig:LVONE_MiVOS_Holi_MiVOS_All_in_One}.

To ensure a fair comparison, we keep the default settings for \XMEM and use the same sampling interval (\ie, $s_r = 10$) for all other methods (in contrast to previous works~\cite{XMEM,AFB-URR} that use dataset-specific sampling intervals).
In our experiments, we follow the evaluation of~\cite{QDMN} and limit the number of memory slots of the baselines and the corresponding READMem counterparts to $20$. 
Note that this value differs from the memory slot limit (\ie, $50$) used by~\cite{XMEM,AFB-URR}.
In contrast to~\cite{XMEM, AFB-URR}, we do not adapt the sampling interval to the sequence length when dealing with long videos as we argue that the sampling interval should not be correlated to the video's length and is not a standard practice when dealing with short videos (\D17 and Y-VOS~\cite{xu2018youtube}). 
Instead, we opt for a first-in-first-out~(\ie, FIFO) replacement strategy when updating the memory of the mentioned baselines.
Our results in \tabref{tab_TableQuantitativeResults} demonstrate that READMem improves long-term performance while preserving good short-term capabilities. 
Moreover, by setting $s_r=1$ we minimize the likelihood of omitting key frames and enhance the stability of the method~\cite{XMEM}.
As a result, we attain competitive results against the state-of-the-art method~(\ie,~\XMEM).
We do not employ READMem on \XMEM, as a long-term memory is already present, consolidating the embeddings of the working memory~\cite{STCN} and updated by a \textit{least-frequently-used} (\textit{LFU}) approach~\cite{XMEM,AFB-URR}.

\textbf{Ablation Studies:}
For clarity and to isolate the benefits of each design decision, we present our ablation results solely for READMem-MiVOS in \tabref{tab_TableAblations_tres}. 
We observe that integrating embeddings of frames that diversify the memory (DME) and enforcing a lower bound on similarity (LSB) results in a significant performance improvement for long videos. 
Moreover, using the robust association of memory embeddings (REA) also leads to a substantial performance increase.
Although including the adjacent frame leads to a slight improvement for long videos, it is particularly important when dealing with short sequences as the previous adjacent frame provides recent information (not guaranteed by our module).

\begin{table}[t]
	\centering
	\resizebox{0.9\columnwidth}{!}{
		\begin{tabular}{l|ccc|ccc}
			\hline
			&\multicolumn{3}{c}{\textbf{\LV1 (Long Sequences)}}&\multicolumn{3}{c}{\textbf{\D17 (Short Sequences)}}\\
			\textbf{Method} & \JandF & \J & \F & \JandF & \J & \F \\
			\hline
			\rowcolor{black!10!white}
			\multicolumn{7}{c}{\textit{Sampling Interval $s_r = 10$}}\\
			\hline
			XMem{$^\dagger$}~\cite{XMEM}~\refConf{ECCV 22}	& $\mathbf{89.8}$ & $\mathbf{88.0}$ & $\mathbf{91.6}$ & $85.5$ & $82.3$ & $88.6$ \\
			\XMEM~\refConf{ECCV 22} 			& $83.6$ & $81.8$ & $85.4$ & $\mathbf{86.7}$ & $83.0$ & $\mathbf{90.1}$ \\
			\hline
			\MiVOS~\refConf{CVPR 21} 				& $64.3$ & $63.6$ & $65.0$ & $84.3$ & $81.4$ & $87.1$ \\
			{\color{IBMINDIGO!80!black}READMem-MiVOS}~\refConf{ours}	& $83.6^{\uparrow19.3}$ & $82.1^{\uparrow18.5}$ & $85.1^{\uparrow20.1}$ & $84.3$ & $81.4$ & $87.1$ \\
			{\color{IBMINDIGO!80!black}READMem-MiVOS}{$^\star$}~\refConf{ours}
			& $\underline{86.0}^{\uparrow21.7}$ & $\underline{84.6}^{\uparrow21.0}$ & $\underline{87.4}^{\uparrow22.4}$ & $84.6^{\uparrow0.3}$ & $81.8^{\uparrow0.4}$ & $87.3^{\uparrow0.2}$ \\
			\hline
			STCN{$^\dagger$}~\cite{STCN}~\refConf{NIPS 21}   				& $71.8$ & $69.4$ & $74.3$ & $83.7$ & $80.7$ & $86.6$ \\
			{\color{IBMINDIGO!80!black}READMem-STCN}{$^\dagger$}~\refConf{ours}	& $82.6^{\uparrow10.8}$ & $81.5^{\uparrow12.1}$ & $83.7^{\uparrow9.4}$ & $83.7$ & $80.7$ & $86.6$ \\
			{\color{IBMINDIGO!80!black}READMem-STCN}{$^{\dagger\star}$}~\refConf{ours}
			& $81.8^{\uparrow10.0}$ & $80.8^{\uparrow11.4}$ & $82.8^{\uparrow8.5}$ & $83.7$ & $80.7$ & $86.6$ \\
			\hline
			QDMN{$^\star$}~\cite{QDMN}~\refConf{ECCV 22}					& $80.7$ & $77.8$ & $83.6$ & $86.0$ & $\underline{83.2}$ & $88.8$ \\
			{\color{IBMINDIGO!80!black}READMem-QDMN}{$^\star$}~\refConf{ours}	& $84.0^{\uparrow3.3}$ & $81.3^{\uparrow3.5}$ & $86.7^{\uparrow3.1}$ & $\underline{86.1}^{\uparrow0.1}$ & $\mathbf{83.3}^{\uparrow0.1}$ & $\underline{88.9}^{\uparrow0.1}$ \\
			\hline
			\rowcolor{black!10!white}
			\multicolumn{7}{c}{\textit{Sampling Interval $s_r = 1$}}\\
			\hline
			XMem{$^\dagger$}~\cite{XMEM}~\refConf{ECCV 22}	& $61.5$ & $60.4$ & $68.4$ & $84.1$ & $81.1$ & $\mathbf{90.8}$ \\
			\XMEM~\refConf{ECCV 22} 			& $79.3$ & $77.3$ & $84.2$ & $83.9$ & $80.6$ & $87.1$ \\
			\hline
			\MiVOS~\refConf{CVPR 21} 				& $27.4$ & $27.4$ & $27.4$ & $84.3$ & $81.3$ & $87.3$ \\
			{\color{IBMINDIGO!80!black}READMem-MiVOS}~\refConf{ours}	& $\underline{83.6}^{\uparrow56.2}$ & $\mathbf{82.3}^{\uparrow54.9}$ & $\underline{85.0}^{\uparrow57.6}$ & $84.3$ & $81.4^{\uparrow0.1}$ & $87.1^{\downarrow0.2}$ \\
			\hline
			\STCN{$^\dagger$}~\refConf{NIPS 21} 					& $26.2$ & $22.9$ & $29.4$ & $83.2$ & $79.9$ & $86.5$ \\
			{\color{IBMINDIGO!80!black}READMem-STCN}{$^\dagger$}~\refConf{ours}	& $80.8^{\uparrow54.6}$ & $78.4^{\uparrow55.5}$ & $83.2^{\uparrow53.8}$ & $83.8^{\uparrow0.6}$ & $80.4^{\uparrow0.5}$ & $87.2^{\uparrow0.7}$ \\
			\hline
			\QDMN~\refConf{ECCV 22} 					& $65.7$ & $63.5$ & $67.9$ & $\underline{85.1}$ & $\underline{82.2}$ & $88.0$ \\
			{\color{IBMINDIGO!80!black}READMem-QDMN}~\refConf{ours}	& $\mathbf{84.3}^{\uparrow18.6}$ & $\underline{81.9}^{\uparrow18.4}$ & $\mathbf{86.7}^{\uparrow18.8}$ & $\mathbf{85.3}^{\uparrow0.2}$ & $\mathbf{82.4}^{\uparrow0.2}$ & $\underline{88.1}^{\uparrow0.1}$ \\
			\hline
		\end{tabular}
	}
	\caption{Quantitative evaluation of sVOS methods~\cite{MiVOS,STCN,QDMN, XMEM} with and without READMem on the~\LV1 and~\D17 datasets. The symbol $^\dagger$ denotes no pre-training on BL30K~\cite{MiVOS}, while $^\star$ indicates the use of a flexible sampling interval as in~\QDMN.
	}
	\label{tab_TableQuantitativeResults}
	\end{table}
	
\vspace{-5mm}

\begin{table}[H]
	\begin{center}
		\resizebox{0.8\columnwidth}{!}{
			\begin{tabular}{l|cc}
				\hline
				\textbf{Configuration} & \JandFLV1 & \JandFD17 \\
				\hline
				\MiVOS + adj. frame~\refConf{baseline} 					& $64.3$ 				 & $84.3$ \\
				\hline
				\MiVOS + DME~\refConf{ours}   								& $69.5^{\uparrow 5.2}$  & $81.3^{\downarrow 3.0}$ \\
				\MiVOS + DME + LSB~\refConf{ours} 						& $75.0^{\uparrow 10.7}$ & $81.3^{\downarrow 3.0}$ \\
				\MiVOS + DME + LSB + adj. frame ~\refConf{ours} 			& $77.4^{\uparrow 13.1}$ & $84.3^{\uparrow 0.0}$ \\
				\MiVOS + DME + LSB + adj. frame + REA~\refConf{ours} 	& $\mathbf{86.0}^{\uparrow 21.7}$ & $\mathbf{84.6}^{\uparrow 0.3}$ \\
				\hline
			\end{tabular}
		}
		\caption{Demonstrating the benefits of our memory extension, \ie, diversity of the embeddings (DME), robust embedding association (REA) and lower similarity bound (LSB).
		}
		\label{tab_TableAblations_tres}
	\end{center}
\end{table}

\vspace{-5mm}
\begin{table}[H]
	\begin{center}
		\resizebox{0.9\columnwidth}{!}{
			\begin{tabular}{c|cc}
				\hline
				\textbf{Computation of the transition matrix~$\mtensor{T}_n$} & \JandFLV1 using~$\mtensor{W}_n$ & \JandFLV1 using~$\mtensor{F}_n$ \\
				\hline
				Hungarian Method~\cite{kuhn1955hungarian} & $76.8$ & $75.1$ \\
				$\argmax$ along the columns & $\mathbf{86.0}$ & $73.4$ \\
				$\argmax$ along the rows    & $79.8$ & $\mathbf{77.1}$\\
				\hline
			\end{tabular}
		}
		\caption{Performance comparison on the~\LV1 dataset when computing the transition matrix~$\mtensor{T}_n$ through different methods. As a reminder, the baseline~(\ie, \MiVOS) achieves a \JandF score of $64.3$ on~\LV1 using the same configuration.}
		\label{TableTransformationAblation}
	\end{center}
\end{table}

\vspace{-5mm}

\begin{table}[H]
	\centering
		\resizebox{0.7\columnwidth}{!}{
			\begin{tabular}{c|cc|cc|cc}
				\hline
				& \multicolumn{2}{c}{\textbf{\textit{blueboy}}} & \multicolumn{2}{c}{\textbf{\textit{dressage}}} &        \multicolumn{2}{c}{\textbf{\textit{rat}}} \\
				$\mathbf{s_r}$ &  \JandF  &  $\lvert \mtensor{G} \rvert$   & \JandF &    $\lvert \mtensor{G} \rvert$    & \JandF &        $\lvert \mtensor{G} \rvert$ \\ 
				\hline
				$1$  & $\mathbf{89.5}$ &     $\mathbf{14.6 \times 10^{-7}}$     &  $83.7$   &     $3.36\times 10^{-8}$      &  $77.7$   & $4.22\times 10^{-10}$ \\ 
			$5$ & $87.0$ & $6.35\times 10^{-7}$ & $83.7$ & $9.38\times 10^{-8}$ & $74.9$ & $1.39\times 10^{-10}$ \\ 
			$10$ & $88.1$ & $4.31\times 10^{-7}$ & $\mathbf{84.0}$ & $\mathbf{18.0\times 10^{-8}}$ & $\mathbf{86.0}$ & $\mathbf{82.3\times 10^{-10}}$ \\ 
			\hline
		\end{tabular}
	}
	\caption{Performance and Gramian of READMEm-MiVOS for different sampling intervals $s_r$ on \LV1. Note that high \JandF scores correlate with high Gramian values.}
\label{Tab:Gramian_over_time_for_blueboy_dressage_n_rat}
\end{table}

In~\tabref{TableTransformationAblation}, we compare different approaches for inferring the transition matrix~$\mtensor{T}_n$. 
Our results indicate, as expected, that using the weight matrix $\mtensor{W}$ to generate the transition matrices~$\mtensor{T}_n$ leads to better performance compared to the affinity matrix $\mtensor{F}$.
Furthermore, we note that using the~$\argmax$ function along the columns axis~(memory) leads to the best performance in comparison to applying~$\argmax$ along the rows axis~(query) or when using a bijective mapping (Hungarian Method~\cite{kuhn1955hungarian}).

\begin{figure}[t]
	\centering
	\begin{minipage}{0.3\textwidth}
		\includegraphics[trim={0mm 0 0 0mm},clip, width=\textwidth]{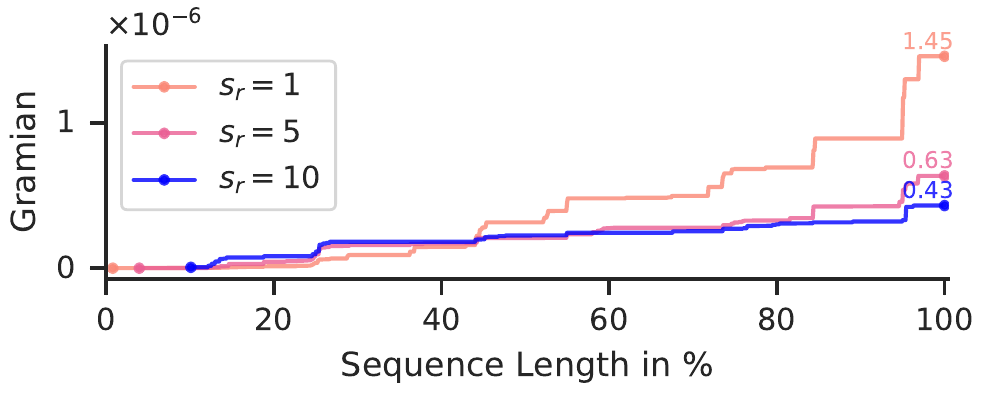}
		\caption{Evolution of~$\lvert \mtensor{G} \rvert$ for different sampling interval on \textit{blueboy}~\cite{AFB-URR}.}
		\label{fig:Gramian_over_time_for_blueboy_dressage_n_rat}
	\end{minipage}
	\hspace{0.3cm}
	\begin{minipage}{0.65\textwidth}
		\centering
		\resizebox{\columnwidth}{!}{
			\begin{tabular}{c|cc|cc|cc}
				\hline
				& \multicolumn{2}{c}{$0.0\% - 26.20\%$} 
				& \multicolumn{2}{c}{$26.20\% - 57.64\%$} 
				& \multicolumn{2}{c}{$57.64\% - 83.63\%$} 
				\\
				$s_r$ & \JandF & $\lvert \mtensor{G} \rvert$ & \JandF & $\lvert \mtensor{G} \rvert$ & \JandF & $\lvert \mtensor{G} \rvert$ \\ 
				\hline 
				1  & $\mathbf{94.3}$ & $0.61\times 10^{-7}$ 			 & $\mathbf{72.5}$ 	& $\mathbf{4.80 \times 10^{-7}}$ 	& $\mathbf{95.6}$ 	& $\mathbf{6.93\times 10^{-7}}$
				\\
				5  & $93.3$ 		 & $1.46\times 10^{-7}$ 			 & $59.3$ 			& $2.34\times 10^{-7}$ 			& $92.2$ 			& $3.45\times 10^{-7}$
				\\
				10 & $90.4$ 		 & $\mathbf{1.69\times 10^{-7}}$  & $66.1$ 			& $2.44\times 10^{-7}$ 			& $95.4$			& $3.09\times 10^{-7}$
				\\
				\hline
			\end{tabular}
		}
		\captionof{table}{Relative performance and Gramian (\ie,~$\lvert \mtensor{G} \rvert$) evolution for three sections on the \textit{blueboy} sequence~\cAFBURR. The final Gramian and \JandF score is reported in \tabref{Tab:Gramian_over_time_for_blueboy_dressage_n_rat}.
		}
		\label{Tab:Gramian_and_JF_metrix_for_boueboy_percentil}
	\end{minipage}
\end{figure}

\tabref{Tab:Gramian_over_time_for_blueboy_dressage_n_rat} tabulates the~\JandF score and Gramian of READMem-MiVOS on the three sequences of~\LV1 for three different sampling interval~$s_r$. 
In~\figref{fig:Gramian_over_time_for_blueboy_dressage_n_rat}, we display the evolution of the Gramian over the observed~\textit{blueboy} sequence along with the specific attained Gramian and the respective \JandF performance of three intermediate sections in~\tabref{Tab:Gramian_and_JF_metrix_for_boueboy_percentil}.
We note that the Gramian continuously increases over the observed sequences and 
provides more assistance in the latter stages as indicated in~\tabref{Tab:Gramian_and_JF_metrix_for_boueboy_percentil}, where a high Gramian generally correlates to higher performance.

%% file: sections/conclusion.tex
\section{Conclusion}
\label{sec:conclusion}
We propose READMem, a modular framework for STM-based sVOS methods, to improve the embeddings-stored in the memory on unconstrained videos and address the expanding memory demand.
Our evaluation displays that sVOS methods~\cite{MiVOS, STCN, QDMN} using our extension consistently improve their performance compared to their baseline on the long video sequences (\ie, \LV1) without compromising their efficiency on shorter sequences (\ie, \D17).
Moreover, we achieve competitive results with~\sota and provide a comprehensive overview of each design decision, supported by an extensive ablation study.

%% file: supplementary.tex
\newpage

\section*{\centering{\huge{Supplementary Material}}\\\vspace{2mm} \centering{READMem: Robust Embedding Association for a Diverse Memory in Unconstrained Video Object Segmentation}}




\definecolor{IBMINDIGO}{HTML}{0000FF}
\definecolor{IBMMAGENTA2}{HTML}{9D02D7}
\definecolor{IBMMAGENTA1}{HTML}{CD34B5}
\definecolor{IBMPINK}{HTML}{EA5F94}
\definecolor{IBMLIGHTORANGE}{HTML}{FA8775}
\definecolor{IBMORANGE}{HTML}{FFB14E}
\definecolor{IBMGOLD}{HTML}{FFD700}

\graphicspath{{./figures/}}

In this supplementary document, we provide additional experiments, visualizations and insights.

\section{Additional Qualitative Results on the Long-time Video (\LV1) Dataset}
We display qualitative results for the READMem variations of \MiVOS, \STCN and \QDMN along with their baseline on the \LV1 dataset.\blfootnote{\hspace{-5mm}$^\star$Fraunhofer IOSB is a member of the Fraunhofer Center for Machine Learning.}
We use the same settings as described in~\secref{sec:quantitativexresults}~(refer to quantitative results). 
We also provide the results for \XMEM, which represents the state-of-the-art. 

\begin{figure}[H]
	\centering
	\begin{tabular}{p{2.4cm}ccccc}
	\centering {\footnotesize \MiVOS \vs READMem-MiVOS}  &
	\bmvaHangBox{\includegraphics[width=0.135\textwidth]{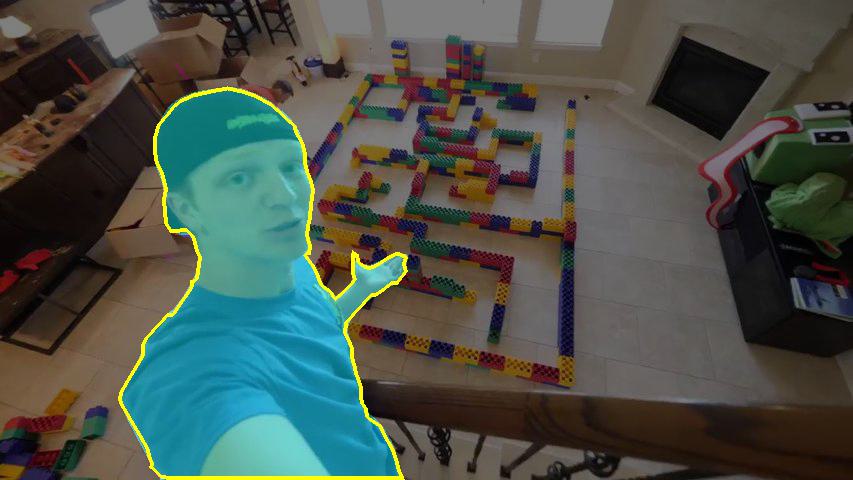}
		\put(-48.0,23.0){\makebox[0pt][l]{{\color{IBMGOLD}\textbf{\tiny \#4941}}}}}
	&\hspace{-4.5mm}
	\bmvaHangBox{\includegraphics[width=0.135\textwidth]{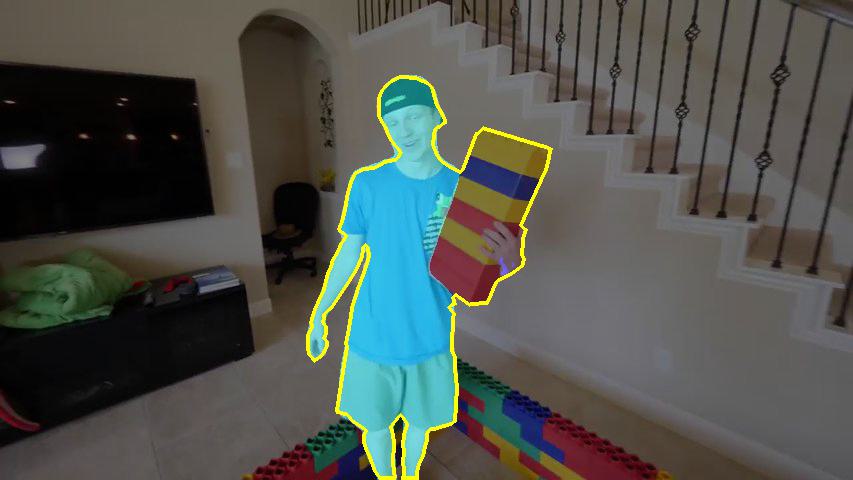}
		\put(-48.0,23.0){\makebox[0pt][l]{{\color{IBMGOLD}\textbf{\tiny \#6057}}}}}
	&\hspace{-4.5mm}
	\bmvaHangBox{\includegraphics[width=0.135\textwidth]{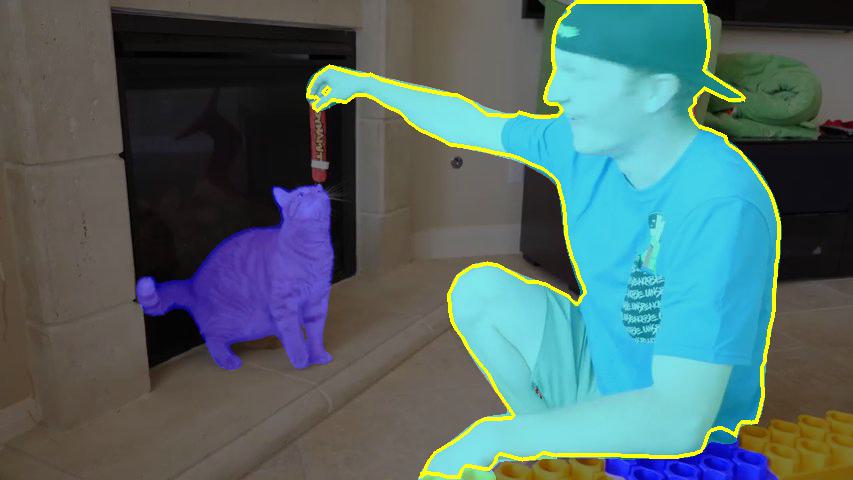}
		\put(-48.0,23.0){\makebox[0pt][l]{{\color{IBMGOLD}\textbf{\tiny \#11139}}}}}
	&\hspace{-4.5mm}
	\bmvaHangBox{\includegraphics[width=0.135\textwidth]{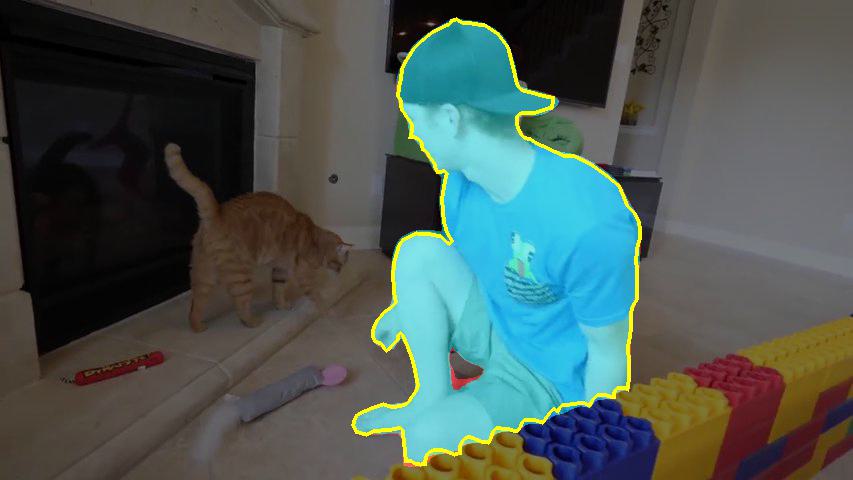}
		\put(-48.0,23.0){\makebox[0pt][l]{{\color{IBMGOLD}\textbf{\tiny \#11895}}}}}
	&\hspace{-4.5mm}
	\bmvaHangBox{\includegraphics[width=0.135\textwidth]{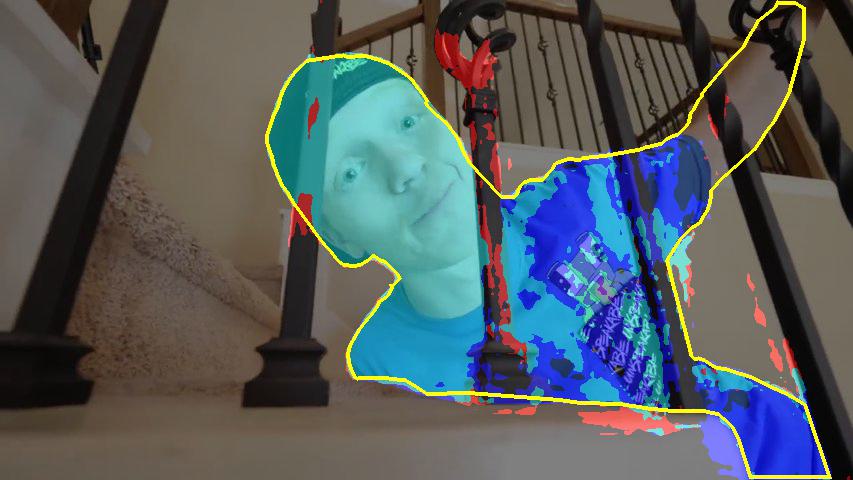}
		\put(-48.0,23.0){\makebox[0pt][l]{{\color{IBMGOLD}\textbf{\tiny \#18252}}}}}
	\\
		\centering {\footnotesize \STCN \vs READMem-STCN} &
		\bmvaHangBox{\includegraphics[width=0.135\textwidth]{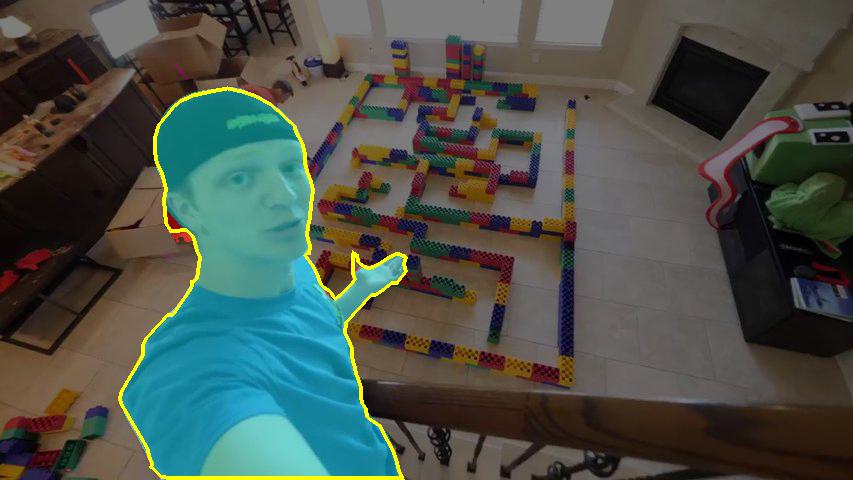}}
		&\hspace{-4.5mm}
		\bmvaHangBox{\includegraphics[width=0.135\textwidth]{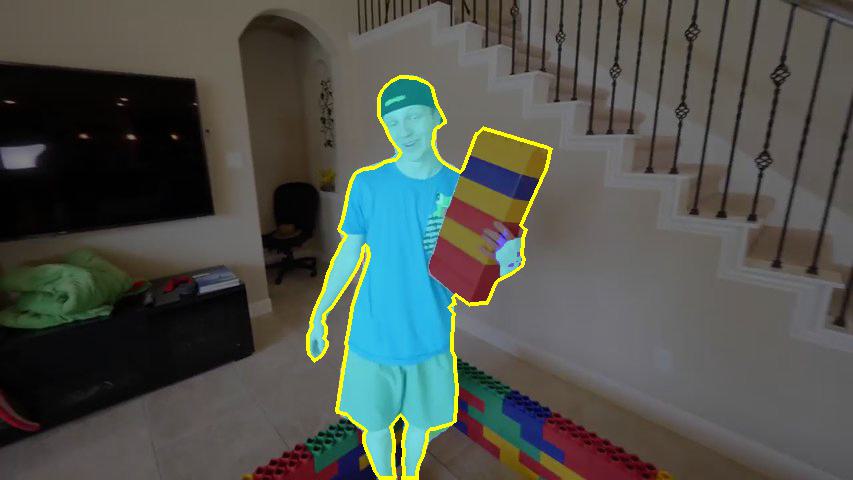}}
		&\hspace{-4.5mm}
		\bmvaHangBox{\includegraphics[width=0.135\textwidth]{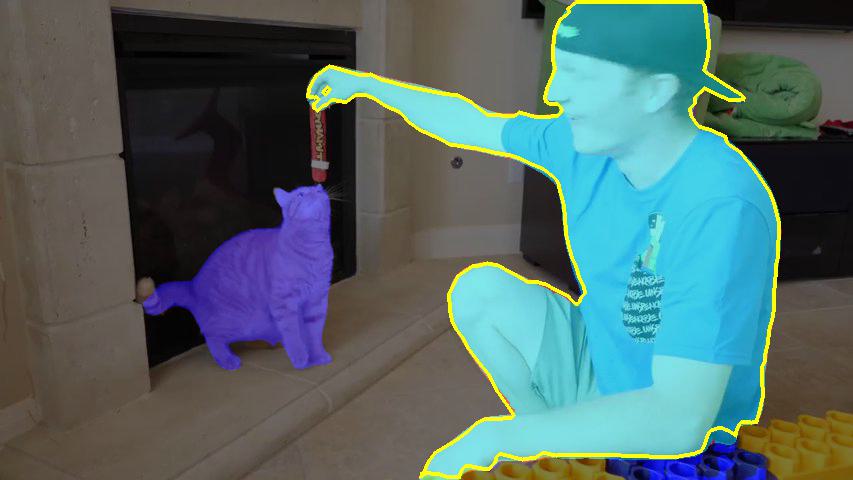}}
		&\hspace{-4.5mm}
		\bmvaHangBox{\includegraphics[width=0.135\textwidth]{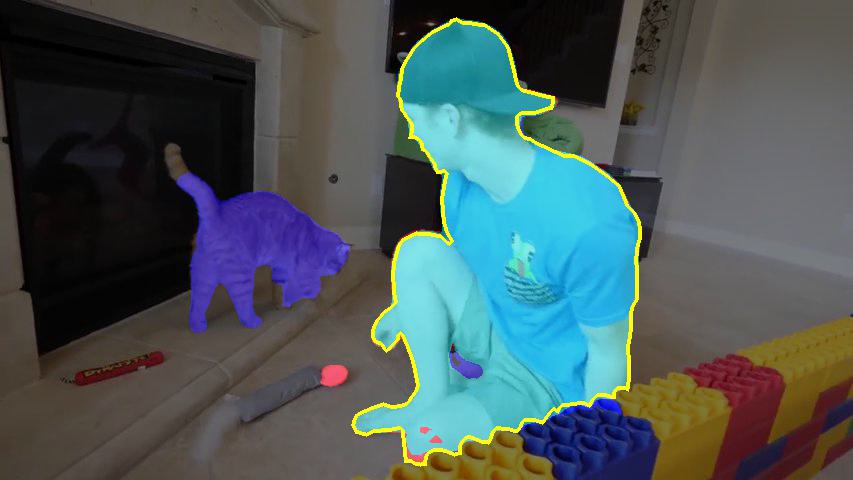}}
		&\hspace{-4.5mm}
		\bmvaHangBox{\includegraphics[width=0.135\textwidth]{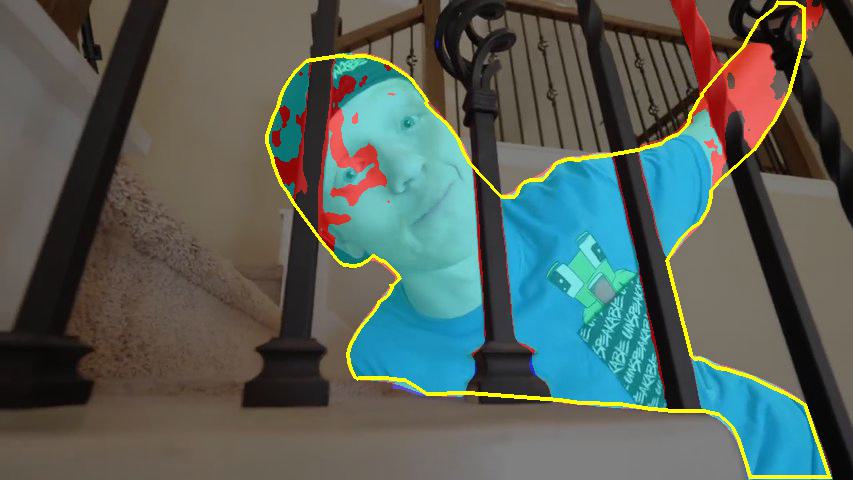}}
		\\
		\centering {\footnotesize \QDMN \vs READMem-QDMN} &
		\bmvaHangBox{\includegraphics[width=0.135\textwidth]{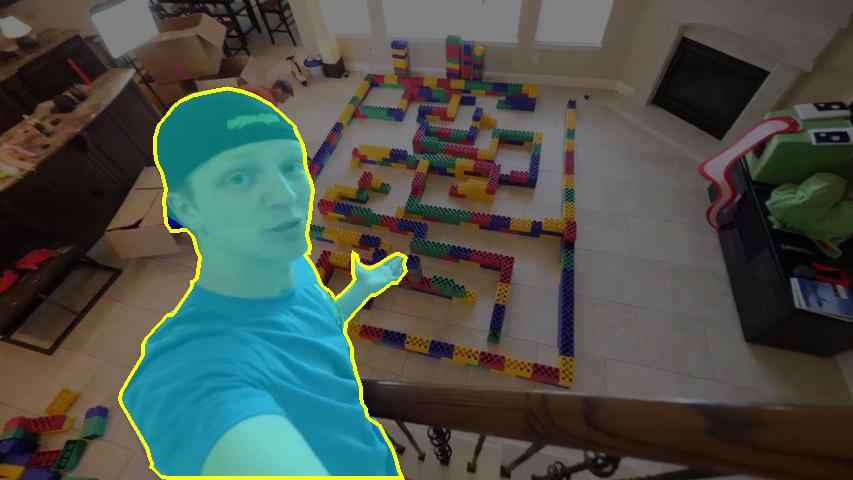}}
		&\hspace{-4.5mm}
		\bmvaHangBox{\includegraphics[width=0.135\textwidth]{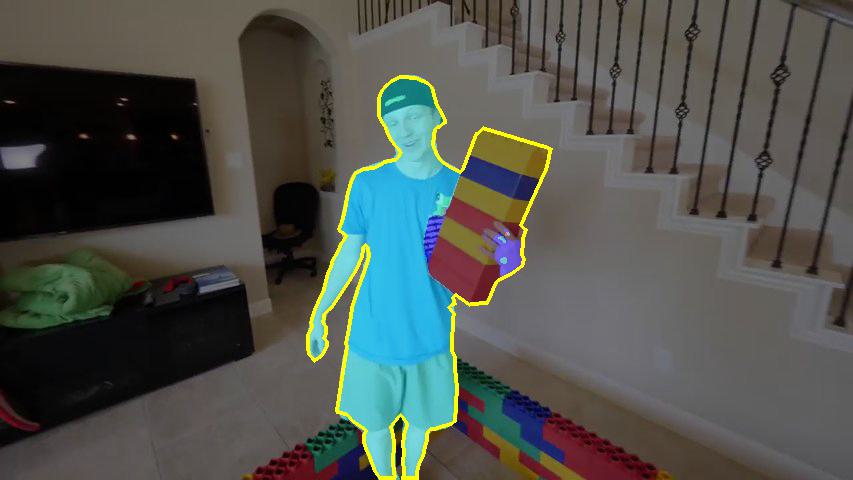}}
		&\hspace{-4.5mm}
		\bmvaHangBox{\includegraphics[width=0.135\textwidth]{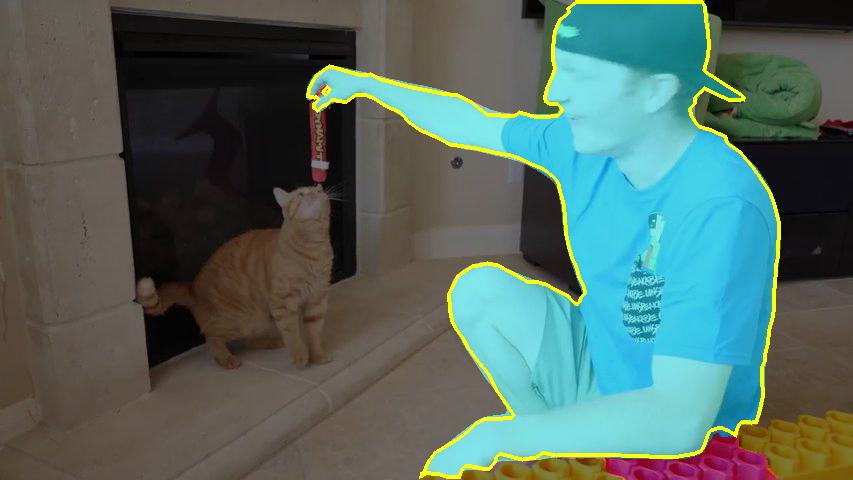}}
		&\hspace{-4.5mm}
		\bmvaHangBox{\includegraphics[width=0.135\textwidth]{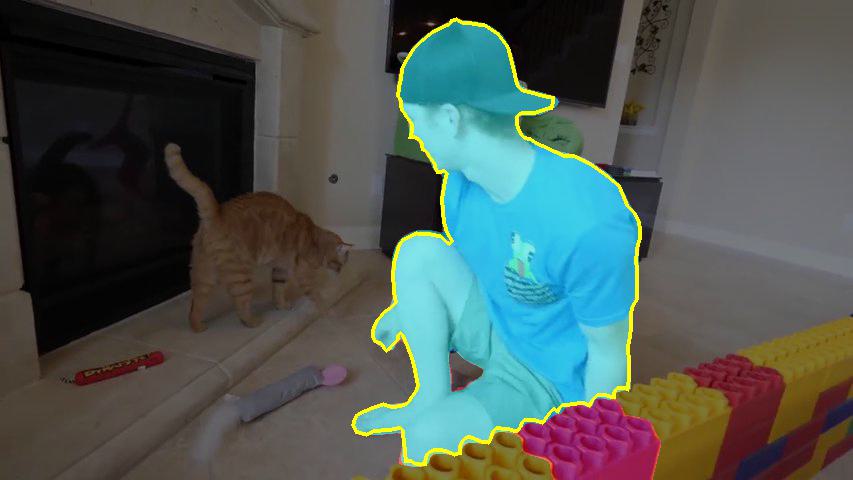}}
		&\hspace{-4.5mm}
		\bmvaHangBox{\includegraphics[width=0.135\textwidth]{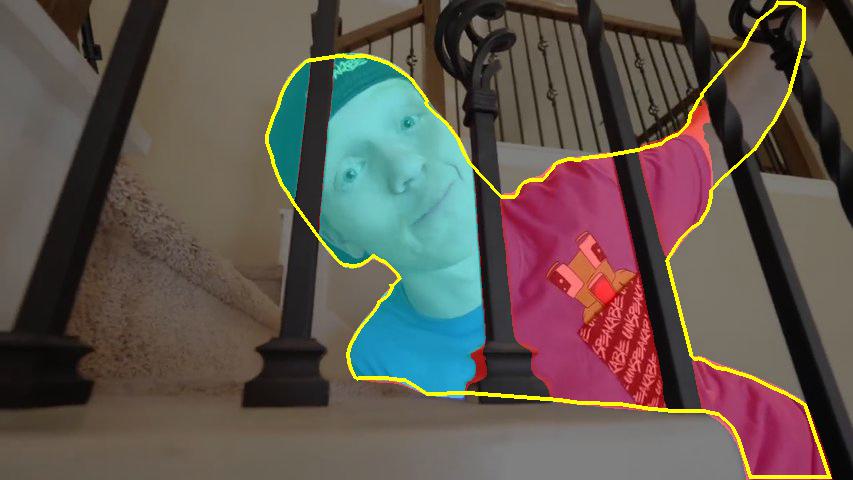}}
		\\
		\centering {\footnotesize \XMEM} &
		\bmvaHangBox{\includegraphics[width=0.135\textwidth]{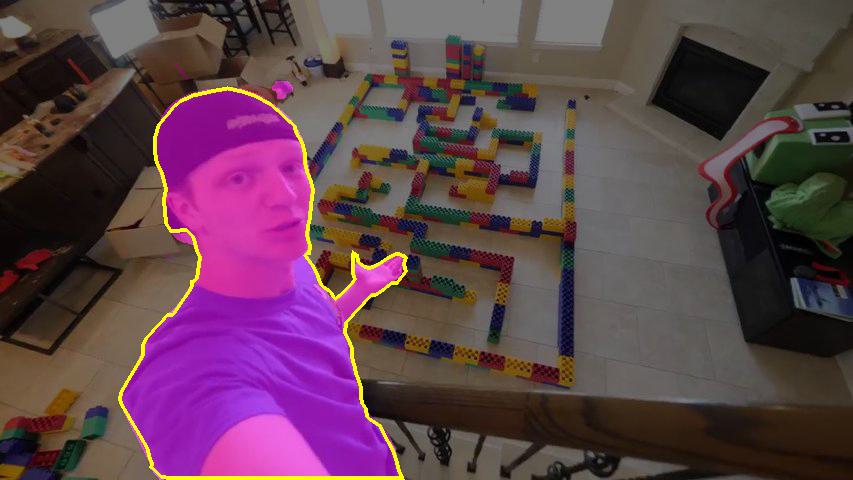}}
		&\hspace{-4.5mm}
		\bmvaHangBox{\includegraphics[width=0.135\textwidth]{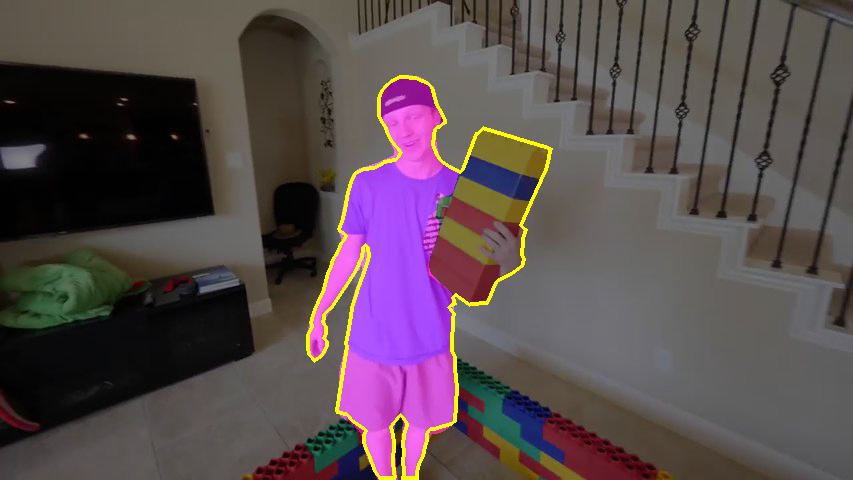}}
		&\hspace{-4.5mm}
		\bmvaHangBox{\includegraphics[width=0.135\textwidth]{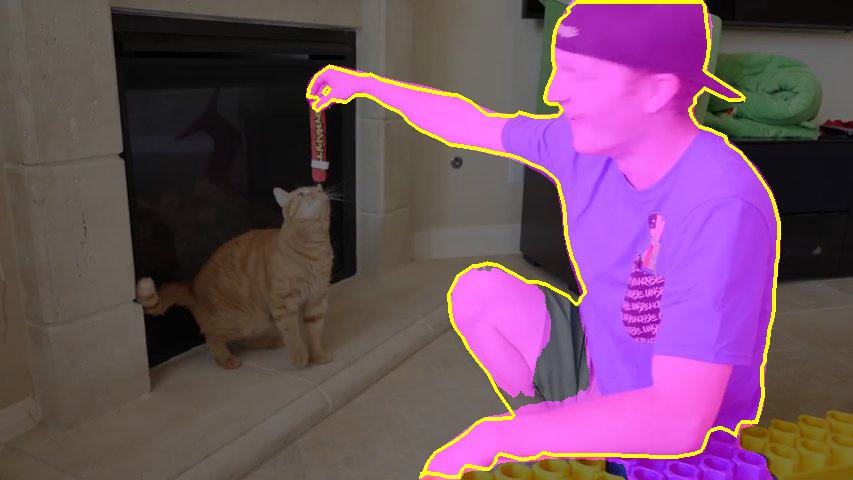}}
		&\hspace{-4.5mm}
		\bmvaHangBox{\includegraphics[width=0.135\textwidth]{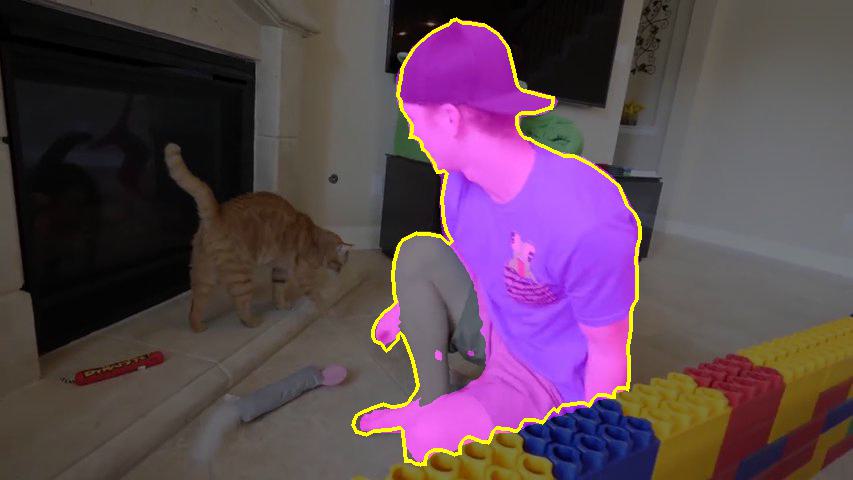}}
		&\hspace{-4.5mm}
		\bmvaHangBox{\includegraphics[width=0.135\textwidth]{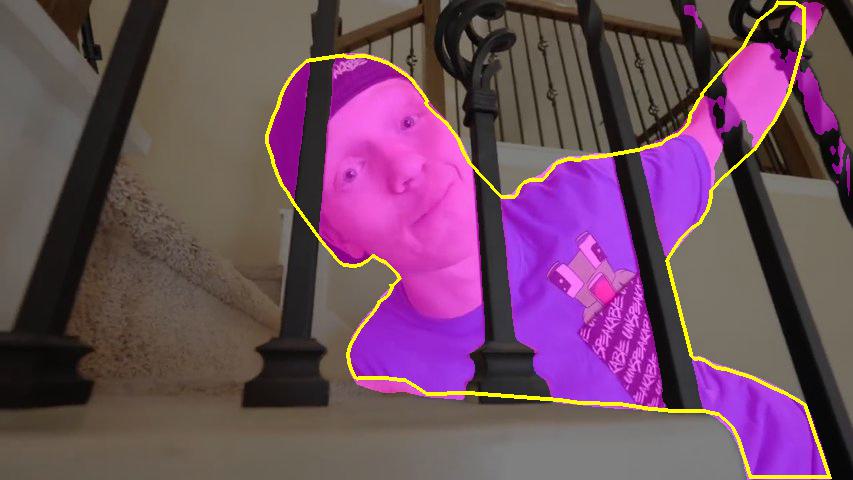}}
	\end{tabular}
\caption{Results on the \textit{blueboy} sequence of \LV1 with $s_r = 10$.
	We depict the results of: the baselines in {\color{red!80!black}red}, the READMem variations in {\color{IBMINDIGO!}blue}, the intersection between both in {\color{IBMTURQUOISE!70!black}turquoise}, the ground-truth contours in {\color{IBMGOLD!70!black}yellow} and \XMEM results in {\color{IBMMAGENTA1!}purple}.
}
\label{fig:SUPP_BLUEBOY_sr_10}
\end{figure}

\begin{figure}[H]
	\centering
	\begin{tabular}{p{2.4cm}ccccc}
		\centering {\footnotesize \MiVOS \vs READMem-MiVOS}
		&
		\bmvaHangBox{\includegraphics[width=0.135\textwidth]{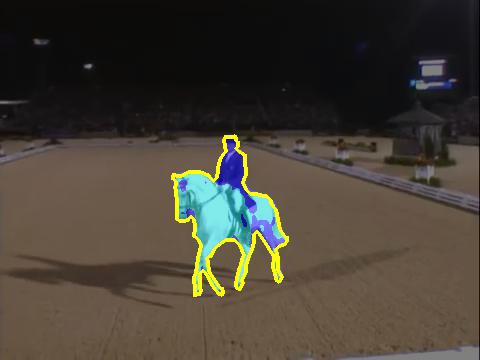}
			\put(-48.0,32.0){\makebox[0pt][l]{{\color{IBMGOLD}\textbf{\tiny \#3075}}}}}
		&\hspace{-4.5mm}
		\bmvaHangBox{\includegraphics[width=0.135\textwidth]{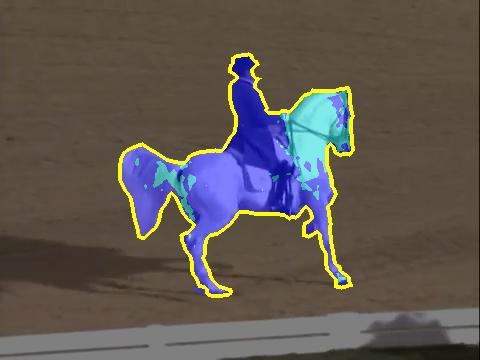}
			\put(-48.0,32.0){\makebox[0pt][l]{{\color{IBMGOLD}\textbf{\tiny \#3639}}}}}
		&\hspace{-4.5mm}
		\bmvaHangBox{\includegraphics[width=0.135\textwidth]{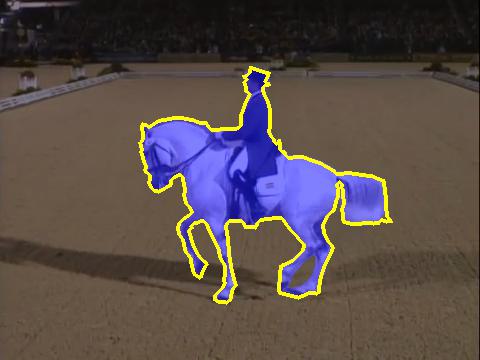}
			\put(-48.0,32.0){\makebox[0pt][l]{{\color{IBMGOLD}\textbf{\tiny \#4767}}}}}
		&\hspace{-4.5mm}
		\bmvaHangBox{\includegraphics[width=0.135\textwidth]{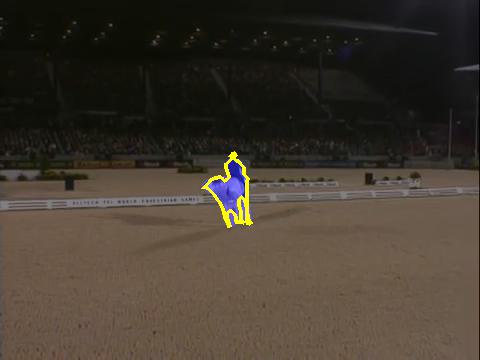}
			\put(-48.0,32.0){\makebox[0pt][l]{{\color{IBMGOLD}\textbf{\tiny \#10407}}}}}
		&\hspace{-4.5mm}
		\bmvaHangBox{\includegraphics[width=0.135\textwidth]{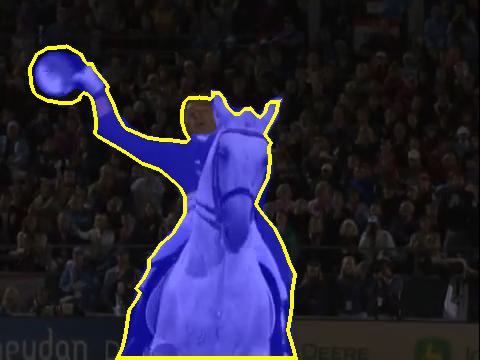}
			\put(-48.0,32.0){\makebox[0pt][l]{{\color{IBMGOLD}\textbf{\tiny \#12711}}}}}
		\\
		\centering {\footnotesize \STCN \vs READMem-STCN}
		&
		\bmvaHangBox{\includegraphics[width=0.135\textwidth]{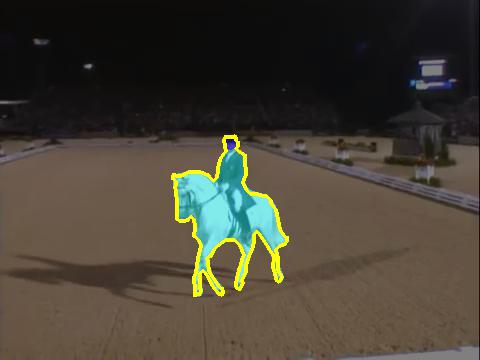}}
		&\hspace{-4.5mm}
		\bmvaHangBox{\includegraphics[width=0.135\textwidth]{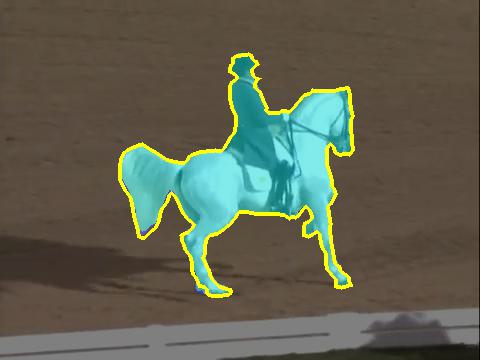}}
		&\hspace{-4.5mm}
		\bmvaHangBox{\includegraphics[width=0.135\textwidth]{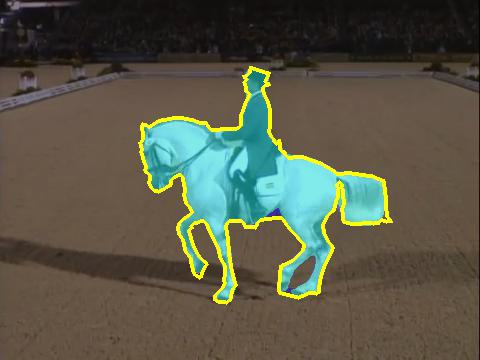}}
		&\hspace{-4.5mm}
		\bmvaHangBox{\includegraphics[width=0.135\textwidth]{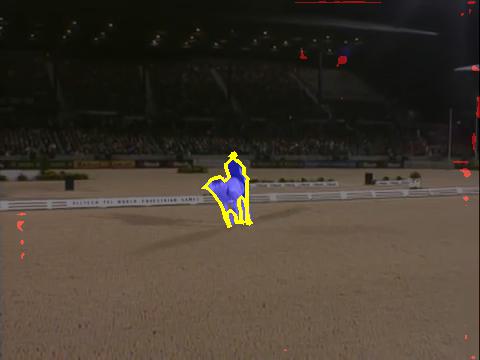}}
		&\hspace{-4.5mm}
		\bmvaHangBox{\includegraphics[width=0.135\textwidth]{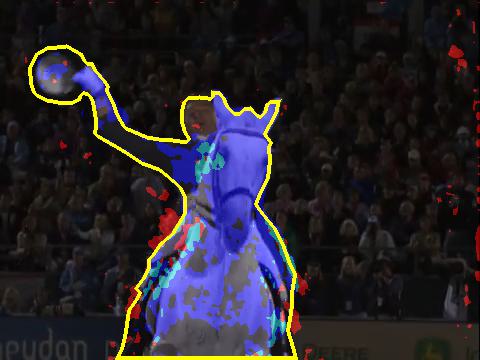}}
		\\
		\centering {\footnotesize \QDMN \vs READMem-QDMN}
		&
		\bmvaHangBox{\includegraphics[width=0.135\textwidth]{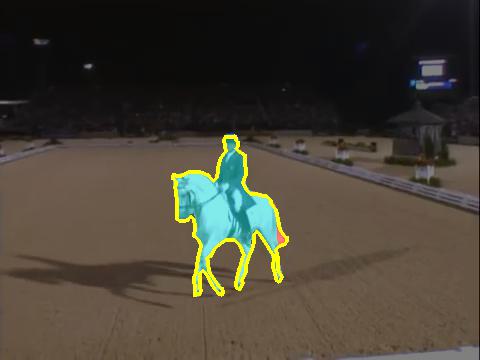}}
		&\hspace{-4.5mm}
		\bmvaHangBox{\includegraphics[width=0.135\textwidth]{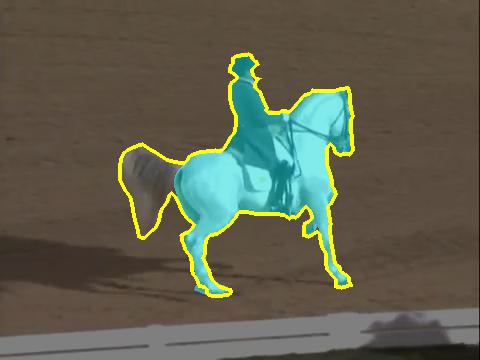}}
		&\hspace{-4.5mm}
		\bmvaHangBox{\includegraphics[width=0.135\textwidth]{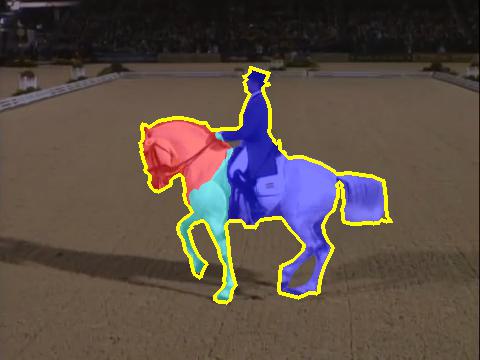}}
		&\hspace{-4.5mm}
		\bmvaHangBox{\includegraphics[width=0.135\textwidth]{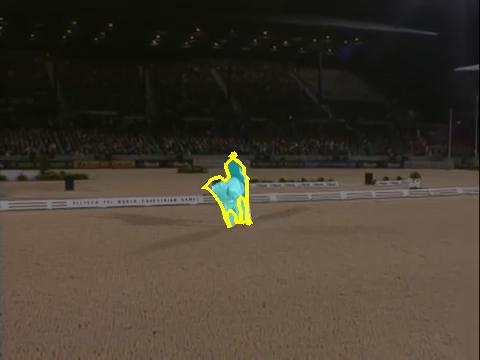}}
		&\hspace{-4.5mm}
		\bmvaHangBox{\includegraphics[width=0.135\textwidth]{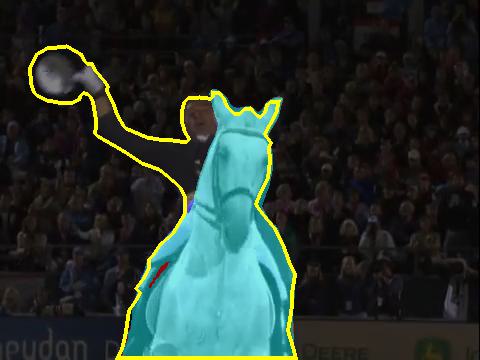}}
		\\
		\centering {\footnotesize \XMEM}
		&
		\bmvaHangBox{\includegraphics[width=0.135\textwidth]{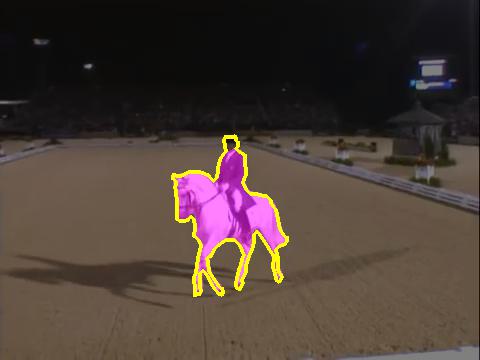}}
		&\hspace{-4.5mm}
		\bmvaHangBox{\includegraphics[width=0.135\textwidth]{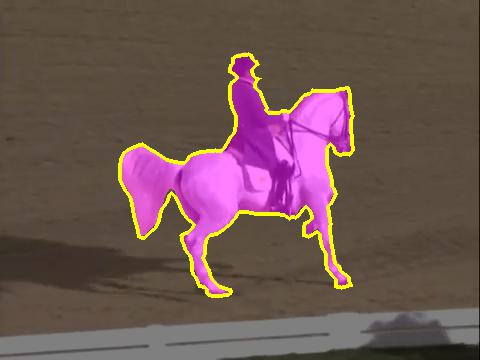}}
		&\hspace{-4.5mm}
		\bmvaHangBox{\includegraphics[width=0.135\textwidth]{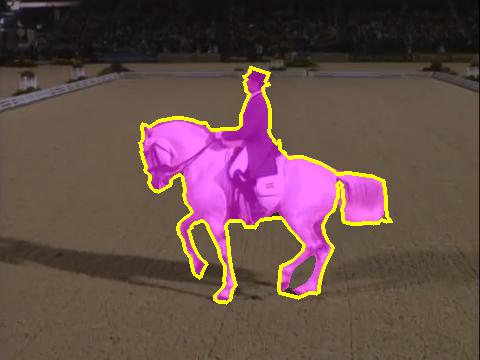}}
		&\hspace{-4.5mm}
		\bmvaHangBox{\includegraphics[width=0.135\textwidth]{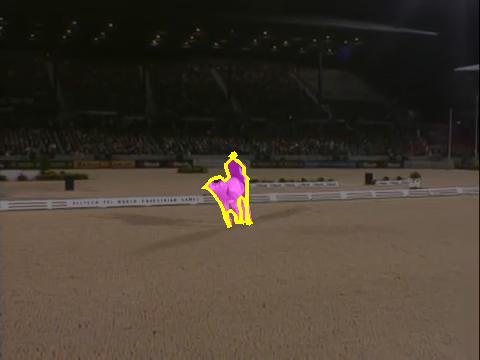}}
		&\hspace{-4.5mm}
		\bmvaHangBox{\includegraphics[width=0.135\textwidth]{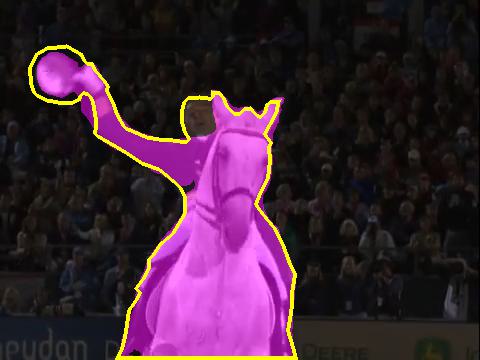}}
	\end{tabular}
	\caption{Results on the \textit{dressage} sequence of \LV1 with $s_r = 10$.
	We depict the results of: the baselines in {\color{red!80!black}red}, the READMem variations in {\color{IBMINDIGO!}blue}, the intersection between both in {\color{IBMTURQUOISE!70!black}turquoise}, the ground-truth contours in {\color{IBMGOLD!70!black}yellow} and \XMEM results in {\color{IBMMAGENTA1!}purple}.
	}
	\label{fig:SUPP_DRESSAGE_sr_10}
\end{figure}

\begin{figure}[H]
	\centering
	\begin{tabular}{p{2.4cm}ccccc}
		\centering {\footnotesize  \MiVOS \vs READMem-MiVOS}
		&
		\bmvaHangBox{\includegraphics[width=0.135\textwidth]{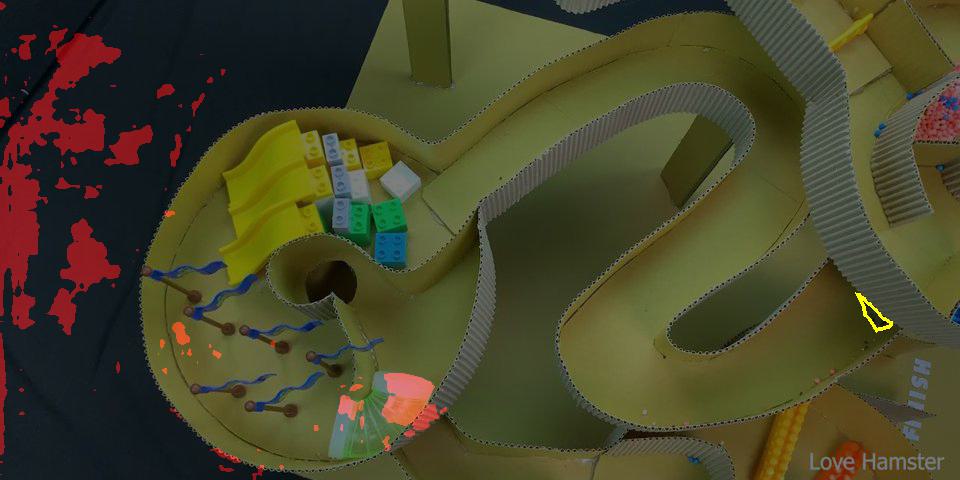}
			\put(-48.0,20.0){\makebox[0pt][l]{{\color{IBMGOLD}\textbf{\tiny \#2931}}}}}
		&\hspace{-4.5mm}
		\bmvaHangBox{\includegraphics[width=0.135\textwidth]{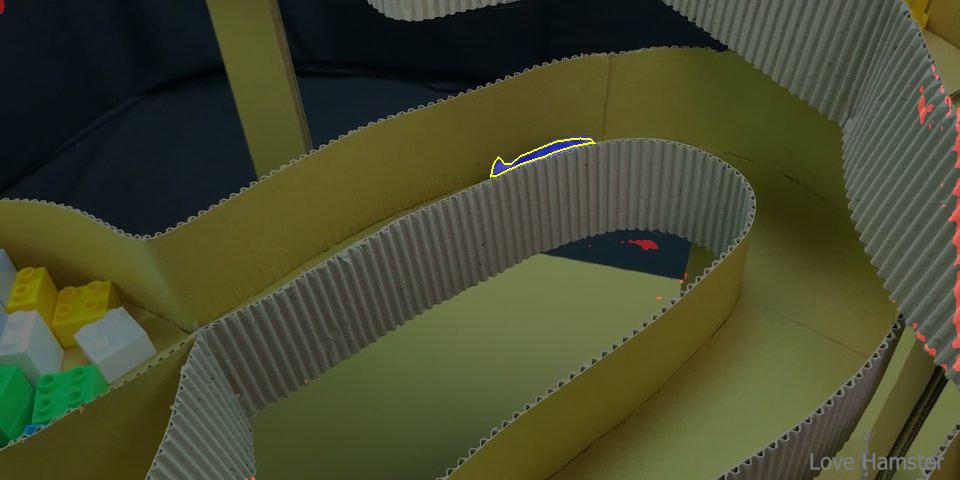}
			\put(-48.0,20.0){\makebox[0pt][l]{{\color{IBMGOLD}\textbf{\tiny \#4485}}}}}
		&\hspace{-4.5mm}
		\bmvaHangBox{\includegraphics[width=0.135\textwidth]{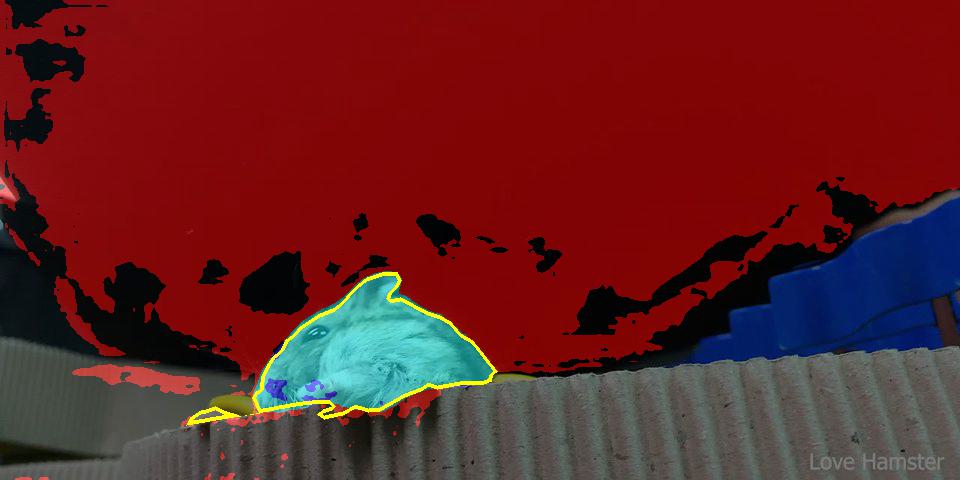}
			\put(-48.0,20.0){\makebox[0pt][l]{{\color{IBMGOLD}\textbf{\tiny \#4929}}}}}
		&\hspace{-4.5mm}
		\bmvaHangBox{\includegraphics[width=0.135\textwidth]{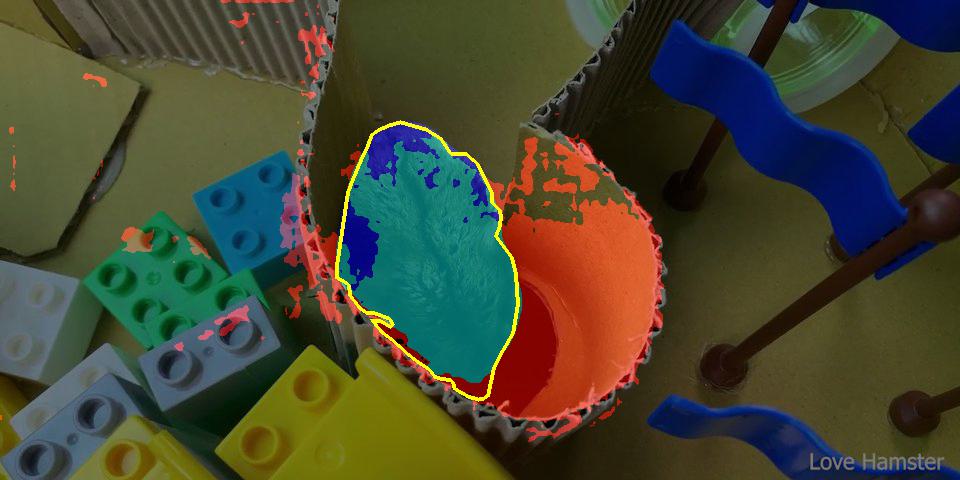}
			\put(-48.0,20.0){\makebox[0pt][l]{{\color{IBMGOLD}\textbf{\tiny \#5373}}}}}
		&\hspace{-4.5mm}
		\bmvaHangBox{\includegraphics[width=0.135\textwidth]{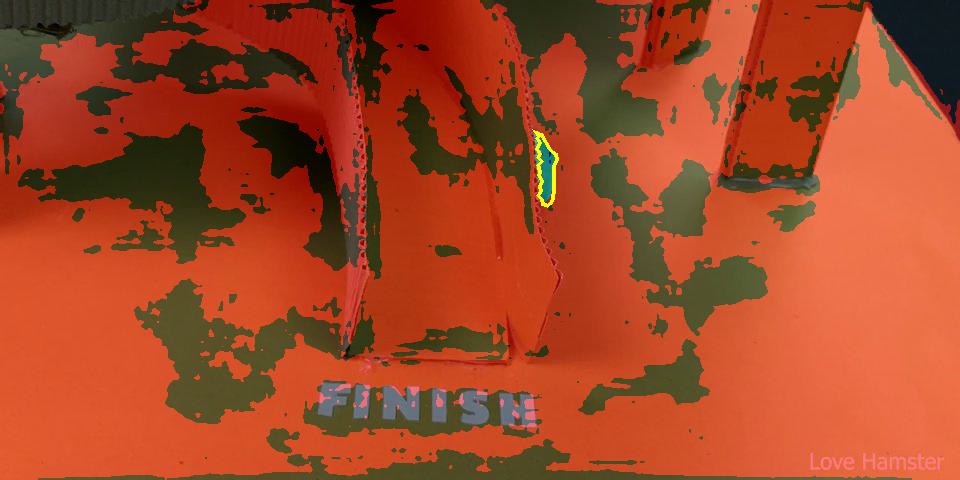}
			\put(-48.0,20.0){\makebox[0pt][l]{{\color{IBMGOLD}\textbf{\tiny \#6039}}}}}
		\\
		\centering {\footnotesize \STCN \vs READMem-STCN}
		&
		\bmvaHangBox{\includegraphics[width=0.135\textwidth]{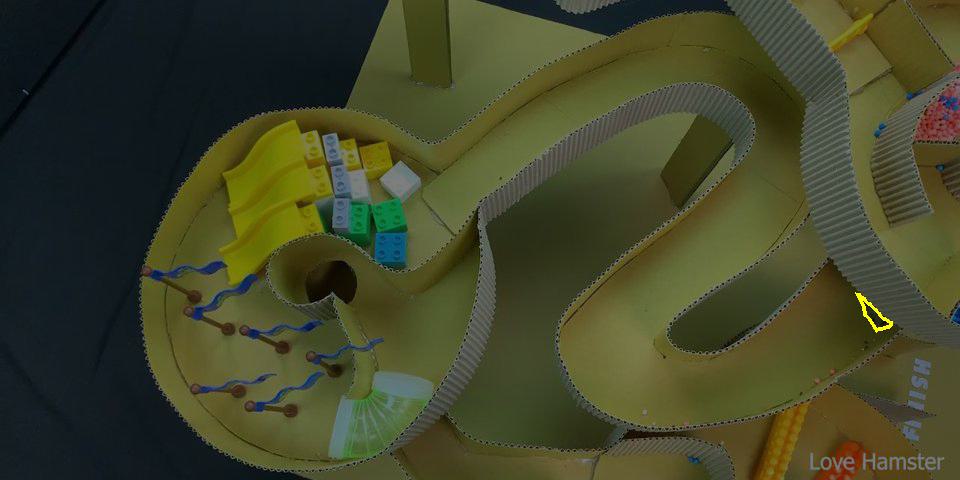}}
		&\hspace{-4.5mm}
		\bmvaHangBox{\includegraphics[width=0.135\textwidth]{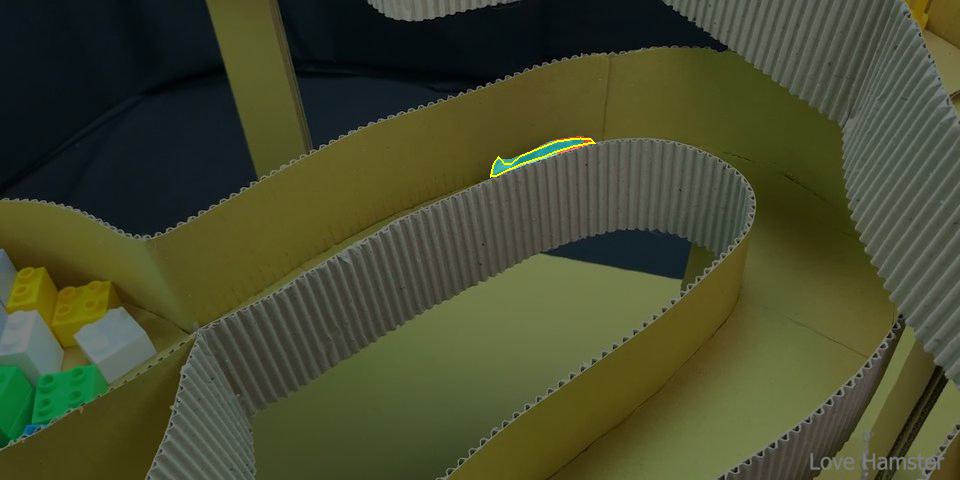}}
		&\hspace{-4.5mm}
		\bmvaHangBox{\includegraphics[width=0.135\textwidth]{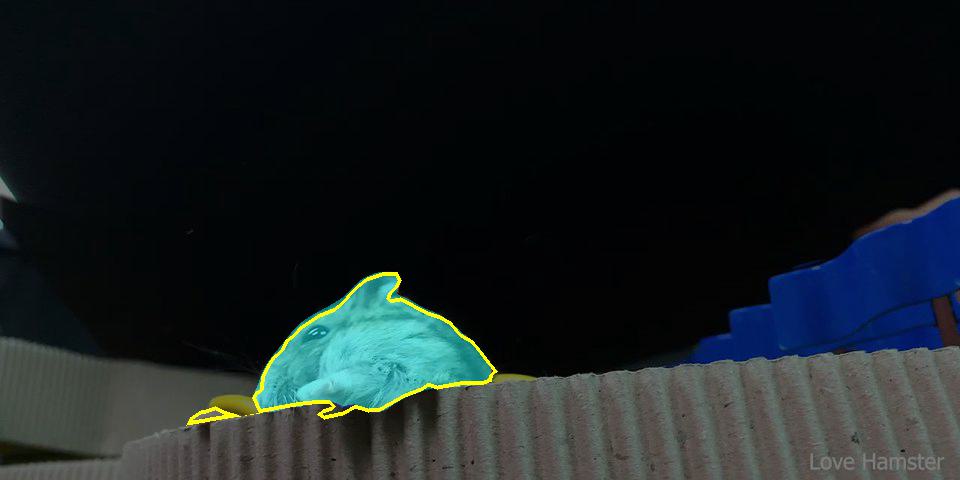}}
		&\hspace{-4.5mm}
		\bmvaHangBox{\includegraphics[width=0.135\textwidth]{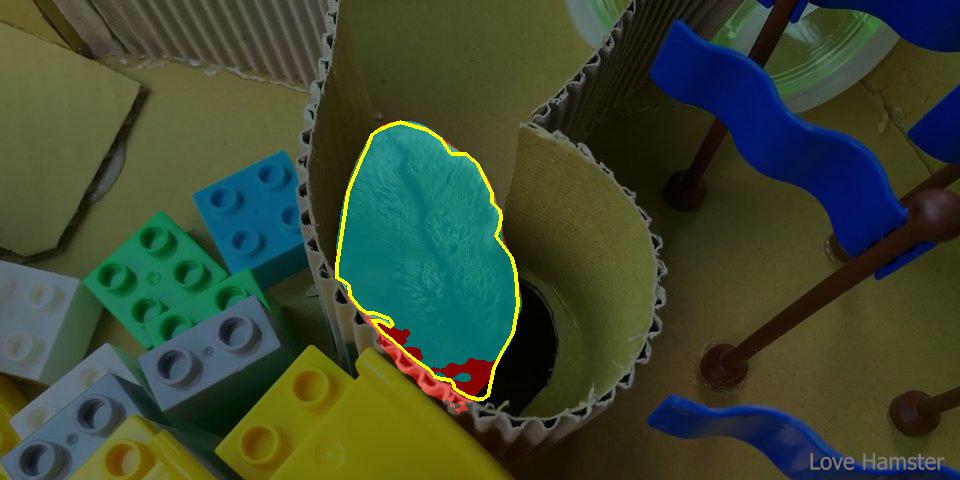}}
		&\hspace{-4.5mm}
		\bmvaHangBox{\includegraphics[width=0.135\textwidth]{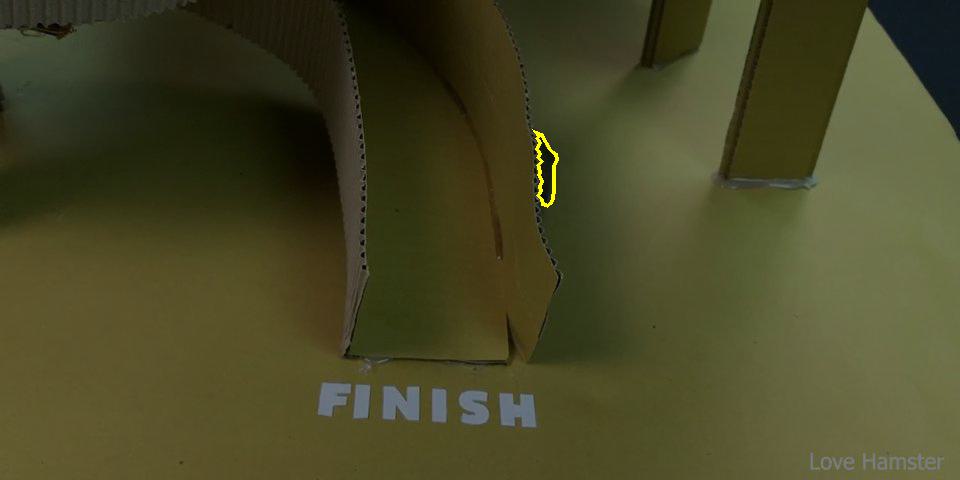}}
		\\
		\centering {\footnotesize \QDMN \vs READMem-QDMN}
		&
		\bmvaHangBox{\includegraphics[width=0.135\textwidth]{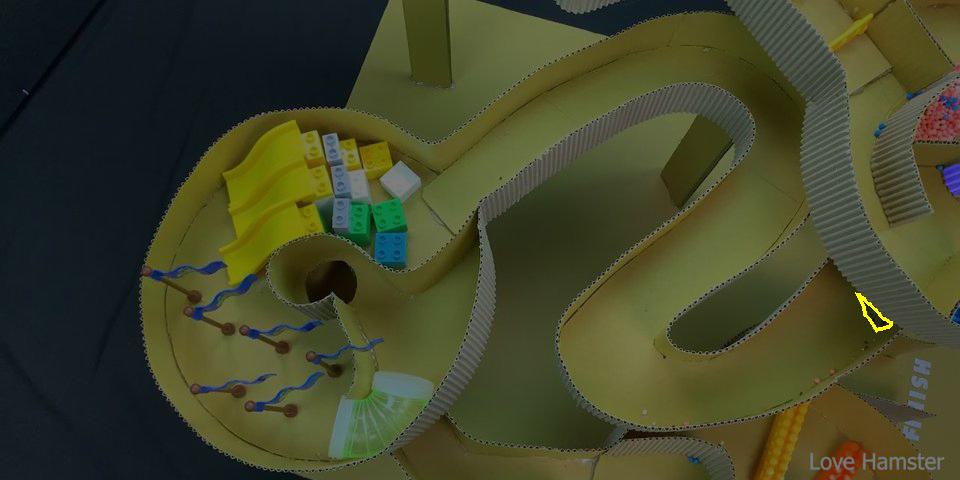}}
		&\hspace{-4.5mm}
		\bmvaHangBox{\includegraphics[width=0.135\textwidth]{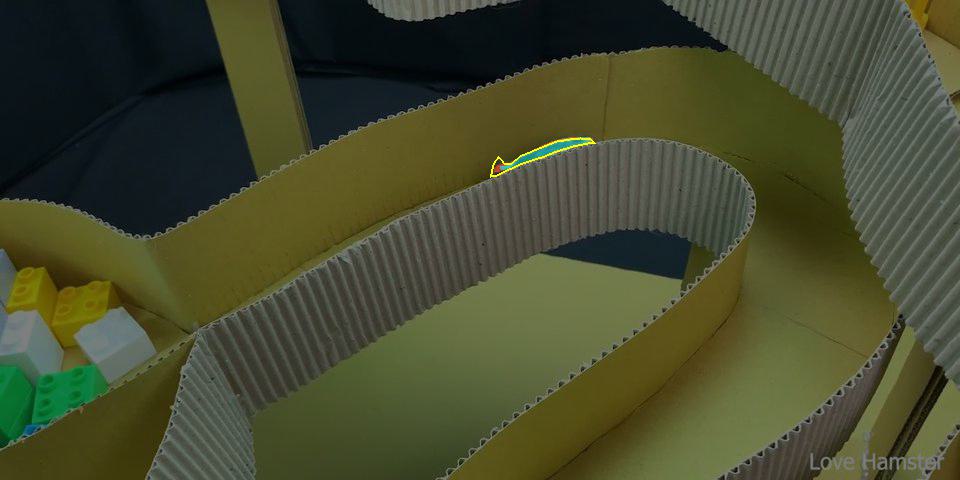}}
		&\hspace{-4.5mm}
		\bmvaHangBox{\includegraphics[width=0.135\textwidth]{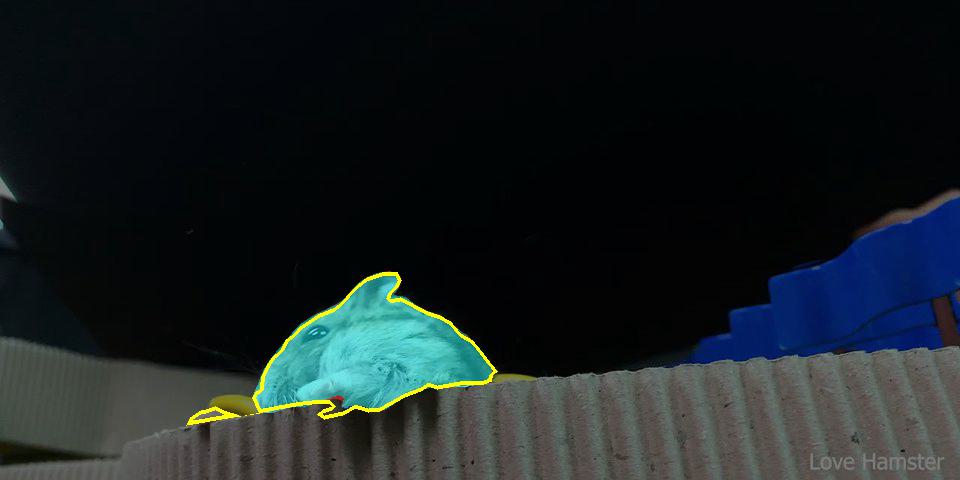}}
		&\hspace{-4.5mm}
		\bmvaHangBox{\includegraphics[width=0.135\textwidth]{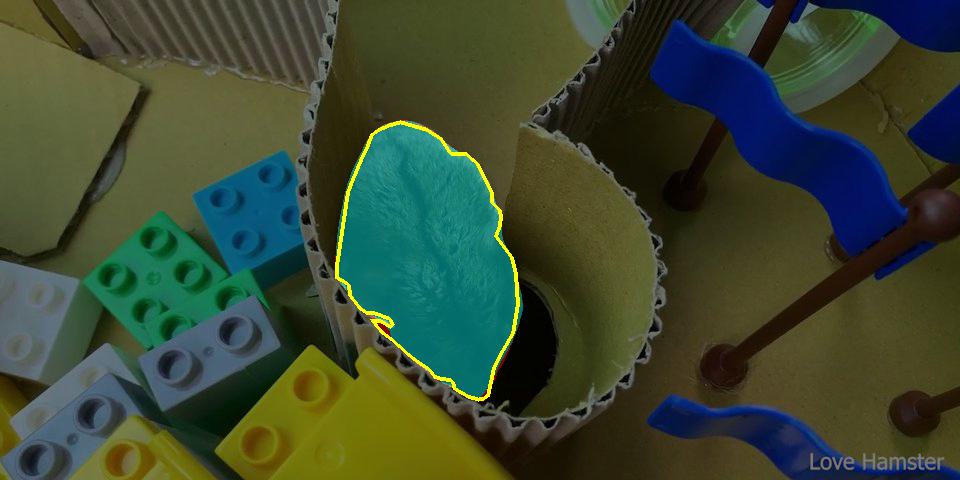}}
		&\hspace{-4.5mm}
		\bmvaHangBox{\includegraphics[width=0.135\textwidth]{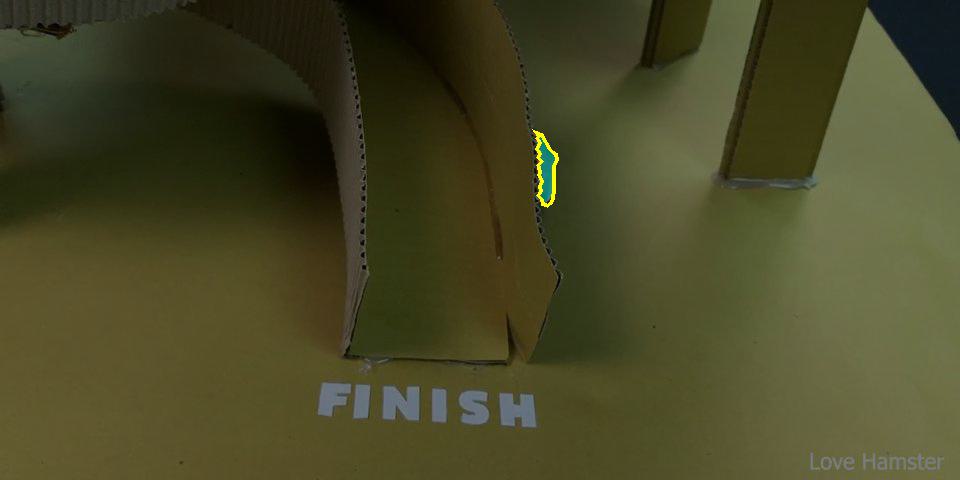}}
		\\
		\centering {\footnotesize \XMEM}
		&
		\bmvaHangBox{\includegraphics[width=0.135\textwidth]{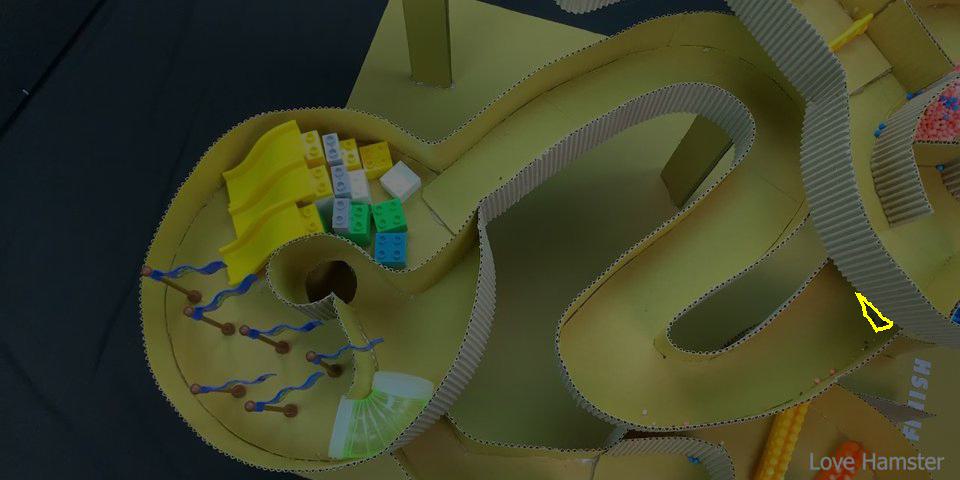}}
		&\hspace{-4.5mm}
		\bmvaHangBox{\includegraphics[width=0.135\textwidth]{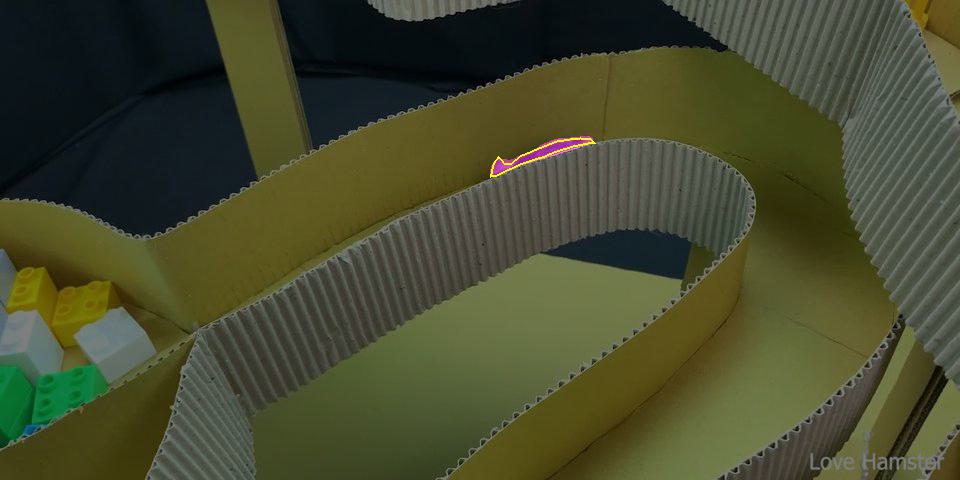}}
		&\hspace{-4.5mm}
		\bmvaHangBox{\includegraphics[width=0.135\textwidth]{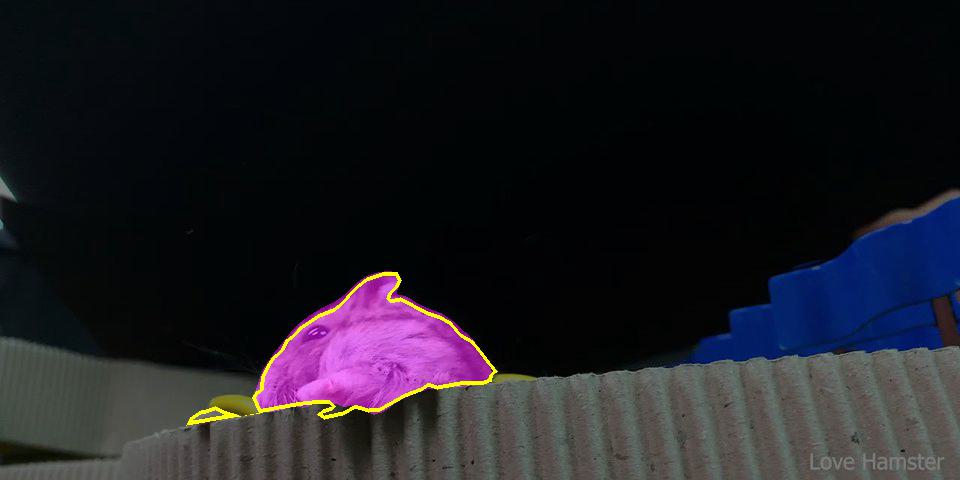}}
		&\hspace{-4.5mm}
		\bmvaHangBox{\includegraphics[width=0.135\textwidth]{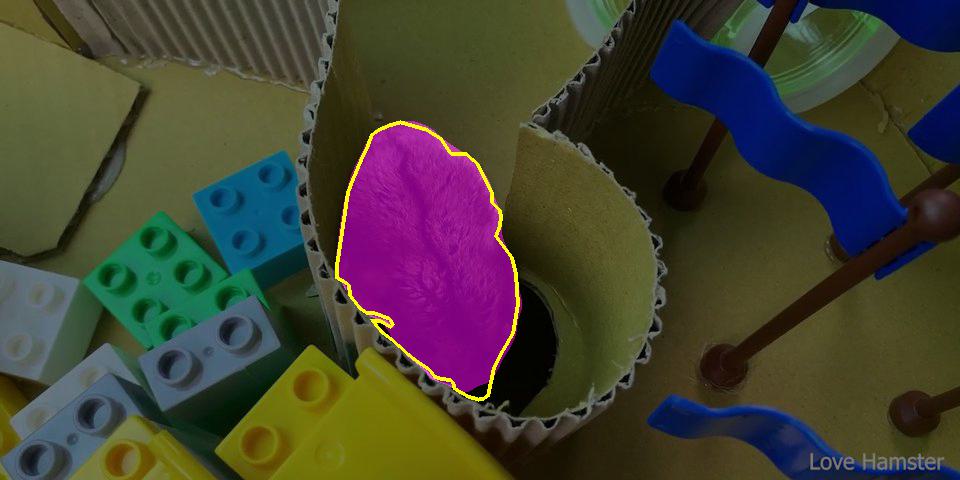}}
		&\hspace{-4.5mm}
		\bmvaHangBox{\includegraphics[width=0.135\textwidth]{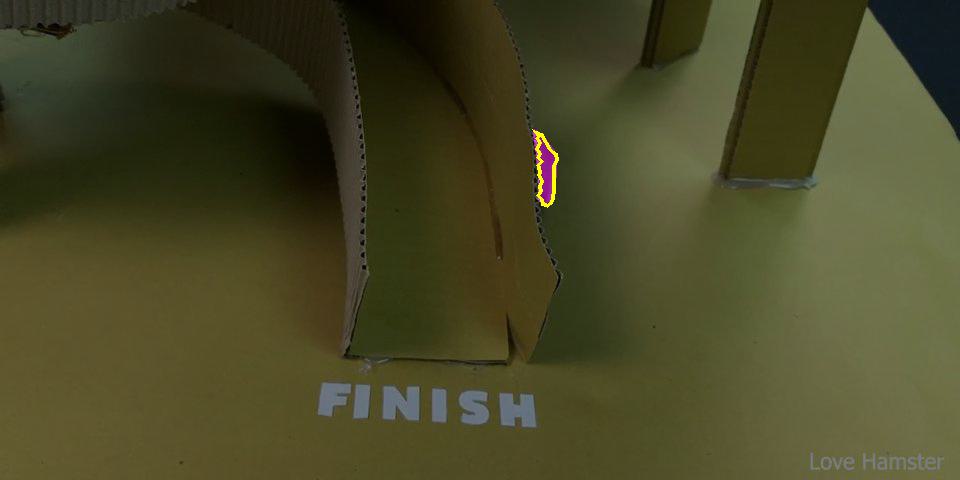}}
		\\
	\end{tabular}
	\caption{Results on the \textit{rat} sequence of \LV1 with $s_r = 10$.
	We depict the results of: the baselines in {\color{red!80!black}red}, the READMem variations in {\color{IBMINDIGO!}blue}, the intersection between both in {\color{IBMTURQUOISE!70!black}turquoise}, the ground-truth contours in {\color{IBMGOLD!70!black}yellow} and \XMEM results in {\color{IBMMAGENTA1!}purple}.
	}
	\label{fig:SUPP_RAT_sr_10}
\end{figure}

\begin{figure}[H]
	\centering
	\begin{tabular}{p{2.4cm}ccccc}
	\centering {\footnotesize  \MiVOS \vs READMem-MiVOS} &
	\bmvaHangBox{\includegraphics[width=0.135\textwidth]{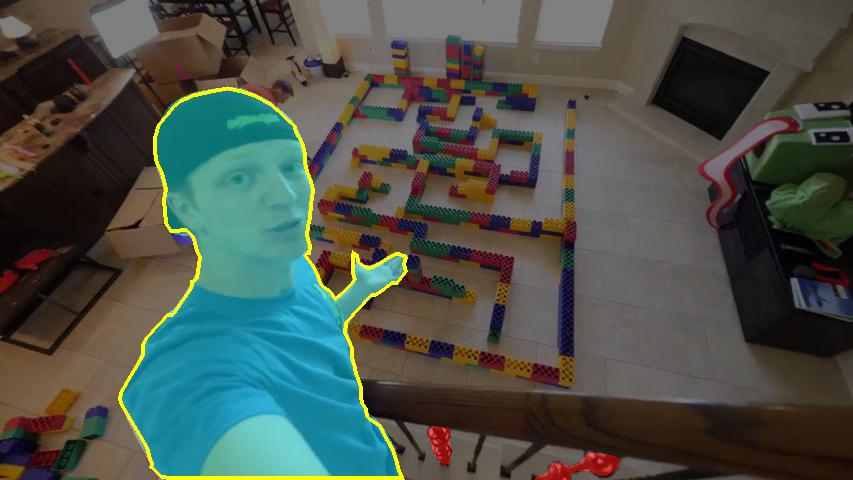}
		\put(-48.0,23.0){\makebox[0pt][l]{{\color{IBMGOLD}\textbf{\tiny \#4941}}}}}
	&\hspace{-4.5mm}
	\bmvaHangBox{\includegraphics[width=0.135\textwidth]{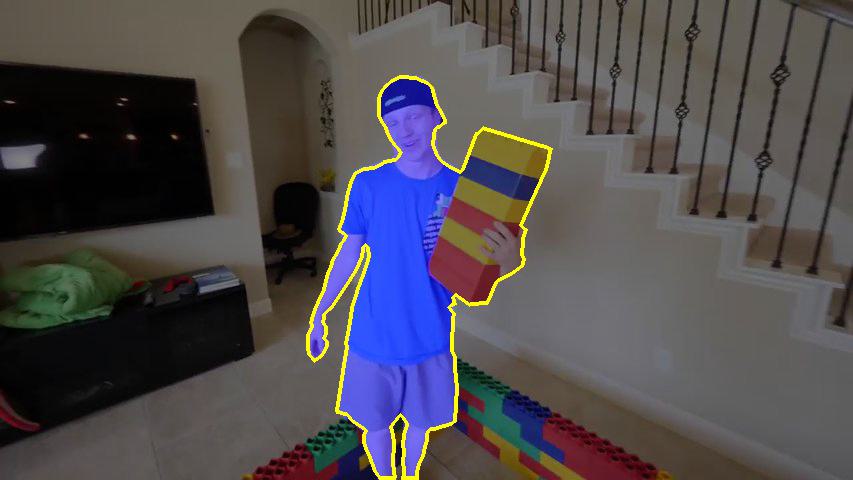}
		\put(-48.0,23.0){\makebox[0pt][l]{{\color{IBMGOLD}\textbf{\tiny \#6057}}}}}
	&\hspace{-4.5mm}
	\bmvaHangBox{\includegraphics[width=0.135\textwidth]{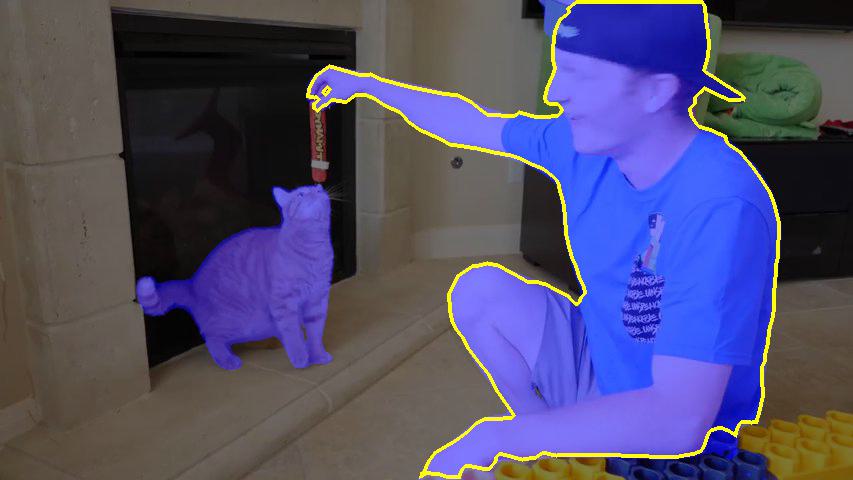}
		\put(-48.0,23.0){\makebox[0pt][l]{{\color{IBMGOLD}\textbf{\tiny \#11139}}}}}
	&\hspace{-4.5mm}
	\bmvaHangBox{\includegraphics[width=0.135\textwidth]{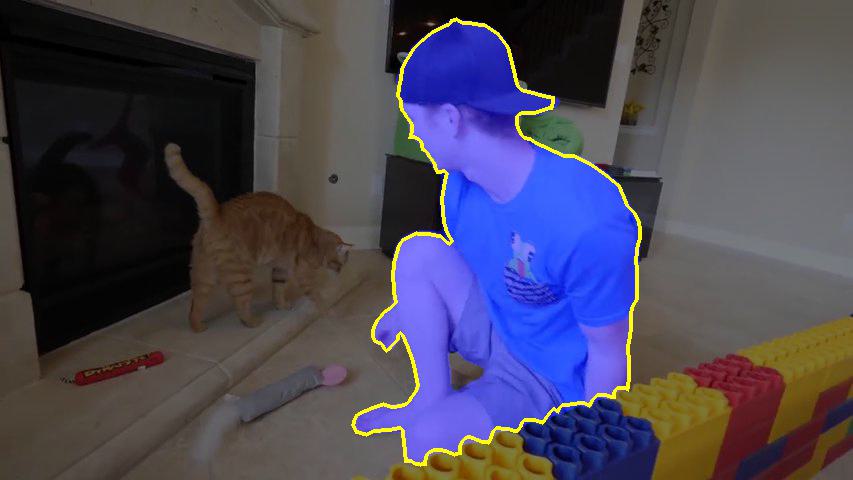}
		\put(-48.0,23.0){\makebox[0pt][l]{{\color{IBMGOLD}\textbf{\tiny \#11895}}}}}
	&\hspace{-4.5mm}
	\bmvaHangBox{\includegraphics[width=0.135\textwidth]{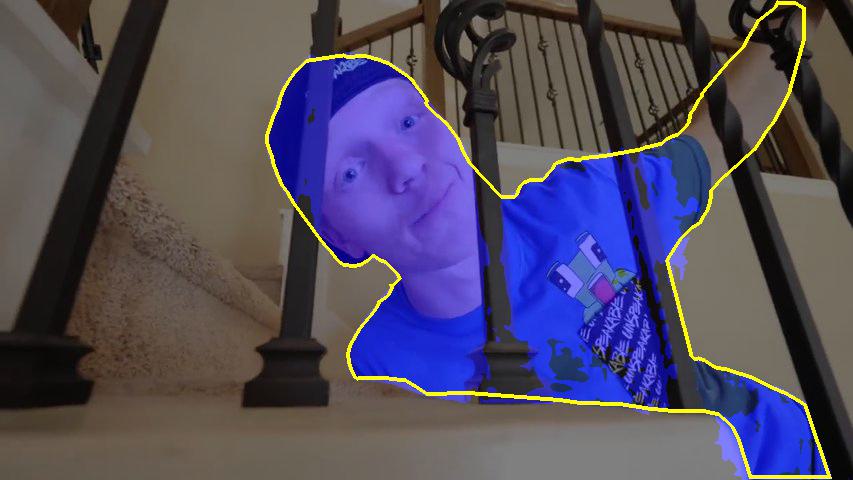}
		\put(-48.0,23.0){\makebox[0pt][l]{{\color{IBMGOLD}\textbf{\tiny \#18252}}}}}
	\\
	\centering {\footnotesize  \STCN \vs READMem-STCN} &
	\bmvaHangBox{\includegraphics[width=0.135\textwidth]{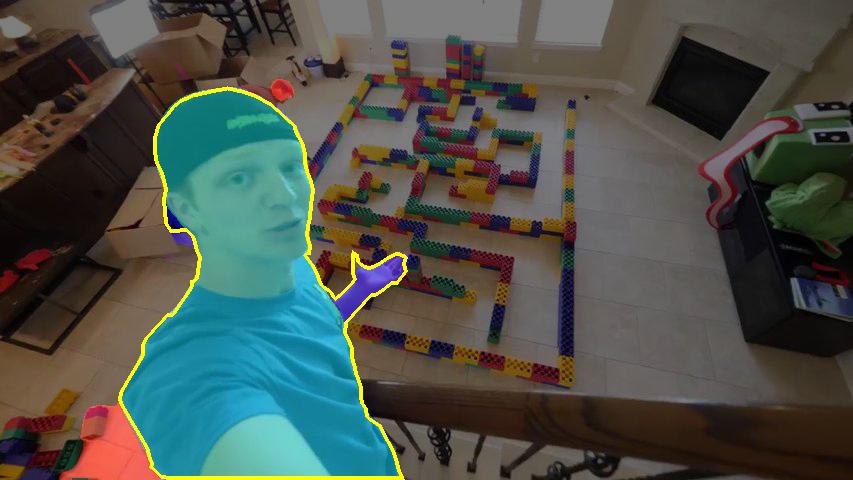}}
	&\hspace{-4.5mm}
	\bmvaHangBox{\includegraphics[width=0.135\textwidth]{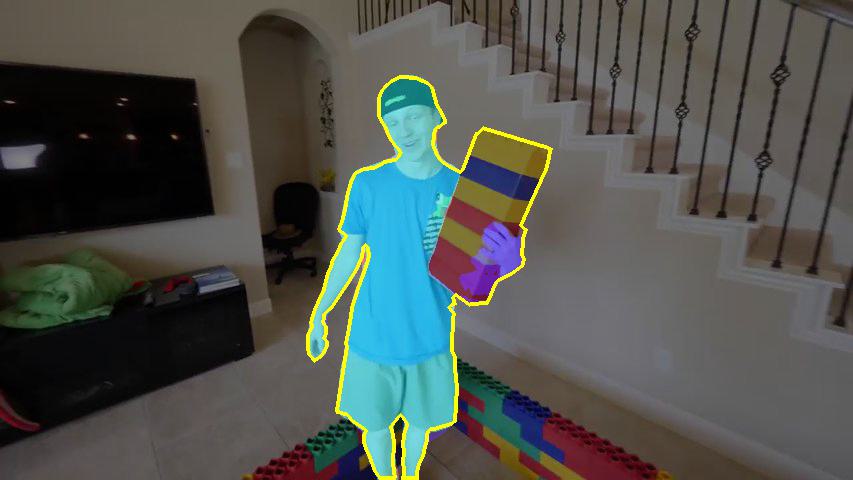}}
	&\hspace{-4.5mm}
	\bmvaHangBox{\includegraphics[width=0.135\textwidth]{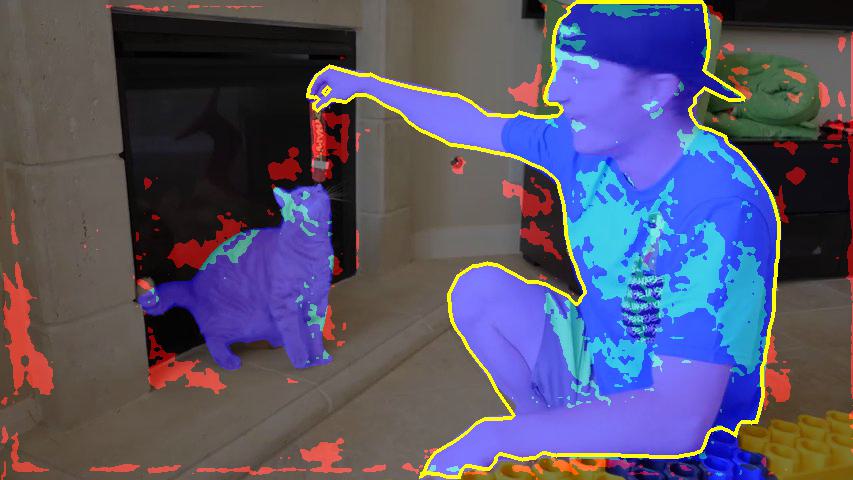}}
	&\hspace{-4.5mm}
	\bmvaHangBox{\includegraphics[width=0.135\textwidth]{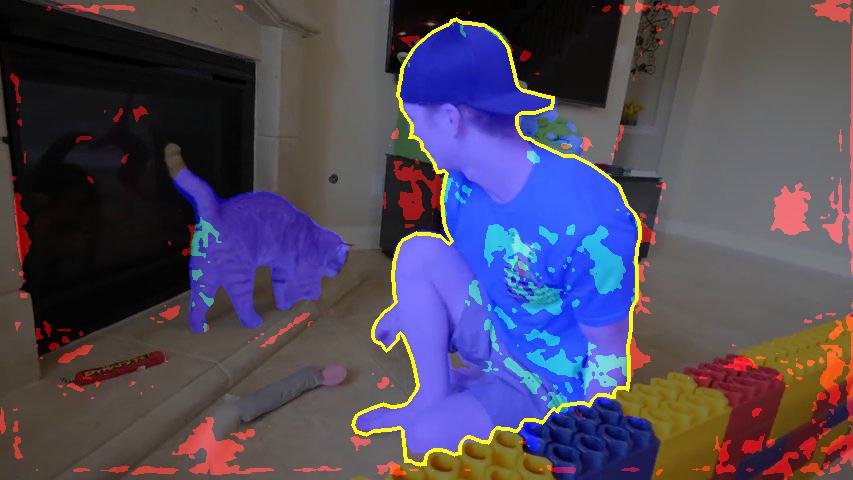}}
	&\hspace{-4.5mm}
	\bmvaHangBox{\includegraphics[width=0.135\textwidth]{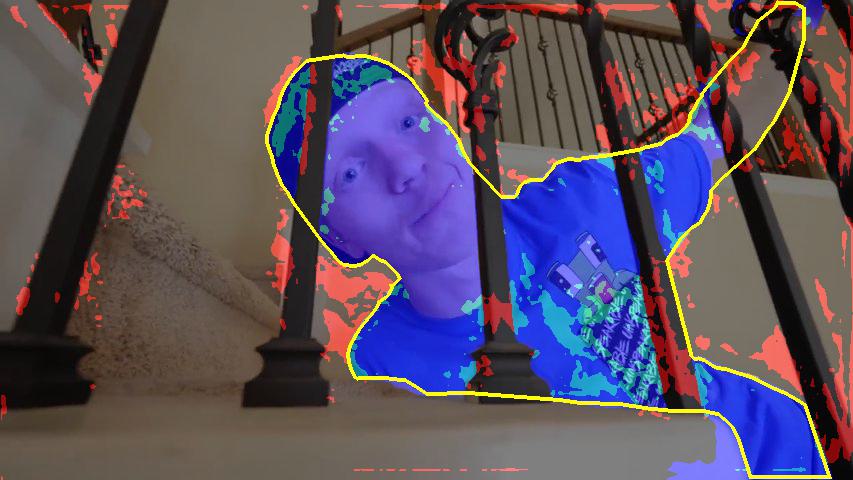}}
	\\
	\centering {\footnotesize \QDMN \vs READMem-QDMN} &
	\bmvaHangBox{\includegraphics[width=0.135\textwidth]{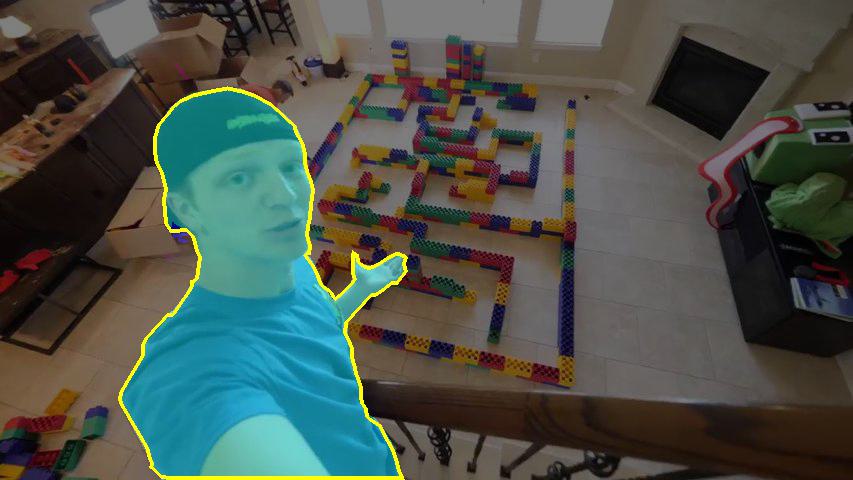}}
	&\hspace{-4.5mm}
	\bmvaHangBox{\includegraphics[width=0.135\textwidth]{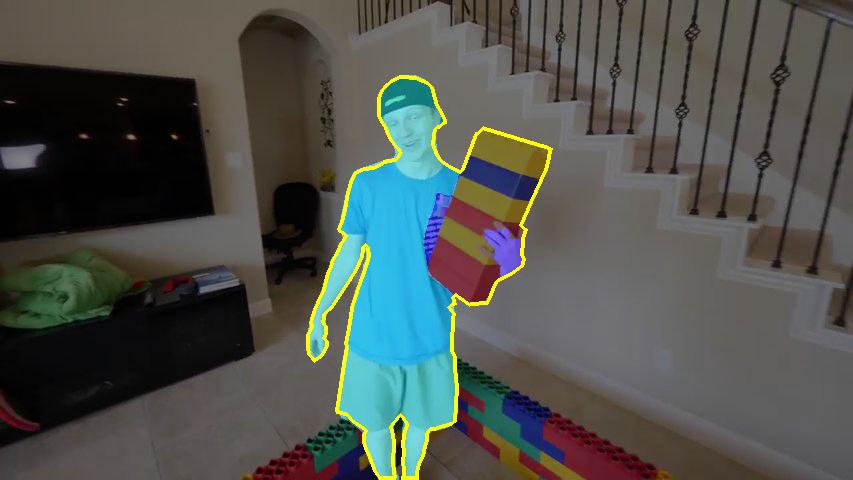}}
	&\hspace{-4.5mm}
	\bmvaHangBox{\includegraphics[width=0.135\textwidth]{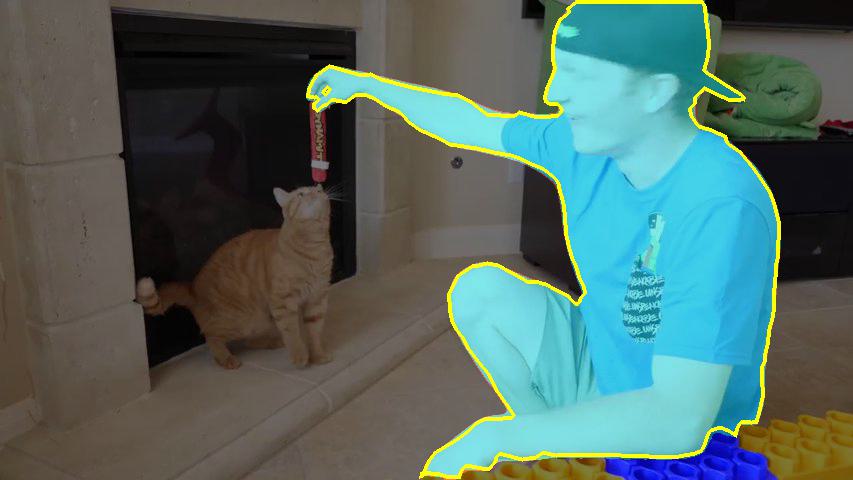}}
	&\hspace{-4.5mm}
	\bmvaHangBox{\includegraphics[width=0.135\textwidth]{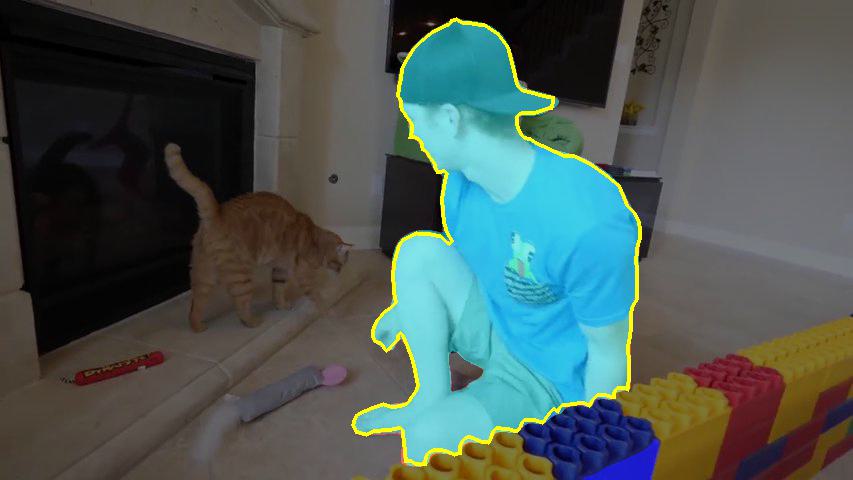}}
	&\hspace{-4.5mm}
	\bmvaHangBox{\includegraphics[width=0.135\textwidth]{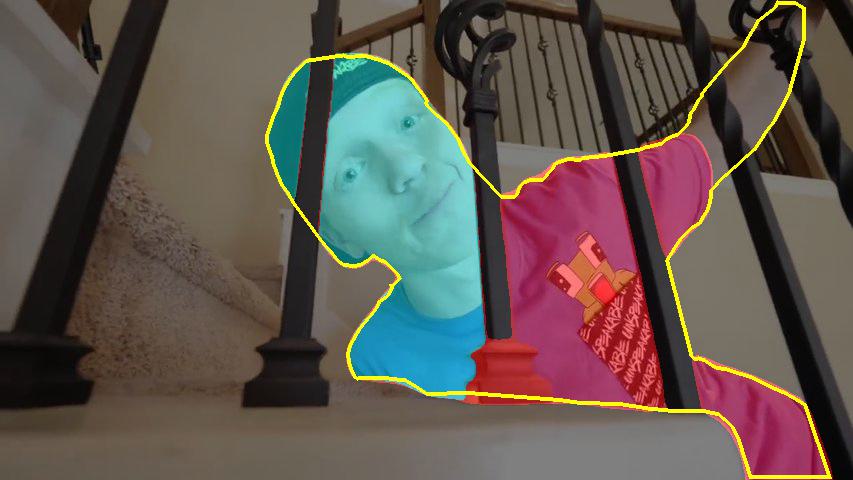}}
	\\
	\centering {\footnotesize  \XMEM} &
	\bmvaHangBox{\includegraphics[width=0.135\textwidth]{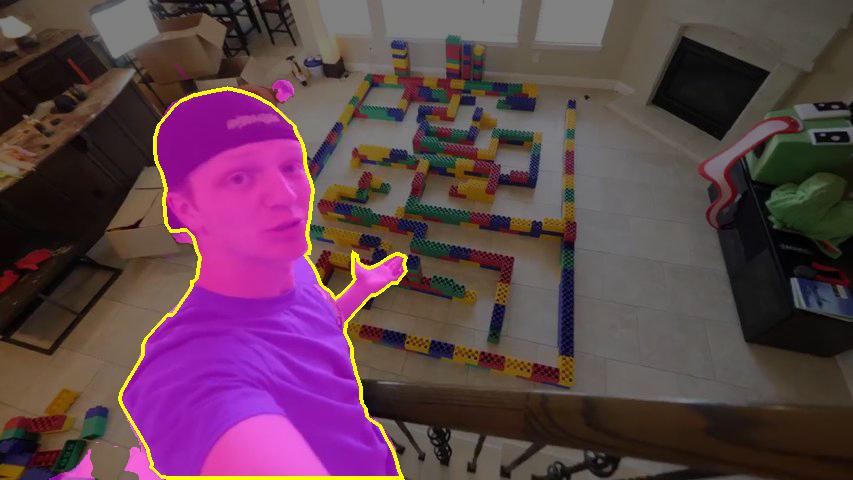}}
	&\hspace{-4.5mm}
	\bmvaHangBox{\includegraphics[width=0.135\textwidth]{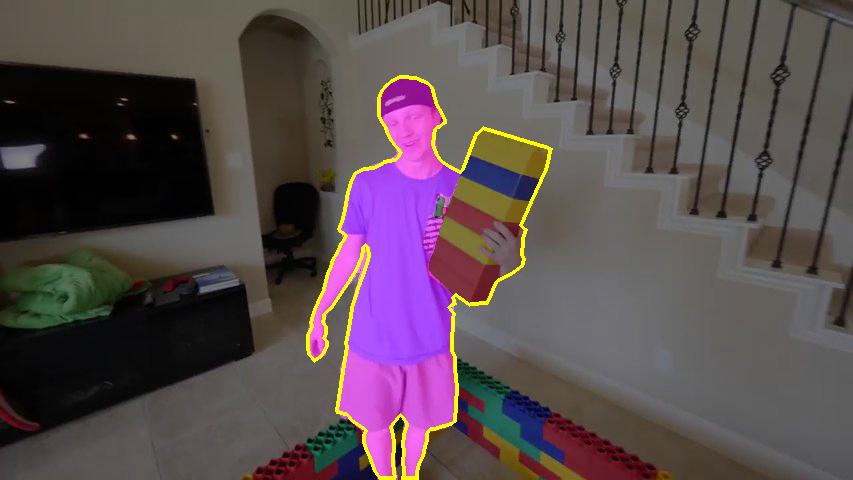}}
	&\hspace{-4.5mm}
	\bmvaHangBox{\includegraphics[width=0.135\textwidth]{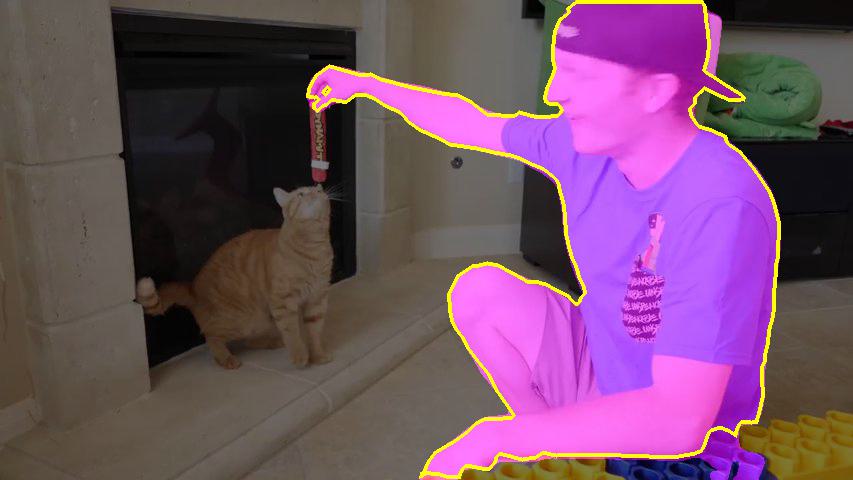}}
	&\hspace{-4.5mm}
	\bmvaHangBox{\includegraphics[width=0.135\textwidth]{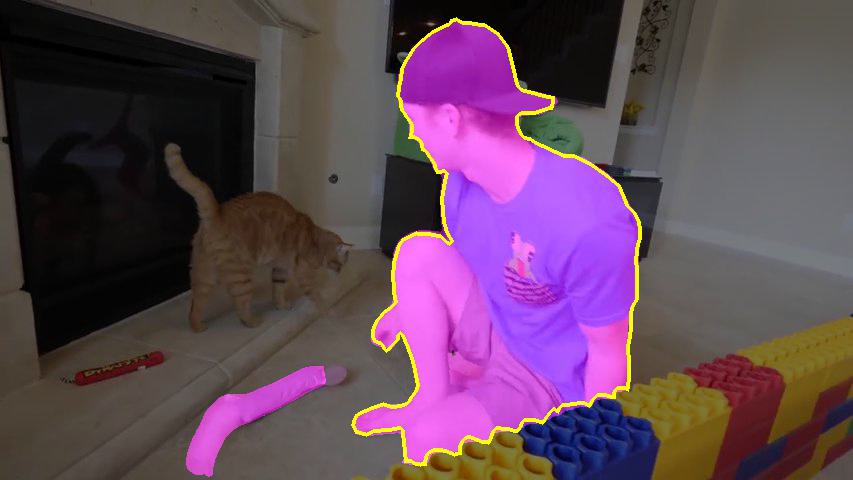}}
	&\hspace{-4.5mm}
	\bmvaHangBox{\includegraphics[width=0.135\textwidth]{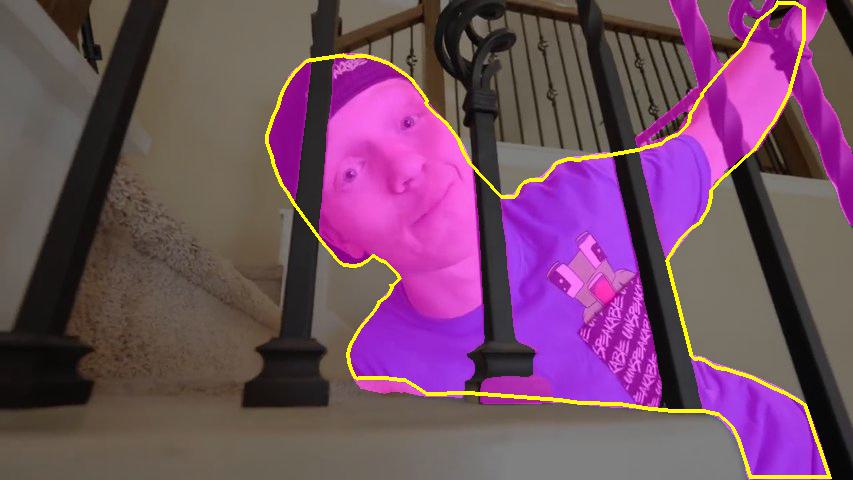}}
	\end{tabular}
	\caption{Results on the \textit{blueboy} sequence of \LV1 with $s_r = 1$.
	We depict the results of: the baselines in {\color{red!80!black}red}, the READMem variations in {\color{IBMINDIGO!}blue}, the intersection between both in {\color{IBMTURQUOISE!70!black}turquoise}, the ground-truth contours in {\color{IBMGOLD!70!black}yellow} and \XMEM results in {\color{IBMMAGENTA1!}purple}.
	}
	\label{fig:SUPP_BLUEBOY}
\end{figure}

\begin{figure}[H]
	\centering
	\begin{tabular}{p{2.4cm}ccccc}
	\centering {\footnotesize  \MiVOS \vs READMem-MiVOS}
	&
		\bmvaHangBox{\includegraphics[width=0.135\textwidth]{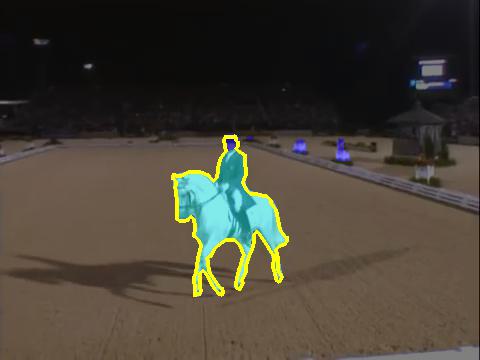}
		\put(-48.0,32.0){\makebox[0pt][l]{{\color{IBMGOLD}\textbf{\tiny \#3075}}}}}
	&\hspace{-4.5mm}
		\bmvaHangBox{\includegraphics[width=0.135\textwidth]{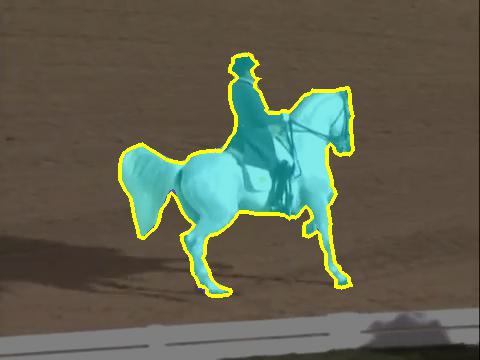}
			\put(-48.0,32.0){\makebox[0pt][l]{{\color{IBMGOLD}\textbf{\tiny \#3639}}}}}
	&\hspace{-4.5mm}
		\bmvaHangBox{\includegraphics[width=0.135\textwidth]{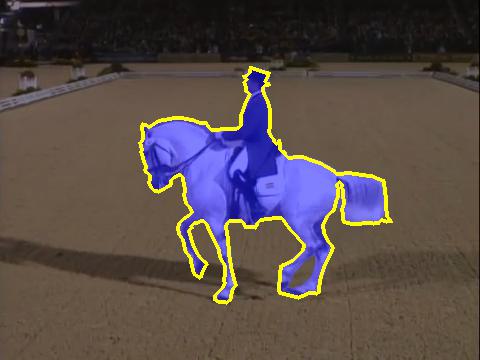}
			\put(-48.0,32.0){\makebox[0pt][l]{{\color{IBMGOLD}\textbf{\tiny \#4767}}}}}
	&\hspace{-4.5mm}
		\bmvaHangBox{\includegraphics[width=0.135\textwidth]{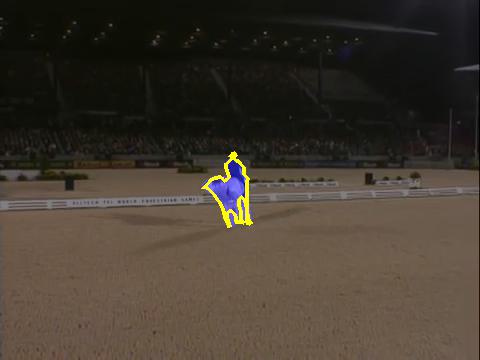}
			\put(-48.0,32.0){\makebox[0pt][l]{{\color{IBMGOLD}\textbf{\tiny \#10407}}}}}
	&\hspace{-4.5mm}
		\bmvaHangBox{\includegraphics[width=0.135\textwidth]{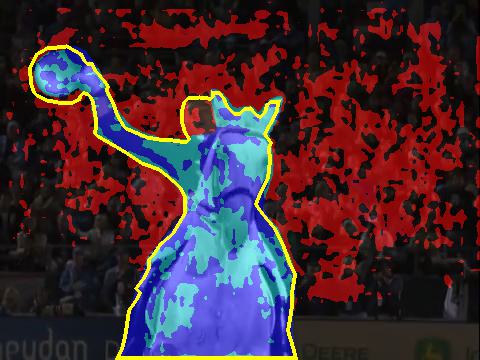}
			\put(-48.0,32.0){\makebox[0pt][l]{{\color{IBMGOLD}\textbf{\tiny \#12711}}}}}
	\\
	\centering {\footnotesize \STCN \vs READMem-STCN}
	&
		\bmvaHangBox{\includegraphics[width=0.135\textwidth]{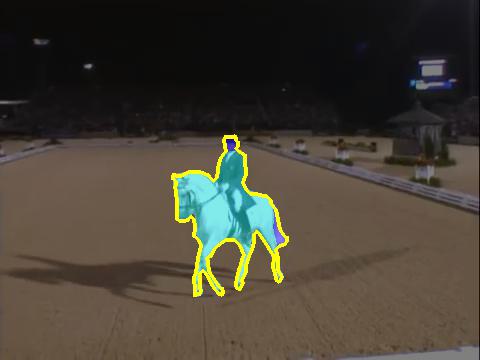}}
	&\hspace{-4.5mm}
		\bmvaHangBox{\includegraphics[width=0.135\textwidth]{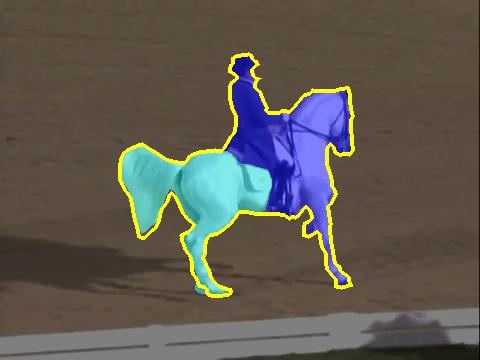}}
	&\hspace{-4.5mm}
		\bmvaHangBox{\includegraphics[width=0.135\textwidth]{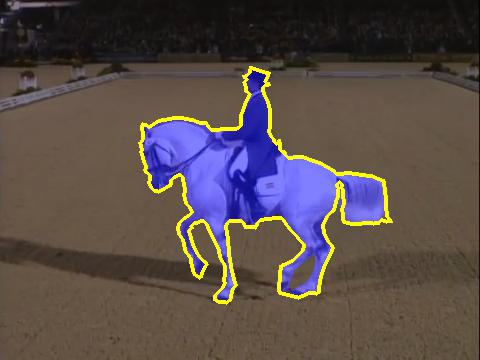}}
	&\hspace{-4.5mm}
		\bmvaHangBox{\includegraphics[width=0.135\textwidth]{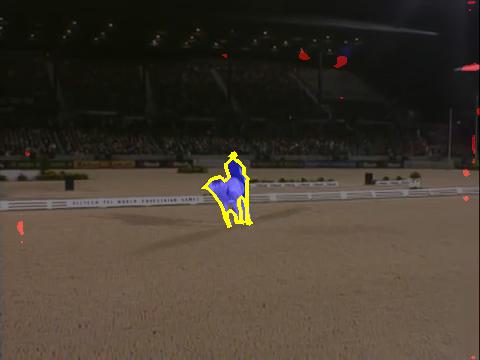}}
	&\hspace{-4.5mm}
		\bmvaHangBox{\includegraphics[width=0.135\textwidth]{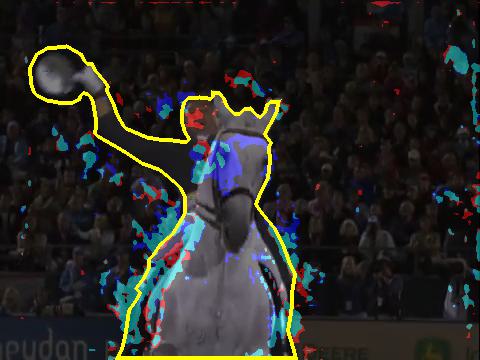}}
	\\
	\centering {\footnotesize \QDMN \vs READMem-QDMN}
	&
		\bmvaHangBox{\includegraphics[width=0.135\textwidth]{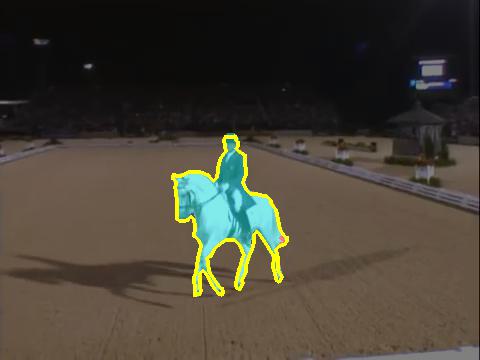}}
	&\hspace{-4.5mm}
		\bmvaHangBox{\includegraphics[width=0.135\textwidth]{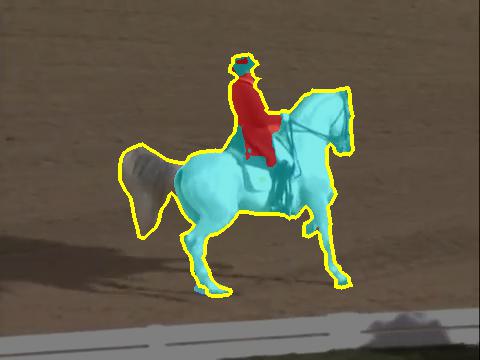}}
	&\hspace{-4.5mm}
		\bmvaHangBox{\includegraphics[width=0.135\textwidth]{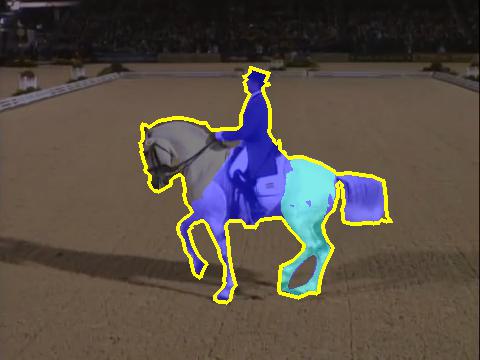}}
	&\hspace{-4.5mm}
		\bmvaHangBox{\includegraphics[width=0.135\textwidth]{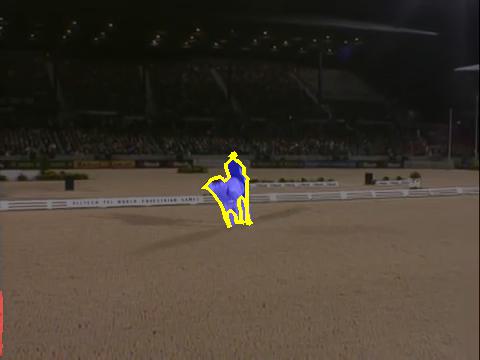}}
	&\hspace{-4.5mm}
		\bmvaHangBox{\includegraphics[width=0.135\textwidth]{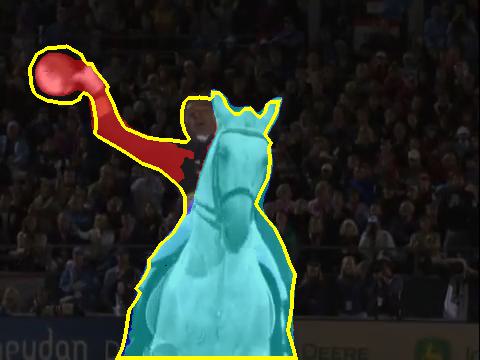}}
	\\
	\centering {\footnotesize  \XMEM}
	&
	\bmvaHangBox{\includegraphics[width=0.135\textwidth]{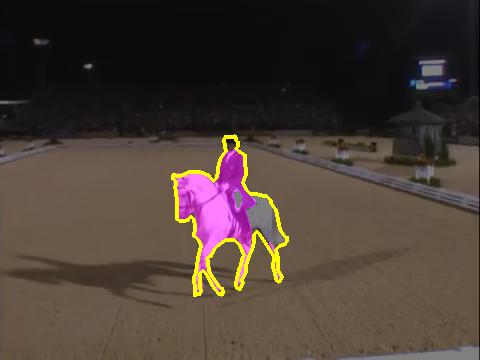}}
	&\hspace{-4.5mm}
	\bmvaHangBox{\includegraphics[width=0.135\textwidth]{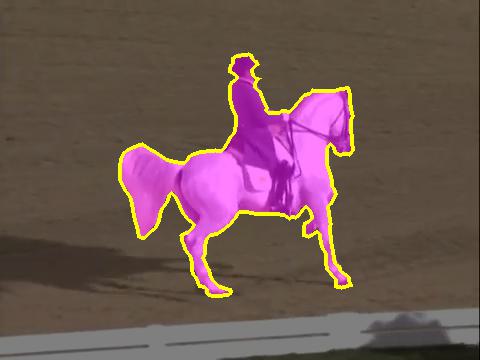}}
	&\hspace{-4.5mm}
	\bmvaHangBox{\includegraphics[width=0.135\textwidth]{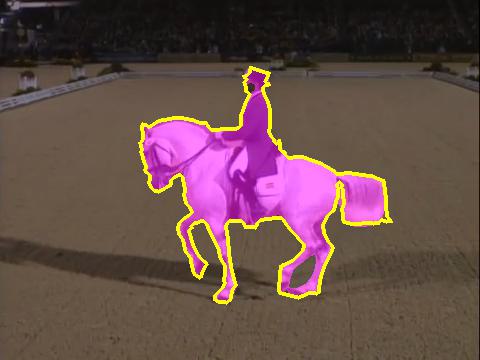}}
	&\hspace{-4.5mm}
	\bmvaHangBox{\includegraphics[width=0.135\textwidth]{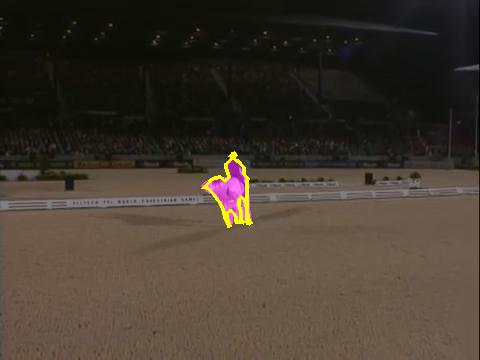}}
	&\hspace{-4.5mm}
	\bmvaHangBox{\includegraphics[width=0.135\textwidth]{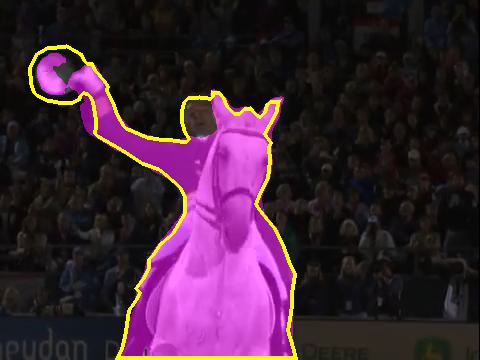}}
	\\
	\end{tabular}
	\caption{Results on the \textit{dressage} sequence of \LV1 with $s_r = 1$.
	We depict the results of: the baselines in {\color{red!80!black}red}, the READMem variations in {\color{IBMINDIGO!}blue}, the intersection between both in {\color{IBMTURQUOISE!70!black}turquoise}, the ground-truth contours in {\color{IBMGOLD!70!black}yellow} and \XMEM results in {\color{IBMMAGENTA1!}purple}.
	}
	\label{fig:SUPP_DRESSAGE}
\end{figure}

\begin{figure}[H]
	\centering
	\begin{tabular}{p{2.4cm}ccccc}
	\centering {\footnotesize \MiVOS \vs READMem-MiVOS}
		&
		\bmvaHangBox{\includegraphics[width=0.135\textwidth]{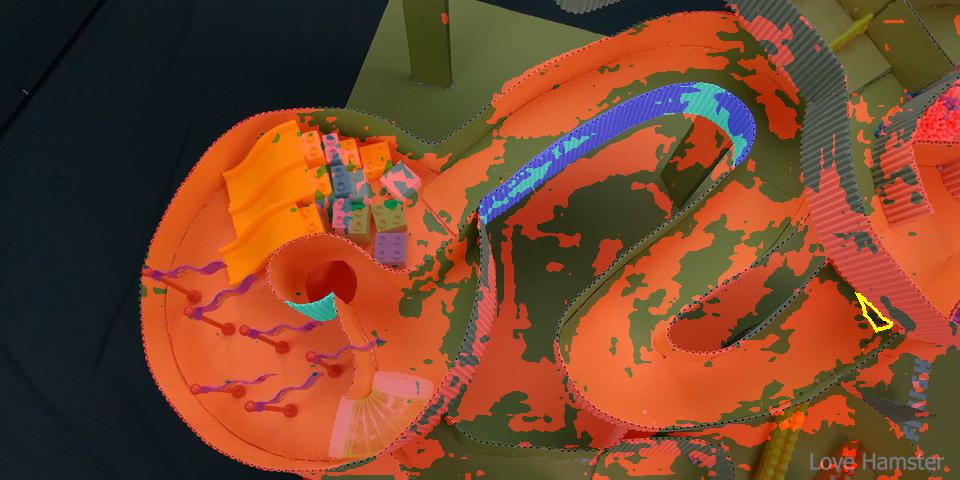}
			\put(-48.0,20.0){\makebox[0pt][l]{{\color{IBMGOLD}\textbf{\tiny \#2931}}}}}
		&\hspace{-4.5mm}
		\bmvaHangBox{\includegraphics[width=0.135\textwidth]{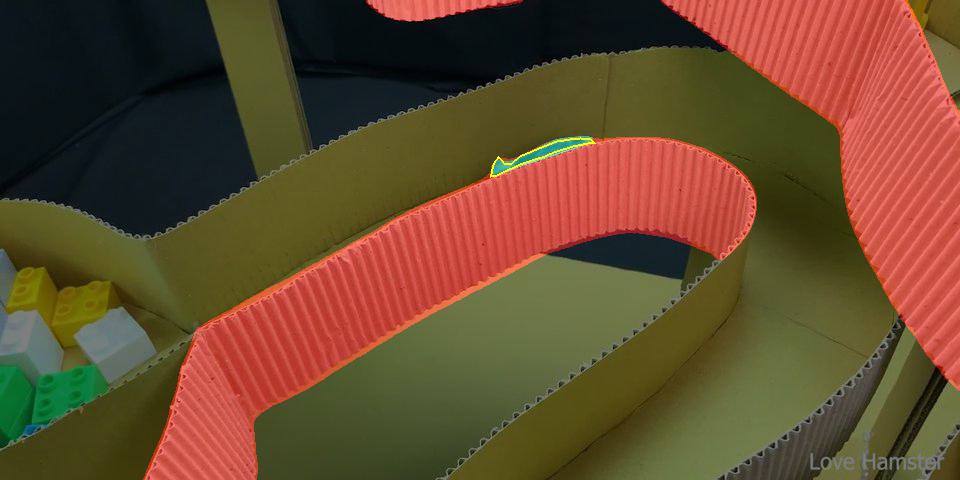}
			\put(-48.0,20.0){\makebox[0pt][l]{{\color{IBMGOLD}\textbf{\tiny \#4485}}}}}
		&\hspace{-4.5mm}
		\bmvaHangBox{\includegraphics[width=0.135\textwidth]{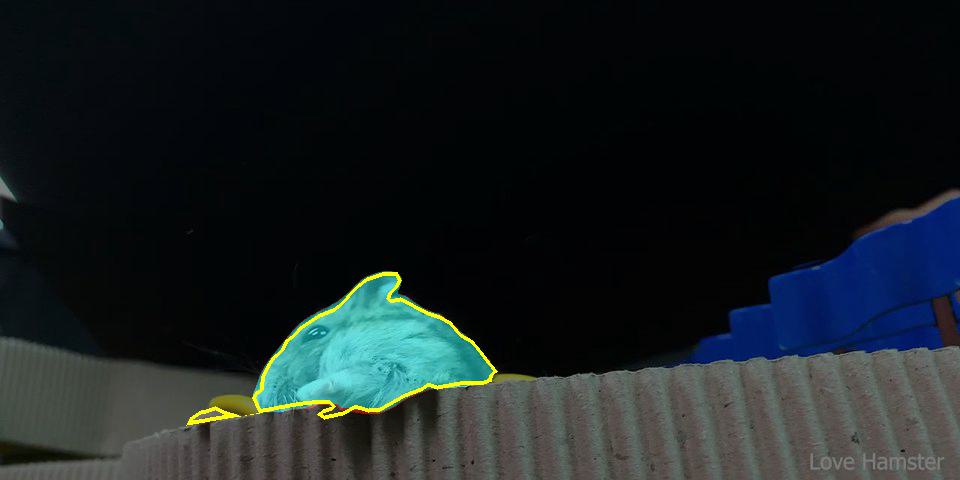}
			\put(-48.0,20.0){\makebox[0pt][l]{{\color{IBMGOLD}\textbf{\tiny \#4929}}}}}
		&\hspace{-4.5mm}
		\bmvaHangBox{\includegraphics[width=0.135\textwidth]{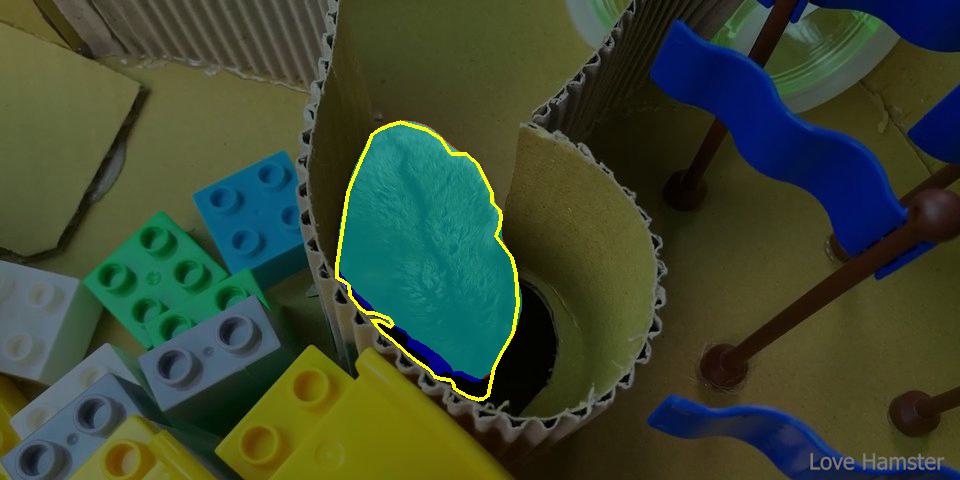}
			\put(-48.0,20.0){\makebox[0pt][l]{{\color{IBMGOLD}\textbf{\tiny \#5373}}}}}
		&\hspace{-4.5mm}
		\bmvaHangBox{\includegraphics[width=0.135\textwidth]{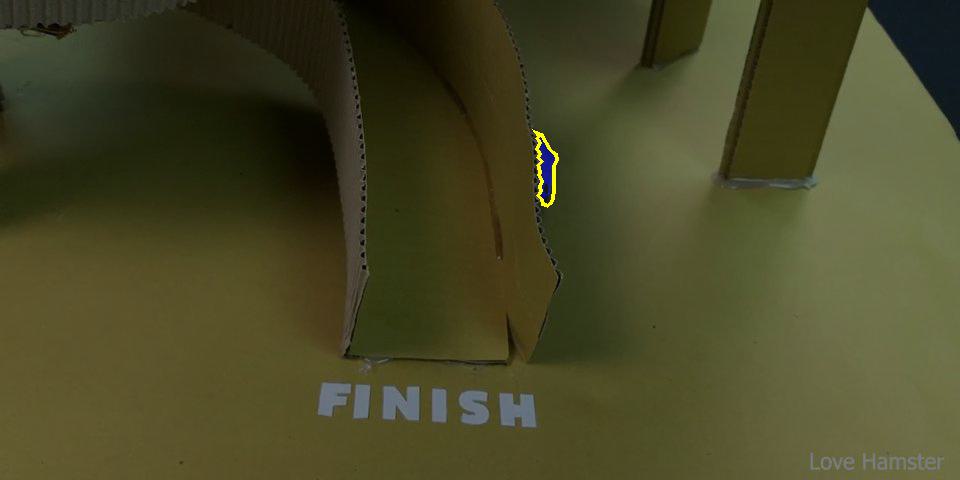}
			\put(-48.0,20.0){\makebox[0pt][l]{{\color{IBMGOLD}\textbf{\tiny \#6039}}}}}
		\\
	\centering {\footnotesize \STCN \vs READMem-STCN}
		&
		\bmvaHangBox{\includegraphics[width=0.135\textwidth]{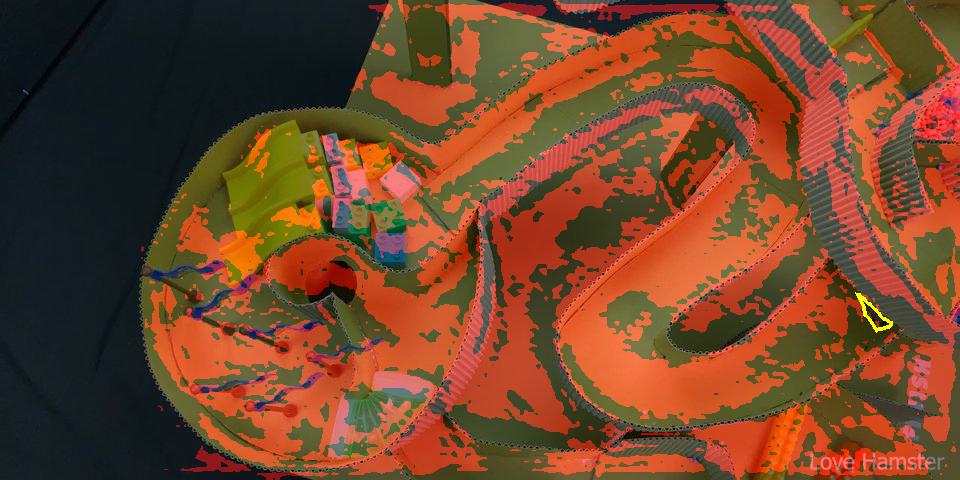}}
		&\hspace{-4.5mm}
		\bmvaHangBox{\includegraphics[width=0.135\textwidth]{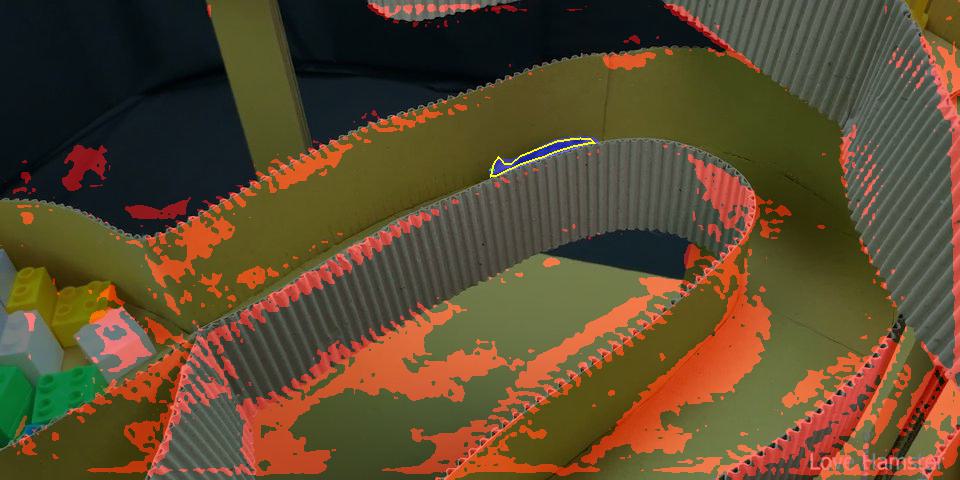}}
		&\hspace{-4.5mm}
		\bmvaHangBox{\includegraphics[width=0.135\textwidth]{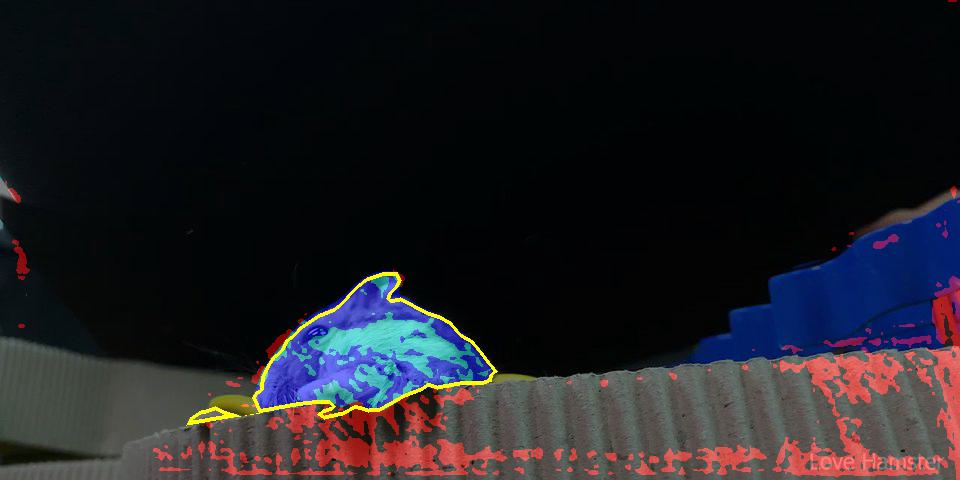}}
		&\hspace{-4.5mm}
		\bmvaHangBox{\includegraphics[width=0.135\textwidth]{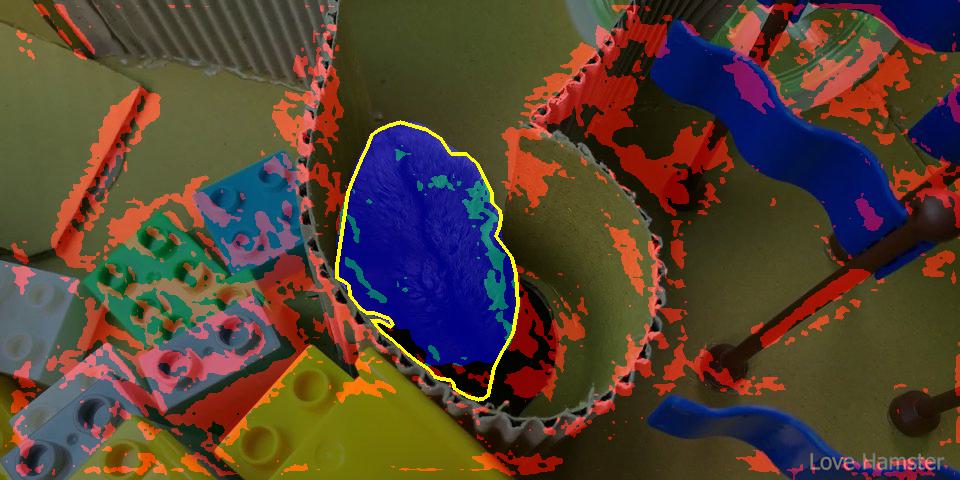}}
		&\hspace{-4.5mm}
		\bmvaHangBox{\includegraphics[width=0.135\textwidth]{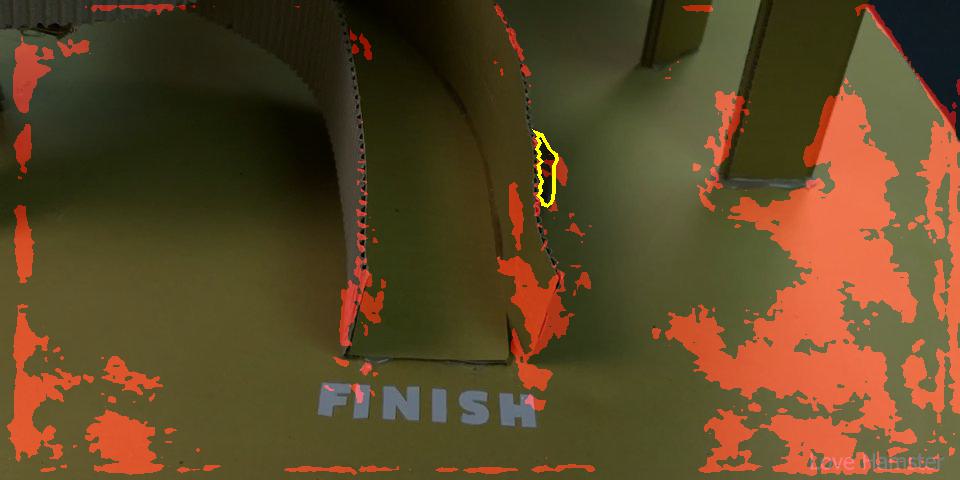}}
		\\
	\centering {\footnotesize \QDMN \vs READMem-QDMN}
		&
		\bmvaHangBox{\includegraphics[width=0.135\textwidth]{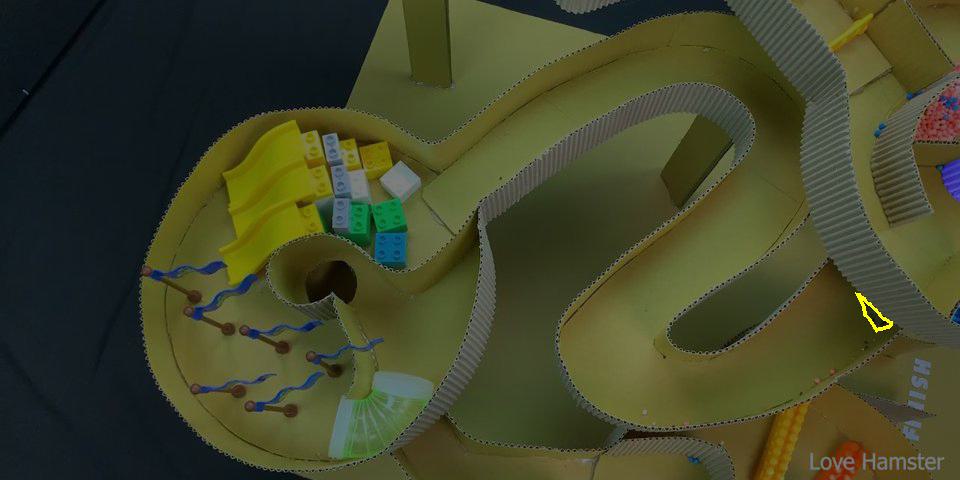}}
		&\hspace{-4.5mm}
		\bmvaHangBox{\includegraphics[width=0.135\textwidth]{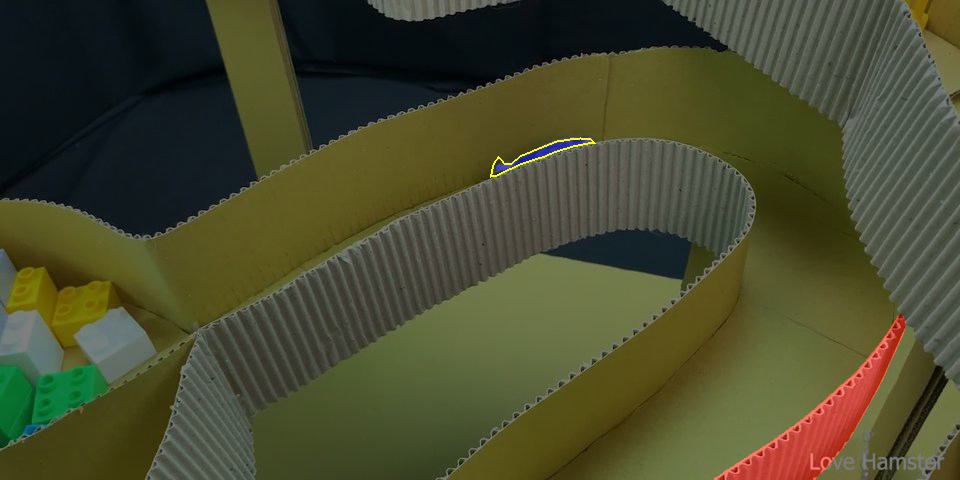}}
		&\hspace{-4.5mm}
		\bmvaHangBox{\includegraphics[width=0.135\textwidth]{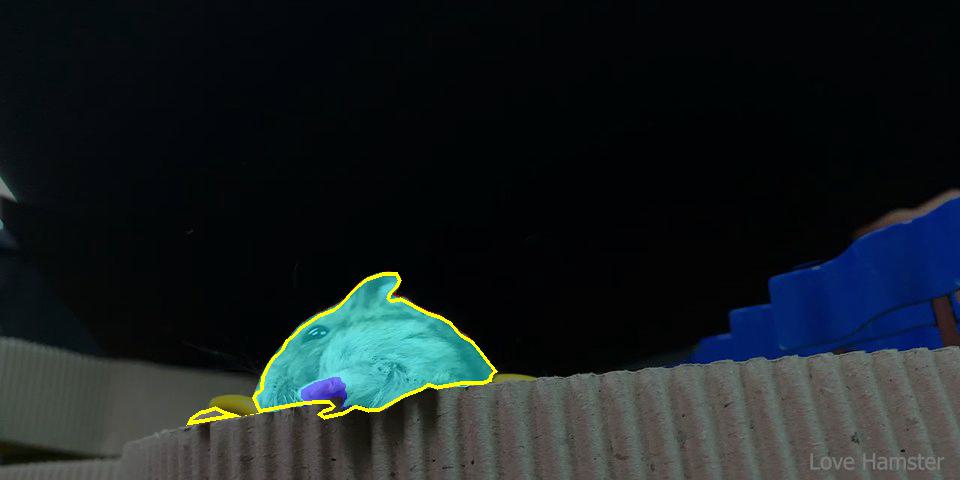}}
		&\hspace{-4.5mm}
		\bmvaHangBox{\includegraphics[width=0.135\textwidth]{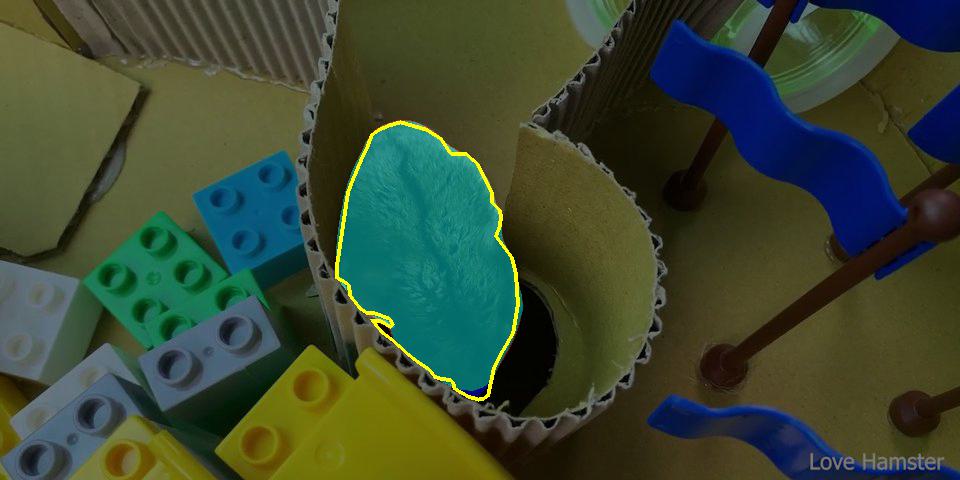}}
		&\hspace{-4.5mm}
		\bmvaHangBox{\includegraphics[width=0.135\textwidth]{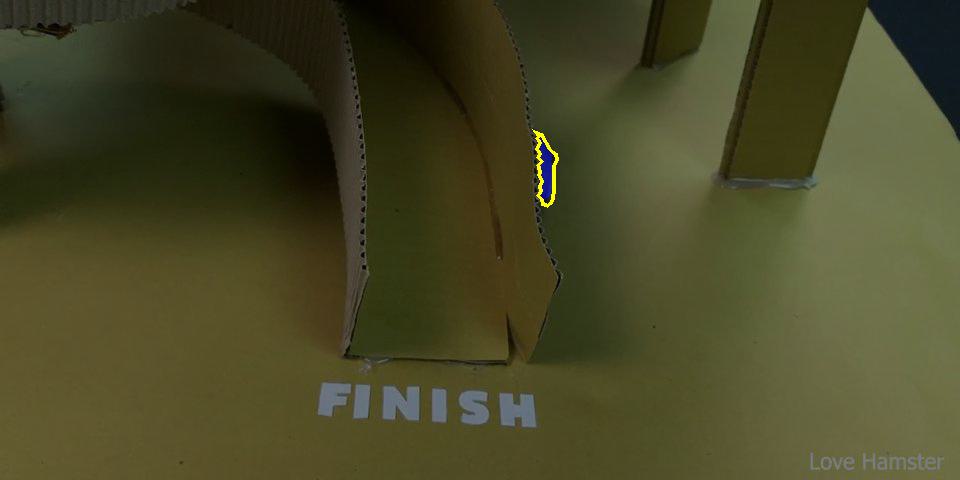}}
		\\
	\centering {\footnotesize \XMEM}
		&
		\bmvaHangBox{\includegraphics[width=0.135\textwidth]{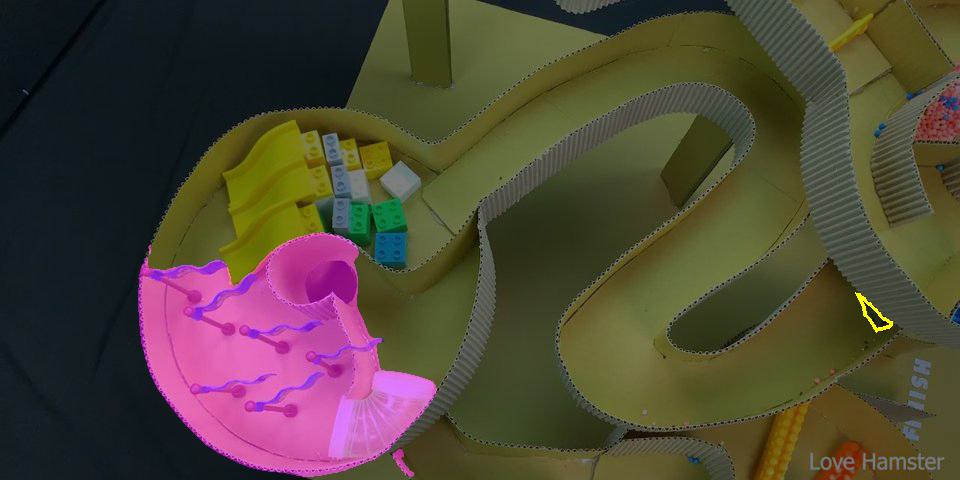}}
		&\hspace{-4.5mm}
		\bmvaHangBox{\includegraphics[width=0.135\textwidth]{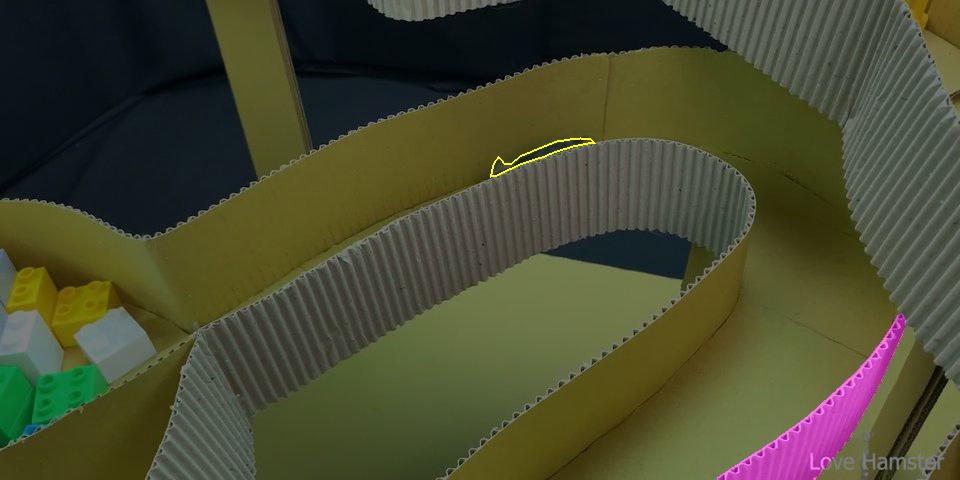}}
		&\hspace{-4.5mm}
		\bmvaHangBox{\includegraphics[width=0.135\textwidth]{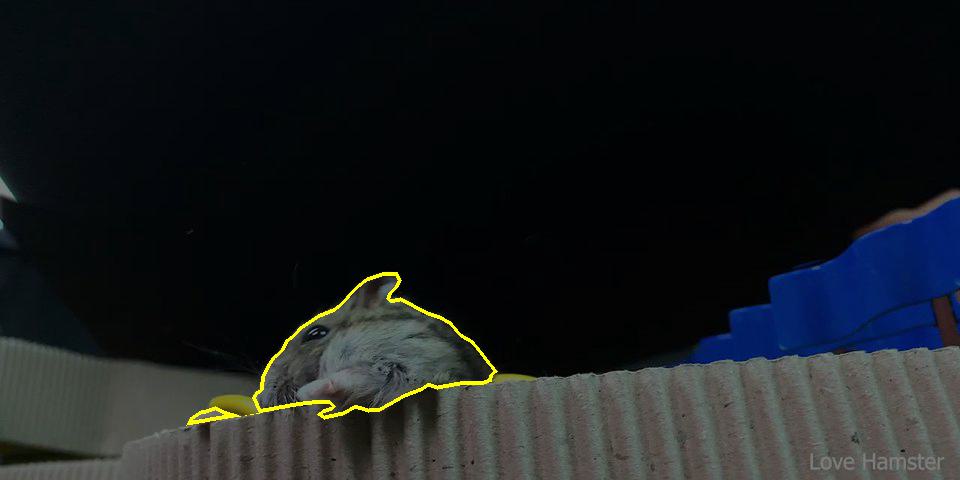}}
		&\hspace{-4.5mm}
		\bmvaHangBox{\includegraphics[width=0.135\textwidth]{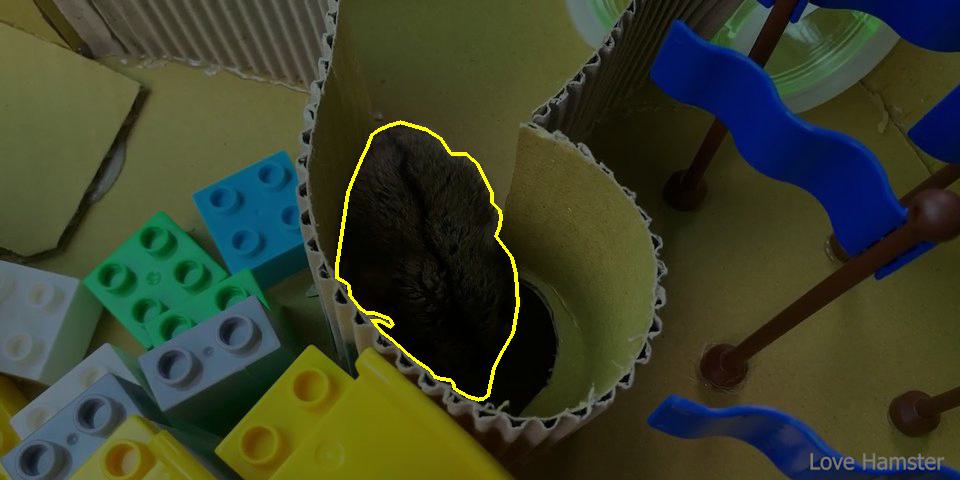}}
		&\hspace{-4.5mm}
		\bmvaHangBox{\includegraphics[width=0.135\textwidth]{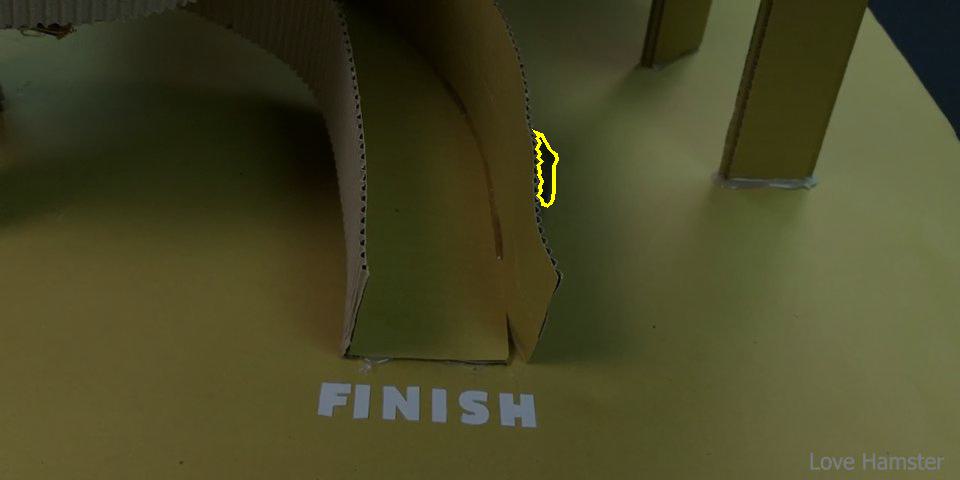}}
	\end{tabular}
	\caption{Results on the \textit{rat} sequence of \LV1 with $s_r = 1$.
	We depict the results of: the baselines in {\color{red!80!black}red}, the READMem variations in {\color{IBMINDIGO!}blue}, the intersection between both in {\color{IBMTURQUOISE!70!black}turquoise}, the ground-truth contours in {\color{IBMGOLD!70!black}yellow} and \XMEM results in {\color{IBMMAGENTA1!}purple}.
	}
	\label{fig:SUPP_RAT_sr_1}
\end{figure}

Figures~\ref{fig:SUPP_BLUEBOY_sr_10},~\ref{fig:SUPP_DRESSAGE_sr_10} and~\ref{fig:SUPP_RAT_sr_10} displays the results for the \textit{blueboy}, \textit{dressage} and \textit{rat} sequences in \LV1 respectively when using $s_r= 10$, while Figures~\ref{fig:SUPP_BLUEBOY},~\ref{fig:SUPP_DRESSAGE} and~\ref{fig:SUPP_RAT_sr_1} display the results for $s_r= 1$.
The estimated segmentation mask of the baselines (\MiVOS, \STCN, and \QDMN) are visualized in {\color{red!80!black}red}, while the results of the READMem-based variations (READMem with a baseline) are highlighted in {\color{IBMINDIGO!}blue}. 
The intersection between the prediction of a baseline and its corresponding READMem variation is depicted in {\color{IBMTURQUOISE!70!black}turquoise}. 
The ground-truth contours are highlighted in {\color{IBMGOLD!70!black}yellow}. 
We depict \XMEM results in {\color{IBMMAGENTA1!}purple}.

\subsection{Qualitative Results on \LV1 with $s_r = 10$}

\vspace{-5mm}
\vspace{-5mm}

\vspace{-5mm}

\subsection{Qualitative Results on \LV1 with $s_r = 1$}
	
\vspace{-5mm}

\vspace{-5mm}
	
\section{Additional Quantitative Evaluation}

We present in~\tabref{DATASETS_STATS} useful statistics for popular (sVOS)~\cite{Pont-Tuset_arXiv_2017, xu2018youtube, AFB-URR} and Visual Object Tracking (VOT)~\cite{kristan2022tenth, kristanvots2023} datasets. 
As our goal is to allow sVOS methods to perform on long video sequence, \tabref{DATASETS_STATS} reveals that the LV1 dataset~\cite{AFB-URR} and the recently introduced VOTS2023 dataset~\cite{kristanvots2023} are ideal candidates for assessing the effectiveness of our READMem extension.

Hence, in the main paper we focus on the LV1 dataset~\cite{AFB-URR}, to allow for a direct comparison with contemporary sVOS methods. 
We include the D17 dataset~\cite{Pont-Tuset_arXiv_2017} in our evaluations to encompass scenarios with shorter sequences.
However, to demonstrate the scalability and versatility of our approach, we also report complementary experiments on VOTS2023 in~\tabref{VOTS2023_Table}.
We want to clarify that our method is originally designed for managing the memory of sVOS task, and as such is not modifying the underlying architecture of the sVOS baselines~\cite{MiVOS, STCN, QDMN}, which are not tailored towards handling specific challenges found only in VOT datasets (\eg, small object-to-image ratio, presence of numerous distractors).

\begin{table*}[t]
	\centering
	\resizebox{1\columnwidth}{!}{
		\begin{tabular}{l|cccccc}
			\hline
			\bf{Dataset} & \bf{\# of sequences} & \bf{avg. length} & \bf{median length} & \bf{std.} & \bf{min. length} & \bf{max. length} \\
			\hline
			D17~\cite{Pont-Tuset_arXiv_2017}~\refConf{validation set} & $30$ & $67$ & $67$ & $21$ & $34$ & $104$ \\
			YVOS~\cite{xu2018youtube}~\refConf{validation set}  & $507$ & $134$ & $144$ & $42$ & $16$ & $180$ \\
			LV1~\cite{AFB-URR} 	 & $3$ & $2470$ & $2406$ & $1088$ & $1416$ & $3589$ \\
			\hline
			VOT2022~\cite{kristan2022tenth} 	 & $62$ & $321$ & $242$ & $295$ & $41$ & $1500$ \\
			VOTS2023~\cite{kristanvots2023} 	 & $144$ & $2073$ & $1810$ & $1856$ & $63$ & $10699$\\
			\hline
		\end{tabular}
	}
	\caption{
		Statistics of popular sVOS and VOT datasets. 
		For more details refer to the original publications.}
	\label{DATASETS_STATS}
\end{table*}

\subsection{Performance on the DAVIS (\D17) Dataset}
We display the performance of \MiVOS, \STCN and \QDMN with and without the READMem extension when varying the sampling interval $s_{r}$ on the \D17 dataset, using the same configuration as in~\secref{sec:quantitativexresults}.

\begin{figure}[t]
	\centering
	\begin{tabular}{c}
		\hspace{-4mm}
		\bmvaHangBox{\includegraphics[trim= {0 3mm 0 0},clip, width=0.98\textwidth]{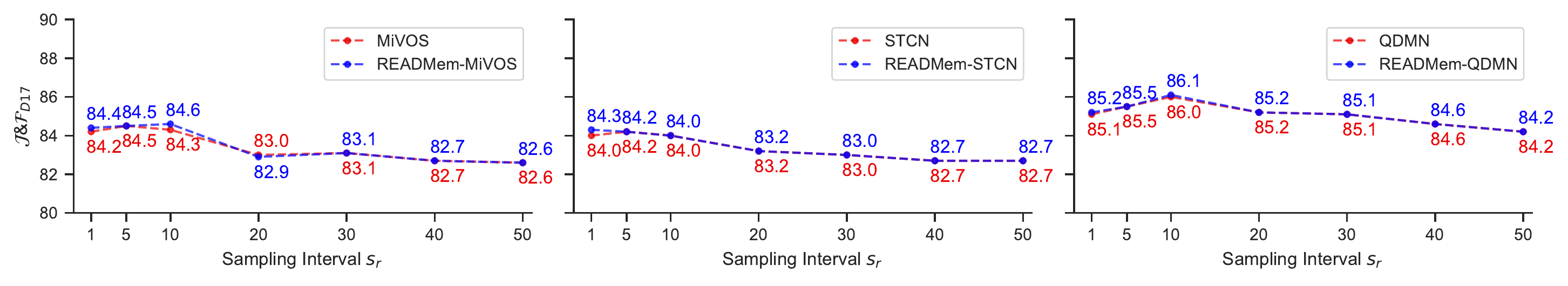}}
	\end{tabular}
	\begin{tabular}{P{4cm}P{4cm}P{4cm}}
		\hspace{+3mm}{\scriptsize (a) MiVOS \vs READMem-MiVOS} & \hspace{-4mm} {\scriptsize (b) STCN \vs READMem-STCN} & \hspace{-12mm} {\scriptsize (c) QDMN \vs READMem-QDMN}
	\end{tabular}
	\caption{Performance comparison of sVOS baselines (\MiVOS, \STCN, \QDMN) with and without the READMem extension on the \D17 dataset, while varying the sampling interval~$s_r$.
		Regardless of the final performance, we observe a general tendency where increasing the sampling interval (\ie, $s_r$ higher than $10$) on short video sequences leads to a performance drop.
	}
	\label{FIGxBaselinesxVSxHoliMemxLV1xALLxinxOnexDAVIS}
\end{figure}

In Figure \ref{FIGxBaselinesxVSxHoliMemxLV1xALLxinxOne} of \secref{sec:introduction}, we observe that increasing the sampling interval generally improves the performance of all methods on long videos, regardless of the baseline employed. 
However, this trend does not hold when working with short video sequences, as shown in Figure \ref{FIGxBaselinesxVSxHoliMemxLV1xALLxinxOnexDAVIS}. 
Here, we notice a degradation in performance for all methods when using larger sampling intervals.

\vspace{5mm}
Therefore, it is essential to utilize a sampling interval that does not negatively impact the performance on both long and short video sequences. 
This is where our READMem extension becomes valuable, as it enables the sVOS pipeline to use a small sampling interval (typically $s_{r} \in [1-10]$) that achieves and maintains high performance for both long and short video sequences.

\subsection{Performance as a Function of Memory Size}
We explore the impact of the size of the memory on the performance of \MiVOS, \STCN and \QDMN with and without our READMem extension on the \LV1 dataset.
We follow the same experimental setup as in~\secref{sec:quantitativexresults} (with $s_r=10$),  except for the varying memory size $N$, which ranges from $5$ to $50$.

\begin{figure}[b]
	\centering
	\begin{tabular}{c}
		\hspace{-4mm}
		\bmvaHangBox{\includegraphics[trim= {0 3mm 0 0},clip, width=0.98\textwidth]{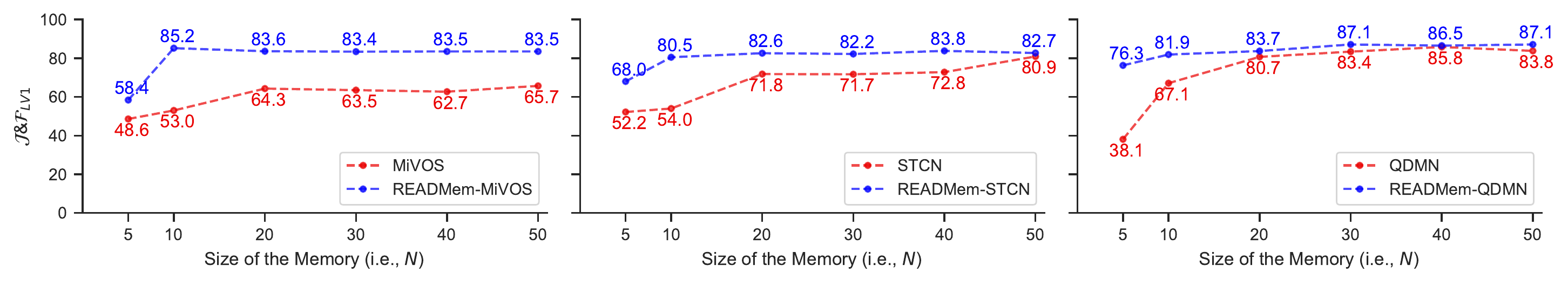}}
	\end{tabular}
	\begin{tabular}{P{4cm}P{4cm}P{4cm}}
		\hspace{+3mm}{\scriptsize (a) MiVOS \vs READMem-MiVOS} & \hspace{-4mm} {\scriptsize (b) STCN \vs READMem-STCN} & \hspace{-12mm} {\scriptsize (c) QDMN \vs READMem-QDMN}
	\end{tabular}
	\caption{We compare the performance of sVOS baselines (\MiVOS, \STCN, \QDMN) with and without the READMem extension on the \LV1 dataset while varying the size of the memory (\ie, $N$).
	A general tendency is that increasing the memory size, leads to better performance.
	}
	\label{FIGxBaselinesxVSxHoliMemxLV1xALLxinxOnexDAVISxMemory}
\end{figure}

From~\figref{FIGxBaselinesxVSxHoliMemxLV1xALLxinxOnexDAVISxMemory}, we observe that the performance of the baselines improves as the memory size increases.
Similarly, although to a lesser extent, the READMem variants also demonstrate improved performance with larger memory sizes.
However, the READMem variations consistently outperform their respective baselines, especially when using a smaller memory size. 
This is desired as a smaller memory requires less GPU resources.

Comparing~\figref{FIGxBaselinesxVSxHoliMemxLV1xALLxinxOnexDAVISxMemory} with~\figref{FIGxBaselinesxVSxHoliMemxLV1xALLxinxOne},
we notice that increasing the sampling interval (\ie, $s_r$) of the baselines leads to a significant boost in performance compared to increasing the memory size~(\ie, $N$).
Hence, storing a diverse set of embeddings in the memory is more beneficial than including additional ones.

\subsection{Performance on the VOTS2023~\cite{kristanvots2023} Dataset}
In our quantitative evaluation (refer to~\tabref{tab_TableQuantitativeResults} of \secref{sec:experiments}), we demonstrate and analyze the effectiveness of our approach on sVOS datasets, encompassing both short~(\ie, D17~\cite{Pont-Tuset_arXiv_2017}) and long~(\ie, LV1~\cite{AFB-URR}) sequences, to allow for a direct comparison with contemporary sVOS approaches~(\ie, \cite{AFB-URR, XMEM}).
In an effort, to enhance the soundness of our READMem extension, we conduct additional experiments on the VOTS2023 dataset~\cite{kristanvots2023}. 
We tabulate in~\tabref{VOTS2023_Table}, the results of sVOS baselines~\cite{MiVOS,STCN,QDMN} with and without READMem on the VOTS2023 tracking benchmark. 

For the evaluation we use the same settings as described in~\secref{sec:quantitativexresults}~(refer to quantitative results) and the official VOT evaluation toolkit (version 0.6.4 released on the 31 May 2023 \textendash~\url{https://github.com/votchallenge/toolkit}).
We observe from~\tabref{VOTS2023_Table}, that the READMem variants consistently outperform their baseline counterpart.

In contrast to previous VOT challenges~\cite{kristan2020eighth, kristan2021ninth, kristan2022tenth}, VOTS2023 introduced new evaluation metrics split into:
(i) a primary performance metric: The Tracking Quality (\textbf{Q}) and 
(ii) secondary metrics: the Accuracy (\textbf{ACC}), Robustness (\textbf{ROB}), Not-Reported Error (\textbf{NRE}), Drift-Rate Error (\textbf{DRE}) and Absence-Detection Quality (\textbf{ADQ}).
Please refer to the VOTS2023 paper for more details.

\begin{table*}[t]
	\centering
	\resizebox{1\columnwidth}{!}{
		\begin{tabular}{l|c|ccc|cc}
			\hline
			& \multicolumn{4}{c}{\refConf{Higher is better}} & \multicolumn{2}{c}{\refConf{Lower is better}}\\
			\bf{Method} 												& \bf{Q} 				& \bf{ACC} 				& \bf{ROB} 				& \bf{ADQ} 					& \bf{NRE} 					& \bf{DRE} 				 \\
			\hline
			MiVOS~\cite{MiVOS}~\refConf{CVPR 21} 			  		   	& $0.38$ 				& $0.55$ 				& $0.54$ 				& $0.75$ 				    & $0.41$ 					& $0.06$ 				 \\
			MiVOS~\cite{MiVOS} ($s_r = 50$)~\refConf{CVPR 21} 		   	& $0.39^{\uparrow0.01}$ & $0.55$ 				& $0.58^{\uparrow0.04}$ & $0.67^{\downarrow0.08}$   & $0.35^{\downarrow0.06}$ 	& $0.07^{\uparrow0.01}$  \\
			{\color{IBMINDIGO!80!black}READMem-MiVOS}~\refConf{ours}   	& $0.43^{\uparrow0.05}$ & $0.57^{\uparrow0.02}$ & $0.60^{\uparrow0.06}$ & $0.67^{\downarrow0.08}$   & $0.33^{\downarrow0.08}$ 	& $0.06$ 				 \\
			\hline
			STCN~\cite{STCN}~\refConf{NIPS 21} 				  			& $0.40$ 				& $0.55$ 				& $0.62$ 					& $0.67$ 					& $0.29$ 					& $0.08$ 				\\
			STCN~\cite{STCN} ($s_r = 50$)~\refConf{NIPS 21}   			& $0.40$ 				& $0.55$ 				& $0.61^{\downarrow0.01}$	& $0.61^{\downarrow0.06}$ 	& $0.29$ 					& $0.10^{\uparrow0.02}$ \\
			{\color{IBMINDIGO!80!black}READMem-STCN}~\refConf{ours}    	& $0.42^{\uparrow0.02}$ & $0.56^{\uparrow0.01}$ & $0.66^{\uparrow0.04}$ 	& $0.57^{\downarrow0.10}$ 	& $0.25^{\downarrow0.04}$ 	& $0.09^{\uparrow0.01}$ \\
			\hline
			QDMN~\cite{QDMN}~\refConf{ECCV 22} 				  			& $0.44$ 					& $0.59$ & $0.62$ 					& $0.69$ 				  & $0.28$ 					& $0.10$ 				  \\
			QDMN~\cite{QDMN} ($s_r = 50$)~\refConf{ECCV 22}   			& $0.42^{\downarrow0.02}$ 	& $0.59$ & $0.60^{\downarrow0.02}$ 	& $0.63^{\downarrow0.06}$ & $0.30^{\uparrow0.02}$ 	& $0.11^{\uparrow0.01}$   \\
			{\color{IBMINDIGO!80!black}READMem-QDMN}~\refConf{ours}	   	& $0.45^{\uparrow0.01}$ 	& $0.59$ & $0.63^{\uparrow0.01}$ 	& $0.67^{\downarrow0.02}$ & $0.27^{\downarrow0.01}$ & $0.09^{\downarrow0.01}$ \\
			\hline
		\end{tabular}
	}
	\caption{
		Quantitative evaluation of sVOS methods~\cite{MiVOS,STCN,QDMN, XMEM} with and without READMem on the VOTS2023~\cite{kristanvots2023} datasets. We use the same settings as described in~\secref{sec:quantitativexresults} and the official VOT evaluation toolkit. 
	}
	\label{VOTS2023_Table}
\end{table*}

\section{Initialization of the Memory}
We investigate the performance variation when employing two different initialization for READMem in~\tabref{tab_initialization}: 
The strategies are as follows:
(1) integrates every~$t$-th frame into the memory until full, while 
(2) fills the memory slots with the embeddings of the annotated frame and includes a new frame to the memory if the conditions on the lower bound on similarity and the Gramian are met (follows a greedy approach). 
The second strategy yields worse results on the short scenarios and is slightly below the performance of strategy (1) on~\LV1. 
We argue that with longer sequences the memory has more opportunities to integrate decisive frame representations in the memory to use as a reference.
Hence, initialization plays a crucial role in short videos, but as the method observes longer videos and has access to a larger pool of frames to select from, the importance diminishes.

\begin{table}[b]
	\centering
		\begin{tabular}{c|cc|cc|cc}
			\hline
			&\multicolumn{2}{c}{\bf{READMem-MiVOS}} & \multicolumn{2}{c}{\bf{READMem-STCN}} & \multicolumn{2}{c}{\bf{READMem-QDMN}}\\
			\bf{Initialization} & \JandFLV1 & \JandFD17 & \JandFLV1 & \JandFD17 & \JandFLV1 & \JandFD17\\
			\hline
			(1)       &  $83.6$   &  $84.6$   &  $82.6$   &  $84.0$   &  $84.0$   &  $86.1$	  \\
			(2)       &  $82.7$   &  $73.7$   &  $85.3$   &  $73.6$   &  $72.5$   &  $73.3$   \\ \hline
		\end{tabular}
	\caption{Performance variation when leveraging two different initialization strategies for READMem-MiVOS. 
	Besides the initialization strategy, the remaining parameters are consistent to~\secref{sec:quantitativexresults} (we set $s_r = 10$).
	}
	\label{tab_initialization}
\end{table}

\section{Discussion and Limitations}
\label{sec:discussion}
We are aware of the limitations imposed by the hand-crafted threshold for the lower similarity bound~$l_{sb}$, although to avoid any fine-tuning, we set the threshold value to~$0.5$.
A more thoughtful approach would incorporate a learnable parameter. 
This approach could potentially lead to improved performance, albeit at the expense of the plug-and-play nature of our extension.
Another point for improvement is to reduce the participation of the background when computing the similarity between two embeddings.
A possible enhancement is to integrate either the segmentation mask estimated by the sVOS pipeline or use the memory values to estimate a filter that can be applied to the memory keys before computing a similarity score. 

\section{Training}
For our experiments, we utilize the original weights provided by the authors of MiVOS~\cite{MiVOS}, STCN~\cite{STCN}, and QDMN~\cite{QDMN}. 
Our primary focus is to showcase the benefits of our extension (\ie, READMem) without modifying the baselines.
To provide a comprehensive overview of the baselines, we briefly elaborate on the training methodology. 
The training procedure follows the regiment presented in STM~\cite{STM} and refined in the subsequent work, MiVOS~\cite{MiVOS}.
The training is divided into two stages employing the bootstrapped cross-entropy loss~\cite{MiVOS} and utilizing the Adam optimizer (refer to the original papers~\cite{MiVOS,STCN,QDMN} and their supplementary materials for detailed insights).

The training comprises the following stages: 
(1) A pre-training stage, in which static image datasets are used as in~\cite{STM} to simulate videos consisting of three frames. 
While all three frames originate from the same image, the second and third frames are modified using random affine transformations
(2) A main-training stage, which uses the DAVIS~\cite{Pont-Tuset_arXiv_2017}~ and the Youtube-VOS~\cite{xu2018youtube} datasets (which provide real videos).
Similar to the pre-training stage, three frames from a video are sampled, gradually increasing the temporal gap from 5 to 25 frames during training. Subsequently, the temporal gap is annealed back to 5 frames, following a curriculum training approach~\cite{zhang2020spatial}.
(Optional) Moreover, after the pre-training stage a synthetic dataset BL30K~\cite{MiVOS} can be leveraged to enhance the ability of the model to better handle complex occlusion patterns.